%% file: main.tex
\documentclass[10pt,twocolumn,letterpaper]{article}

\newif\ifhidepagenum
\hidepagenumfalse
\newif\ifusehyperref
\usehyperreftrue
\input{preamble}

\begin{document}

\title{Tracking Anything with Decoupled Video Segmentation}

\author{Ho Kei Cheng\textsuperscript{1}\thanks{This work was done during an internship at Adobe Research.} \hspace{1em} Seoung Wug Oh\textsuperscript{2} \hspace{1em} Brian Price\textsuperscript{2} \hspace{1em} Alexander Schwing\textsuperscript{1} \hspace{1em} Joon-Young Lee\textsuperscript{2}\\
\textsuperscript{1}University of Illinois Urbana-Champaign \hspace{2em} \textsuperscript{2}Adobe Research\\
{\tt\small\{hokeikc2,aschwing\}@illinois.edu, \{seoh,bprice,jolee\}@adobe.com}
\vspace{-1ex}
}

\input{00-abstract}
\input{01-introduction}
\input{02-related_works}
\input{03-method}
\input{04-experiments}
\input{05-conclusion}

{\small
\bibliographystyle{ieee_fullname}
\bibliography{main}
}

\clearpage
\input{09-appendix}

\end{document}

%% file: preamble.tex
\usepackage{titlesec}
\titlespacing*{\paragraph}{0pt}{.5ex plus .5ex minus .5ex}{1em}

\usepackage[dvipsnames]{xcolor}
\usepackage{iccv}
\usepackage{times}
\usepackage{epsfig}
\usepackage{graphicx}
\usepackage{amsmath}
\usepackage{amssymb}
\usepackage{color}
\usepackage{colortbl}
\usepackage{booktabs}
\usepackage[format=plain,labelformat=simple,labelsep=period,font=small,compatibility=false]{caption}
\usepackage{multirow}
\usepackage{makecell}
\usepackage{comment}
\usepackage{pifont}
\usepackage{enumitem}
\usepackage{overpic}
\usepackage[accsupp]{axessibility}

\usepackage{tabularx,stackengine,collcell}
\let\endminwd\relax
\newcolumntype{L}[1]{>{\collectcell\xminwd l{#1}}l<{\endminwd\endcollectcell}}
\newcolumntype{C}[1]{>{\collectcell\xminwd c{#1}}c<{\endminwd\endcollectcell}}
\newcolumntype{R}[1]{>{\collectcell\xminwd r{#1}}r<{\endminwd\endcollectcell}}
\def\minwd#1#2#3\endminwd{\stackengine{0pt}{#3}{\rule{#2}{0pt}}{O}{#1}{F}{F}{L}}
\newcommand\xminwd[1]{\minwd#1}

\usepackage{pgfplots}
\pgfplotsset{compat=1.9}

\ifusehyperref
\usepackage[pagebackref,breaklinks,colorlinks]{hyperref}
\else
\usepackage[pagebackref,breaklinks,colorlinks,bookmark=false]{hyperref}
\fi
\hypersetup{
	colorlinks,
	linkcolor={red!80!black},
	citecolor={blue!80!black},
	urlcolor={blue!80!black},
    pdftitle={Tracking Anything with Decoupled Video Segmentation},
}

\iccvfinalcopy 

\ifhidepagenum
    \ificcvfinal\fi
\fi

\newcommand{\mj}{$\mathcal{J}$}
\newcommand{\mf}{$\mathcal{F}$}
\newcommand{\mjf}{$\mathcal{J}\&\mathcal{F}$}

\newcommand{\mjs}{$\mathcal{J}_s$}
\newcommand{\mfs}{$\mathcal{F}_s$}
\newcommand{\mju}{$\mathcal{J}_u$}
\newcommand{\mfu}{$\mathcal{F}_u$}
\newcommand{\mg}{$\mathcal{G}$}

\newcommand{\vpq}{\text{VPQ}}
\newcommand{\prop}{\text{Prop}}
\newcommand{\imseg}{\text{Seg}}
\newcommand{\seg}{\mathbf{M}}
\newcommand{\mem}{\mathbf{H}}
\newcommand{\proposal}{\mathbf{P}}
\newcommand{\consen}{\mathbf{C}}
\newcommand{\propseg}{\mathbf{R}}
\newcommand{\readout}{\mathbf{F}}

\makeatletter
\def\@fnsymbol#1{\ensuremath{\ifcase#1\or \dagger\or \ddagger\or
		\mathsection\or \mathparagraph\or \|\or **\or \dagger\dagger
		\or \ddagger\ddagger \else\@ctrerr\fi}}
\makeatother

\newcommand{\beginsupplement}{
	\setcounter{table}{0}
	\renewcommand{\thetable}{S\arabic{table}}%
	\setcounter{figure}{0}
	\renewcommand{\thefigure}{S\arabic{figure}}%
	\setcounter{equation}{0}
	\renewcommand{\theequation}{S\arabic{equation}}
}

%% file: 00-abstract.tex
\twocolumn[{%
\renewcommand\twocolumn[1][]{#1}%
\ifhidepagenum
    \pagenumbering{gobble} 
\fi
\maketitle
\centering
\captionsetup{type=figure}
\input{figs/fig-teaser}
\captionof{figure}{
    \small 
    Visualization of our semi-online video segmentation results. 
    Top: our algorithm (\textbf{DEVA}) extends Segment Anything (SAM)~\cite{kirillov2023segment} to video for open-world video segmentation with no user input required.
    Bottom: DEVA performs text-prompted video segmentation for novel objects (with prompt \textit{``beyblade''}, a type of spinning-top toy) by integrating Grounding-DINO~\cite{liu2023grounding} and SAM~\cite{kirillov2023segment}. 
    }%
\label{fig:teaser}%
\vspace{2ex}
}]

\begin{abstract}
Training data for video segmentation are expensive to annotate. 
This impedes extensions of end-to-end algorithms to new video segmentation tasks, especially in large-vocabulary settings. 
To `track anything' without training on video data for every individual task, we develop a \textbf{de}coupled \textbf{v}ideo segment\textbf{a}tion approach (\textbf{DEVA}), composed of task-specific image-level segmentation and class/task-agnostic bi-directional temporal propagation.
Due to this design, we only need an image-level model for the target task (which is cheaper to train) and a universal temporal propagation model which is trained once and generalizes across tasks.
To effectively combine these two modules, we use bi-directional propagation for (semi-)online fusion of segmentation hypotheses from different frames to generate a coherent segmentation.
We show that this decoupled formulation compares favorably to end-to-end approaches in several data-scarce tasks including large-vocabulary video panoptic segmentation, open-world video segmentation, referring video segmentation, and unsupervised video object segmentation.
Code is available at: {\href{https://hkchengrex.github.io/Tracking-Anything-with-DEVA}{\nolinkurl{hkchengrex.github.io/Tracking-Anything-with-DEVA}}.}
\end{abstract}

%% file: figs/fig-teaser.tex
\centering
\begin{tabular}{c@{\hspace{2pt}}c@{\hspace{2pt}}c}
\includegraphics[width=0.32\linewidth]{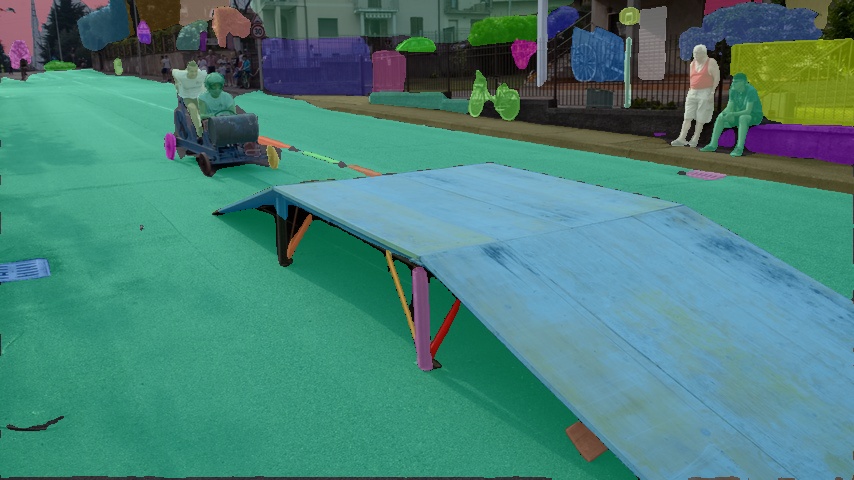} &
\includegraphics[width=0.32\linewidth]{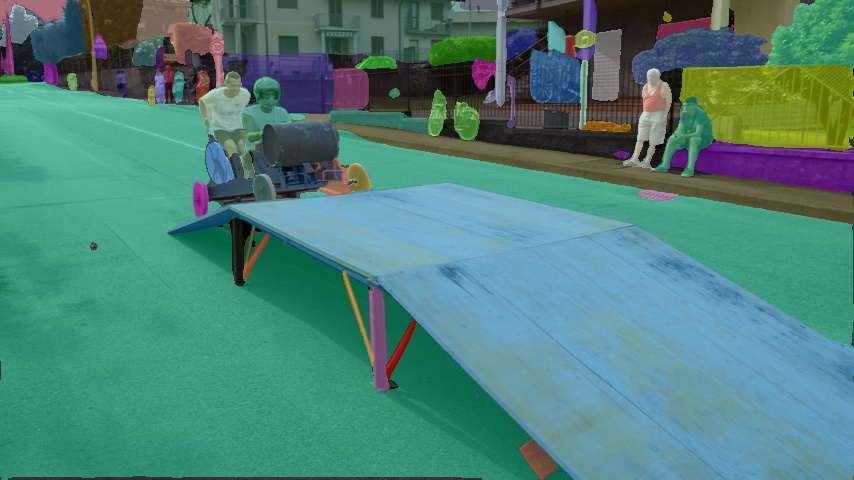} &
\includegraphics[width=0.32\linewidth]{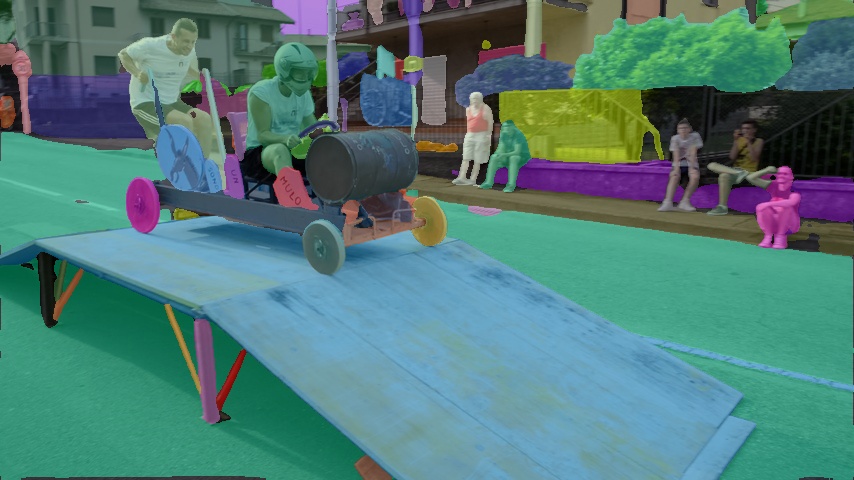} \\
\includegraphics[width=0.32\linewidth]{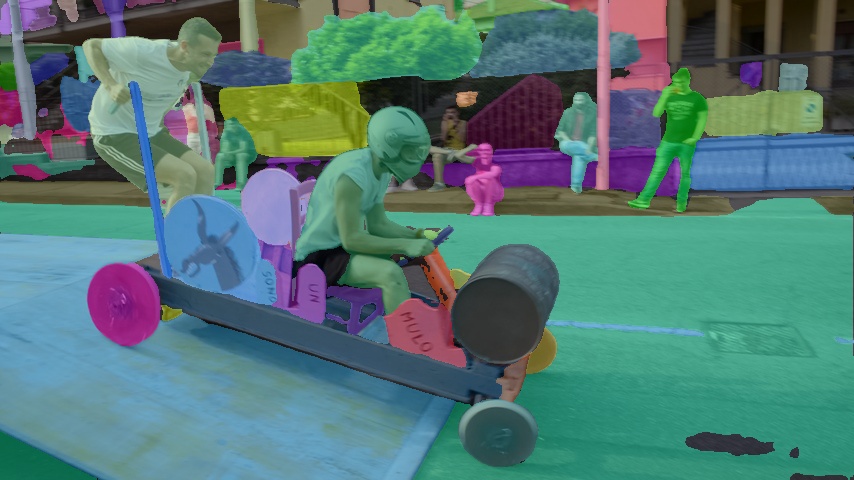} &
\includegraphics[width=0.32\linewidth]{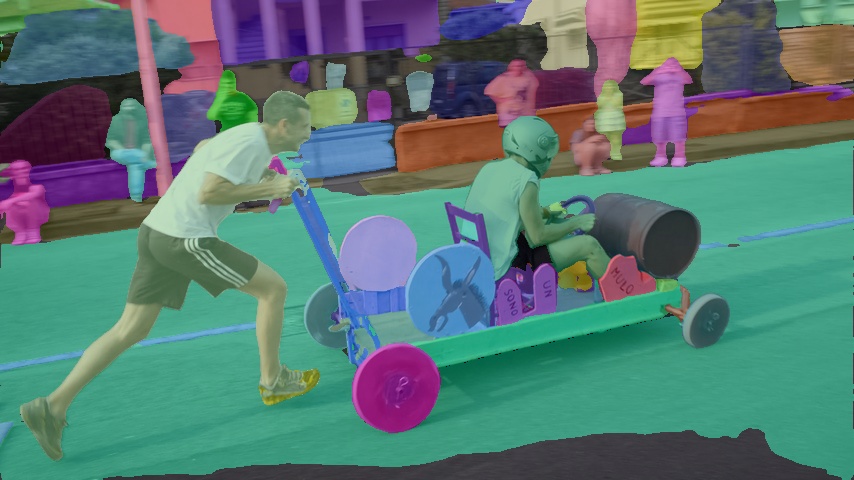} &
\includegraphics[width=0.32\linewidth]{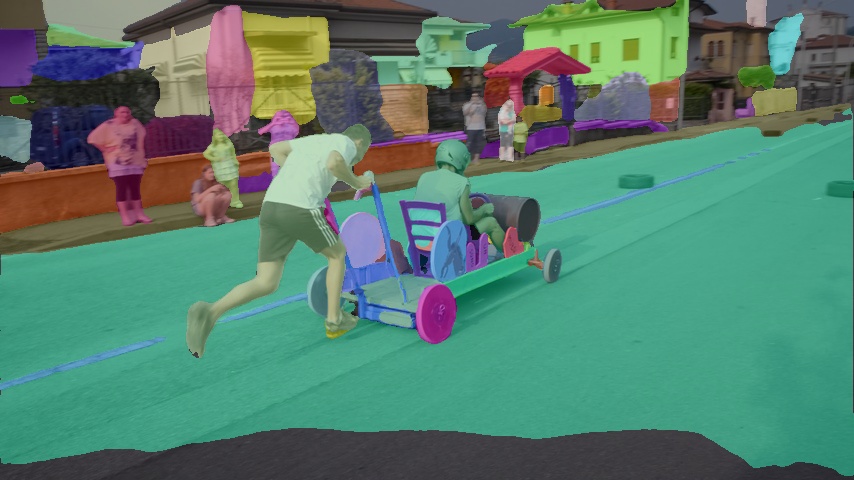} 
\end{tabular}
\begin{tabular}{c@{\hspace{2pt}}c@{\hspace{2pt}}c@{\hspace{2pt}}c@{\hspace{2pt}}c}
\includegraphics[width=0.19\linewidth]{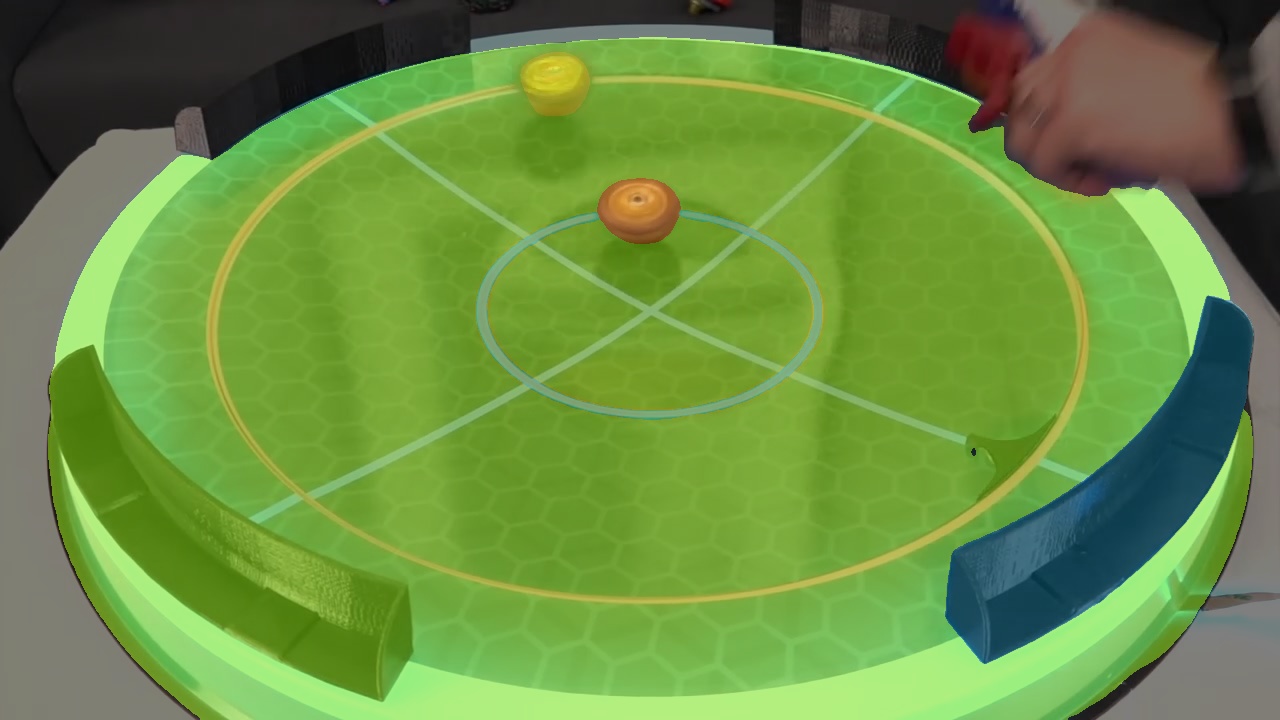} &
\includegraphics[width=0.19\linewidth]{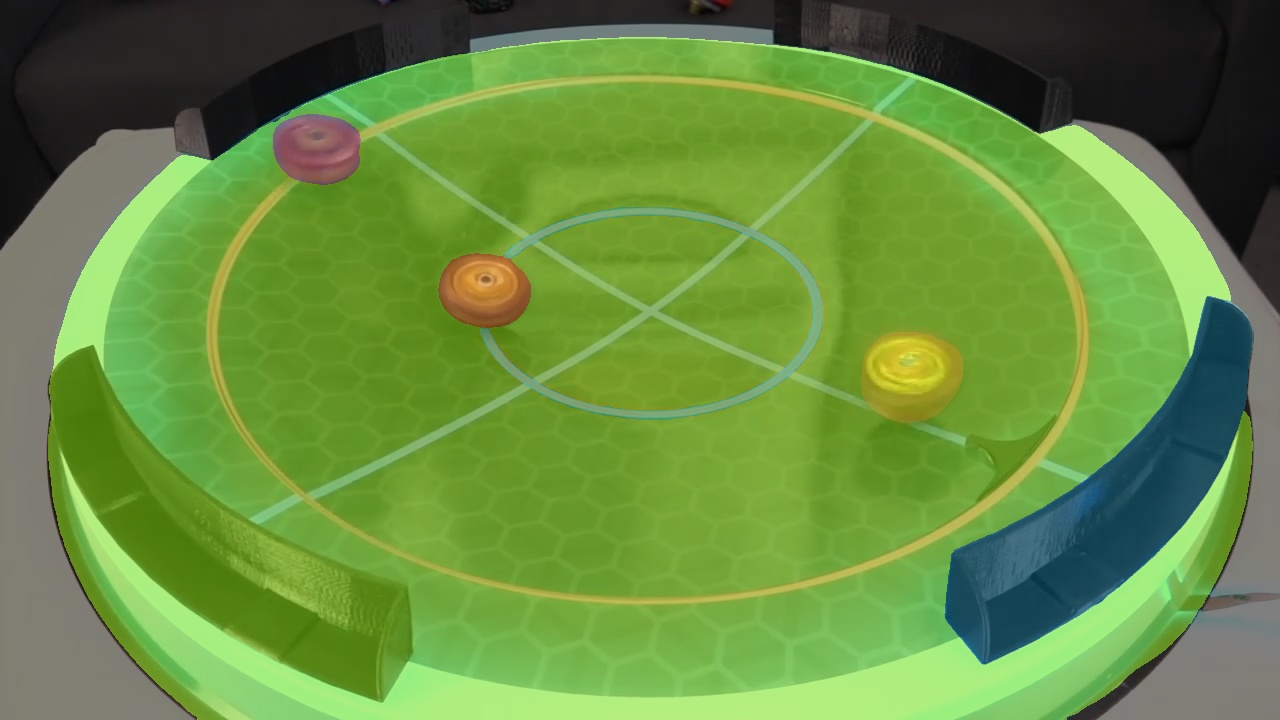} &
\includegraphics[width=0.19\linewidth]{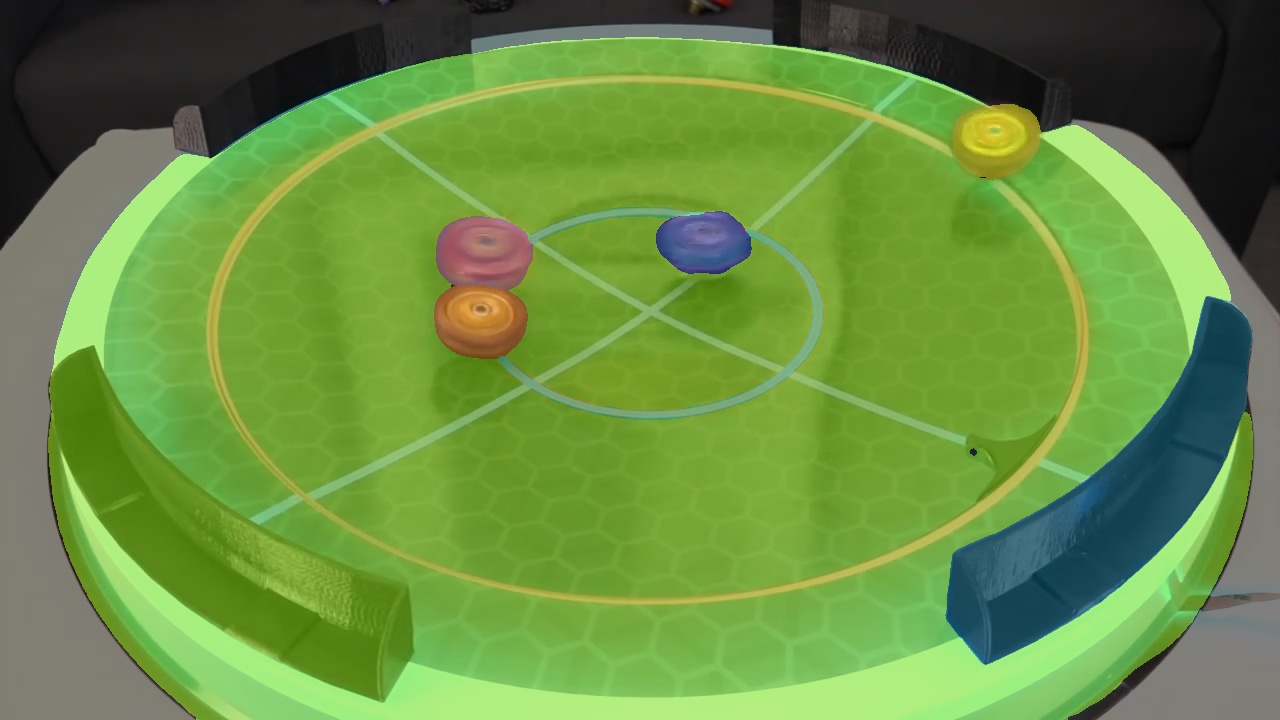} &
\includegraphics[width=0.19\linewidth]{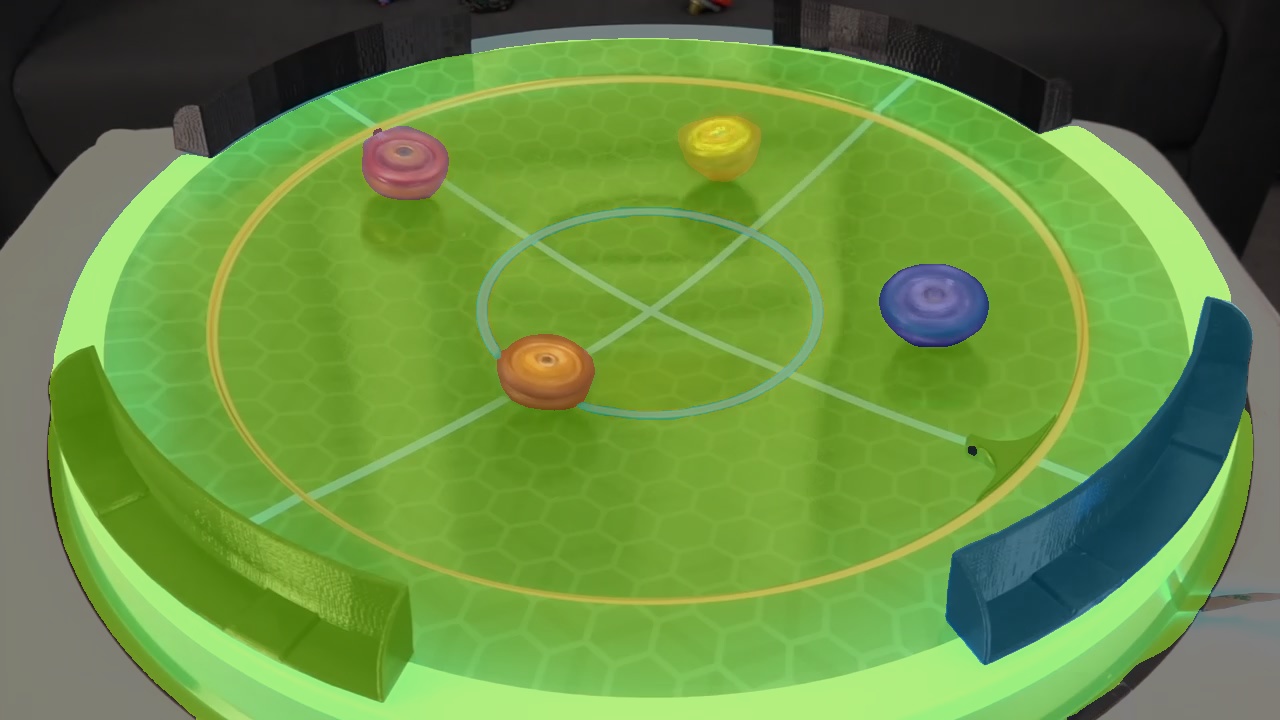} &
\includegraphics[width=0.19\linewidth]{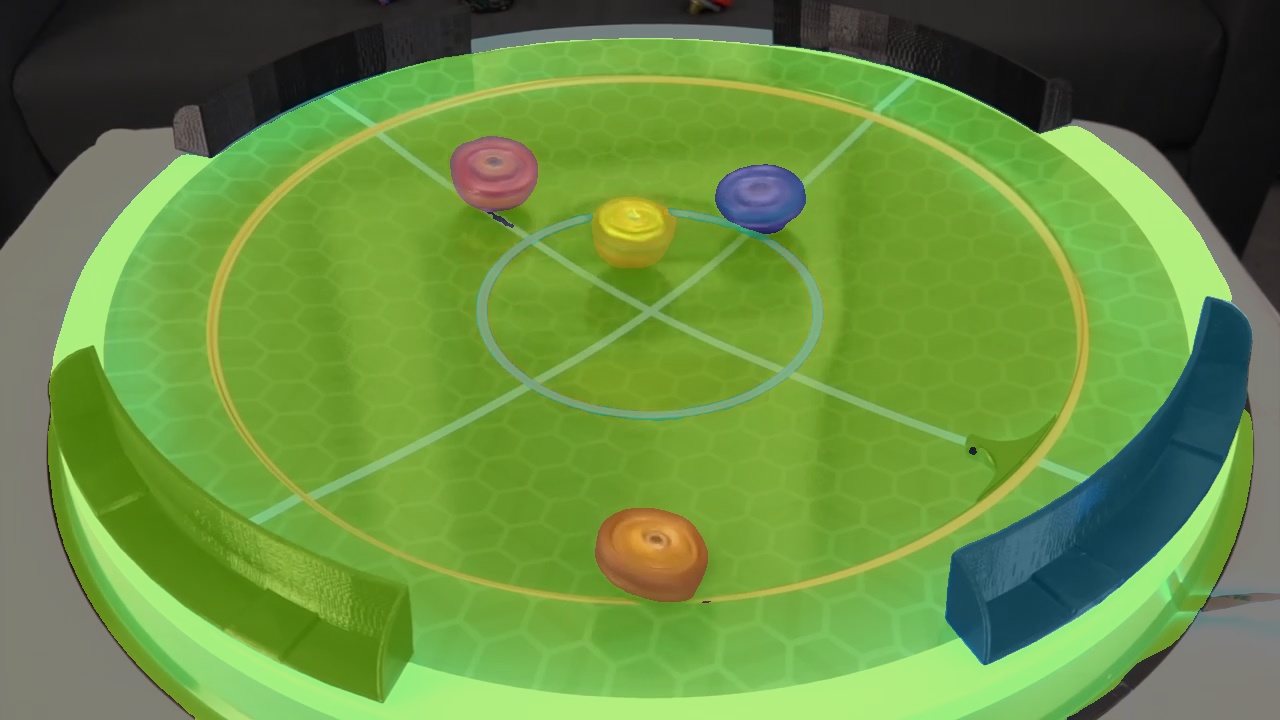} 
\end{tabular}

%% file: 01-introduction.tex
\section{Introduction}

Video segmentation aims to segment and associate objects in a video. It is a fundamental task in computer vision and is crucial for many video understanding applications.

Most existing video segmentation approaches train end-to-end video-level networks on annotated video datasets. They have made significant strides on common benchmarks like YouTube-VIS~\cite{yang2019video} and Cityscape-VPS~\cite{kim2020video}.
However, these datasets have small vocabularies: YouTube-VIS contains 40 object categories, and Cityscape-VPS only has 19.
It is questionable whether recent end-to-end paradigms are scalable to large-vocabulary, or even open-world video data.  
A recent larger vocabulary (124 classes) video segmentation dataset, VIPSeg~\cite{miao2022large}, has been shown to be more difficult -- using the same backbone, a recent method~\cite{li2022video} achieves only 26.1 VPQ compared with 57.8 VPQ on Cityscape-VPS.
To the best of our knowledge, recent video segmentation methods~\cite{athar2023burst,liu2022opening} developed for the open-world setting (e.g., BURST~\cite{athar2023burst}) are not end-to-end and are based on tracking of per-frame segmentation -- further highlighting the difficulty of end-to-end training on large-vocabulary datasets.
As the number of classes and scenarios in the dataset increases, it becomes more challenging to train and develop end-to-end video models to jointly solve segmentation and association, especially if annotations are scarce.

In this work, we aim to reduce reliance on the amount of target training data by leveraging external data \textit{outside of the target domain}. For this, we propose to study \textit{decoupled video segmentation}, which combines task-specific image-level segmentation and task-agnostic temporal propagation.
Due to this design, we only need an image-level model for the target task (which is cheaper) and a 
universal temporal propagation model which is trained once and generalizes across tasks.
Universal promptable image segmentation models like `segment anything' (SAM)~\cite{kirillov2023segment} and others~\cite{zou2023segment,li2023semantic,sam_hq,mobile_sam,zhao2023fast} have recently become available and serve as excellent candidates for the image-level model in a `track anything' pipeline -- Figure~\ref{fig:teaser} shows some promising results of our integration with these methods.

\begin{figure}[t]
    \centering
    \input{figs/fig-data-percentage}
    \vspace{-1em}
    \caption{
    We plot relative $\overline{\vpq}$ increase of our decoupled approach over the end-to-end baseline when we vary the training data in the target domain (VIPSeg~\cite{miao2022large}). Common/rare classes are the top/bottom 50\% most annotated object category in the training set.
    Our improvement is most significant ($>$60\%) in rare classes when there is a small amount of training data.
    This is because our decoupling allows the use of external class-agnostic temporal propagation data -- data that cannot be used by existing end-to-end baselines.  Details in Section~\ref{sec:expr-data}.
    }
    \label{fig:data-percentage}
\end{figure}
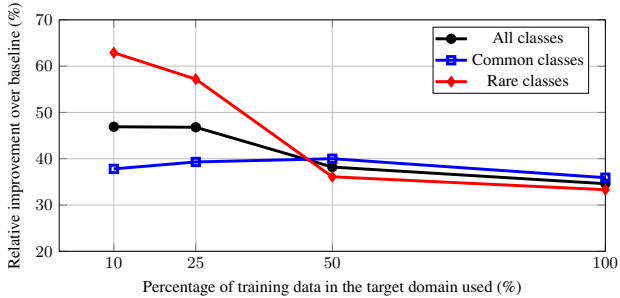

Researchers have studied decoupled formulations before, as `tracking-by-detection'~\cite{kim2015multiple,tang2017multiple,bergmann2019tracking}. However, these approaches often consider image-level detections immutable, while the temporal model only associates detected objects. This formulation depends heavily on the quality of per-image detections and is sensitive to image-level errors.

In contrast, we develop a (semi-)online bi-directional propagation algorithm to 1) denoise image-level segmentation with in-clip consensus (Section~\ref{sec:in-clip-consensus}), and 2) combine results from temporal propagation and in-clip consensus gracefully (Section~\ref{sec:merging-past-future}).
This bi-directional propagation allows temporally more coherent  and potentially better results than those of an image-level model (see Figure~\ref{fig:data-percentage}).

We do not aim to replace end-to-end video approaches. Indeed, we emphasize that specialized frameworks on video tasks with sufficient video-level training data (e.g., YouTubeVIS~\cite{yang2019video}) outperform the developed method.
Instead, we show that our decoupled approach acts as a strong baseline when an image model is available but video data is scarce. 
This is in spirit similar to pretraining of large language models~\cite{radford2018improving}: a \emph{task-agnostic} understanding of natural language is available before being finetuned on specific tasks -- in our case, we learn propagation of segmentations of \emph{class-agnostic} objects in videos via a temporal propagation module and make technical strides in applying this knowledge to specific tasks.
The proposed decoupled approach transfers well to large-scale or open-world datasets, and achieves state-of-the-art results in large-scale video panoptic segmentation~(VIPSeg~\cite{miao2022large}) and open-world video segmentation~(BURST~\cite{athar2023burst}). 
It also performs competitively on referring video segmentation~(Ref-YouTubeVOS~\cite{seo2020urvos}, Ref-DAVIS~\cite{khoreva2019video}) and unsupervised video object segmentation~(DAVIS-16/17\cite{caelles2019}) without end-to-end training.

To summarize:
\begin{itemize}[noitemsep,topsep=0pt]
    \item We propose using decoupled video segmentation that leverages external data, which allows it to generalize better to target tasks with limited annotations than end-to-end video approaches and allows us to seamlessly incorporate existing universal image segmentation models like SAM~\cite{kirillov2023segment}.
    \item We develop bi-directional propagation that denoises image segmentations and merges image segmentations with temporally propagated segmentations gracefully.
    \item We empirically show that our approach achieves favorable results in several important tasks including large-scale video panoptic segmentation, open-world video segmentation, referring video segmentation, and unsupervised video object segmentation.
\end{itemize}

%% file: figs/fig-data-percentage.tex
\resizebox{\columnwidth}{!}{%

\begin{tikzpicture}

\begin{axis}[
    xlabel={Percentage of training data in the target domain used (\%)},
    ylabel={Relative improvement over baseline (\%)},
    xmin=0, xmax=100,
    ymin=20, ymax=70,
    xtick={10,25,50,100},
    ytick={20,30,40,50,60,70},
    legend pos=north west,
    ymajorgrids=true,
    xmajorgrids=true,
    width=12cm,
    height=6cm,
    legend pos=north east,
    label style = {font=\small},
    tick label style = {font=\small} , 
    legend style = {font=\small, inner sep=0.5pt},
    every axis plot/.append style={ultra thick},
]

\addplot[
    color=black,
    mark=*,
    ]
    coordinates {
    (100, 34.6)
    (50, 38.2)
    (25, 46.8)
    (10, 46.9)
    };
    \addlegendentry{All classes}
\addplot[
    color=blue,
    mark=square,
    ]
    coordinates {
    (100, 35.9)
    (50, 40.0)
    (25, 39.3)
    (10, 37.8)
    };
    \addlegendentry{Common classes}
\addplot[
    color=red,
    mark=diamond,
    ]
    coordinates {
    (100, 33.3)
    (50, 36.1)
    (25, 57.2)
    (10, 62.9)
    };
    \addlegendentry{Rare classes}
\end{axis}
\end{tikzpicture}
}

%% file: 02-related_works.tex
\section{Related Works}

\paragraph{End-to-End Video Segmentation.}
Recent end-to-end video segmentation approaches~\cite{qiao2021vip,hwang2021video,wang2021end,bertasius2020classifying,cheng2021,ChoudhuriCVPR2023,ChoudhuriICCV2021} have made significant progress in tasks like Video Instance Segmentation (VIS) and Video Panoptic Segmentation (VPS), especially in closed and small vocabulary datasets like YouTube-VIS~\cite{yang2019video} and Cityscape-VPS~\cite{kim2020video}.
However, these methods require end-to-end training and their scalability to larger vocabularies, where video data and annotations are expensive, is questionable. 
MaskProp~\cite{bertasius2020classifying} uses mask propagation  
to provide temporal information, but still needs to be trained end-to-end on the target task. This is because their mask propagation is not class-agnostic.
We circumvent this training requirement and instead decouple the task into image segmentation and temporal propagation, each of which is easier to train with image-only data and readily available class-agnostic mask propagation data respectively. 

\paragraph{Open-World Video Segmentation.}
Recently, an open-world video segmentation dataset BURST~\cite{athar2023burst} has been proposed. 
It contains 482 object classes in diverse scenarios and evaluates open-world performance by computing metrics for the common classes (78, overlap with COCO~\cite{lin2014microsoft}) and uncommon classes (404) separately. 
The baseline in BURST~\cite{athar2023burst} predicts a set of object proposals using an image instance segmentation model trained on COCO~\cite{lin2014microsoft} and associates the proposals frame-by-frame using either box IoU or STCN~\cite{cheng2021stcn}. OWTB~\cite{liu2022opening} additionally associates proposals using optical flow and pre-trained Re-ID features. 
Differently, we use bi-directional propagation that generates segmentations instead of simply associating existing segmentations -- this reduces sensitivity to image segmentation errors.
UVO~\cite{du2021uvo} is another open-world video segmentation dataset and focuses on human actions. We mainly evaluate on BURST~\cite{athar2023burst} as it is much more diverse and allows separate evaluation for common/uncommon classes.

\paragraph{Decoupled Video Segmentation.}
`Tracking-by-detection' approaches~\cite{kim2015multiple,tang2017multiple,bergmann2019tracking} often consider image-level detections immutable and use a short-term temporal tracking model to associate detected objects. This formulation depends heavily on the quality of per-image detections and is sensitive to image-level errors.
Related long-term temporal propagation works exist~\cite{goel2021msn,garg2021mask}, but they consider 
a single task and do not filter the image-level segmentation. We instead propose a general framework, with a bi-directional propagation mechanism that denoises the image segmentations and allows our result to potentially perform better than the image-level model.

\paragraph{Video Object Segmentation.}
Semi-supervised Video Object Segmentation (VOS) aims to propagate an initial ground-truth segmentation through a video~\cite{perazzi2016benchmark,oh2019videoSTM,yang2021associating,cheng2022xmem}. 
However, it does not account for any errors in the initial segmentation, and cannot incorporate new segmentation given by the image model at later frames.
SAM-PT~\cite{rajivc2023segment} combines point tracking with SAM~\cite{cheng2023segment} to create a video object segmentation pipeline, while our method tracks masks directly.
We find a recent VOS algorithm~\cite{cheng2022xmem} works well for our temporal propagation model. Our proposed bi-directional propagation is essential for bringing image segmentation models and propagation models together as a unified video segmentation framework.

\begin{figure*}[t]
    \centering
    \includegraphics[width=\linewidth]{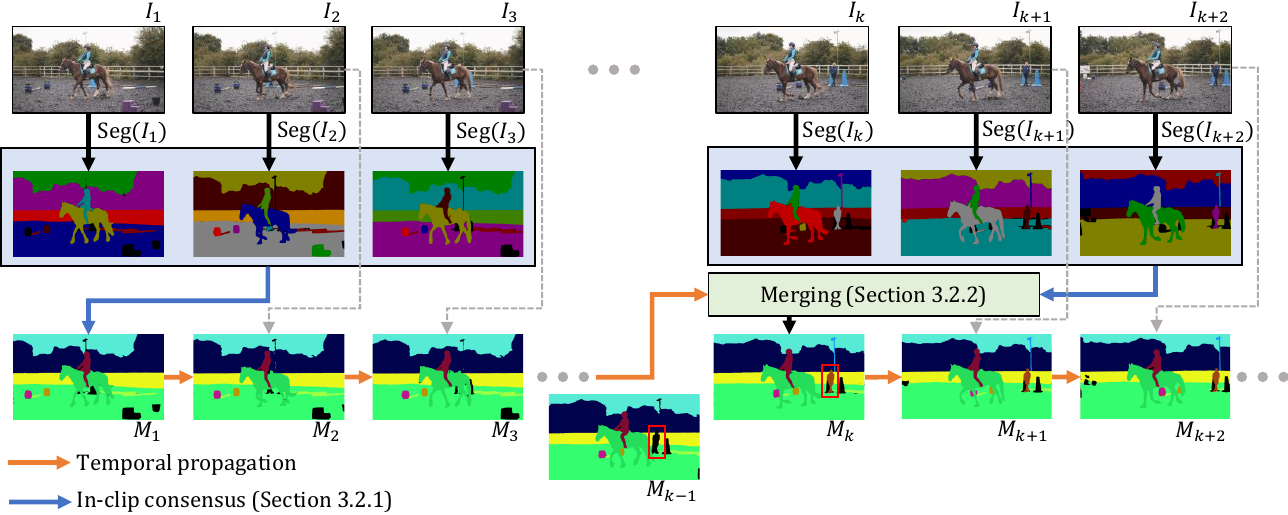}
    \caption{Overview of our framework. We first filter image-level segmentations with in-clip consensus (Section~\ref{sec:in-clip-consensus}) and temporally propagate this result forward. 
    To incorporate a new image segmentation at a later time step (for previously unseen objects, e.g., red box), we merge the propagated results with in-clip consensus as described in Section~\ref{sec:merging-past-future}. 
    Specifics of temporal propagation are in the appendix.
    }
    \label{fig:overview}
\end{figure*}

\paragraph{Unified Video Segmentation.}
Recent Video-K-Net~\cite{li2022video} uses a unified framework for multiple video tasks but requires separate end-to-end training for each task. Unicorn~\cite{yan2022towards},  TarViS~\cite{athar2023tarvis}, and UNINEXT~\cite{yan2023universal} share model parameters for different tasks, and train on all the target tasks end-to-end. 
They report lower tracking accuracy for objects that are not in the target tasks during training compared with class-agnostic VOS approaches, which might be caused by  joint learning with class-specific features.
In contrast, we only train an image segmentation model for the target task, while the temporal propagation model is always fully class-agnostic for generalization across tasks.

\paragraph{Segmenting/Tracking Anything.}
Concurrent to our work, Segment Anything (SAM)~\cite{kirillov2023segment} demonstrates the effectiveness and generalizability of large-scale training for universal image segmentation, serving as an important foundation for open-world segmentation. 
Follow-up works~\cite{yang2023track,cheng2023segment} extend SAM to video data by propagating the masks generated by SAM with video object segmentation algorithms.
However, they rely on single-frame segmentation and lack the denoising capability of our proposed in-clip consensus approach.

%% file: 03-method.tex
\section{Decoupled Video Segmentation}

\subsection{Formulation}
\paragraph{Decoupled Video Segmentation.}
Our decoupled video segmentation approach is driven by an image segmentation model and a universal temporal propagation model. 
The image model, trained specifically on the target task, provides task-specific image-level segmentation hypotheses. The temporal propagation model, trained on class-agnostic mask propagation datasets, associates and propagates these hypotheses to segment the whole video.
This design separates the learning of task-specific segmentation and the learning of general video object segmentation, leading to a robust framework even when data in the target domain is scarce and insufficient for end-to-end learning.

\paragraph{Notation.}
Using $t$ as the time index, we refer to the corresponding frame and its final segmentation as $I_t$ and $\seg_t$ respectively.
In this paper, we represent a segmentation as a set of non-overlapping per-object binary segments, \ie, $\seg_t=\{m_{i}, 0<i\leq\lvert\seg_t\rvert\}$, where $m_i \cap m_j = \emptyset$ if $i\neq j$.

The image segmentation model~$\imseg(I)$ takes an image $I$ as input and outputs a segmentation.
We denote its output segmentation at time $t$ as $\imseg(I_t)=\imseg_t=\{s_{i}, 0<i\leq\lvert\imseg_t\rvert\}$, which is also a set of non-overlapping binary segments.
This segmentation model can be swapped for different target tasks, and users can 
be in the loop to correct the segmentation as we do not limit its internal architecture.

The temporal propagation model~$\prop(\mem, I)$ takes a collection of segmented frames (memory) $\mem$ and a query image $I$ as input and segments the query frame with the objects in the memory. 
For instance, $\prop\left(\{I_1,\seg_1\}, I_2\right)$ propagates the segmentation $\seg_1$ from the first frame $I_1$ to the second frame $I_2$. 
Unless mentioned explicitly, the memory $\mem$ contains all past segmented frames.

\paragraph{Overview.}
Figure~\ref{fig:overview} illustrates the overall pipeline. At a high level, we aim to propagate segmentations discovered by the image segmentation model to the full video with temporal propagation. 
We mainly focus on the (semi-)online setting. 
Starting from the first frame, we use the image segmentation model for initialization. To denoise errors from single-frame segmentation, we look at a small clip of a few frames in the near future (in the online setting, we only look at the current frame) and reach an in-clip consensus (Section~\ref{sec:in-clip-consensus}) as the output segmentation.
Afterward, we use the temporal propagation model to propagate the segmentation to subsequent frames.
We modify an off-the-shelf state-of-the-art video object segmentation XMem~\cite{cheng2022xmem} as our temporal propagation model, with details given in the appendix.
The propagation model itself cannot segment new objects that appear in the scene. Therefore, we periodically incorporate new image segmentation results using the same in-clip consensus as before and merge the consensus with the propagated result (Section~\ref{sec:merging-past-future}). 
This pipeline combines the strong temporal consistency from the propagation model (past) and the new semantics from the image segmentation model (future), hence the name \textit{bi-directional propagation}. Next, we will discuss the bi-directional propagation pipeline in detail.

\subsection{Bi-Directional Propagation}

\begin{figure}[t]
    \centering
    \includegraphics[width=\linewidth]{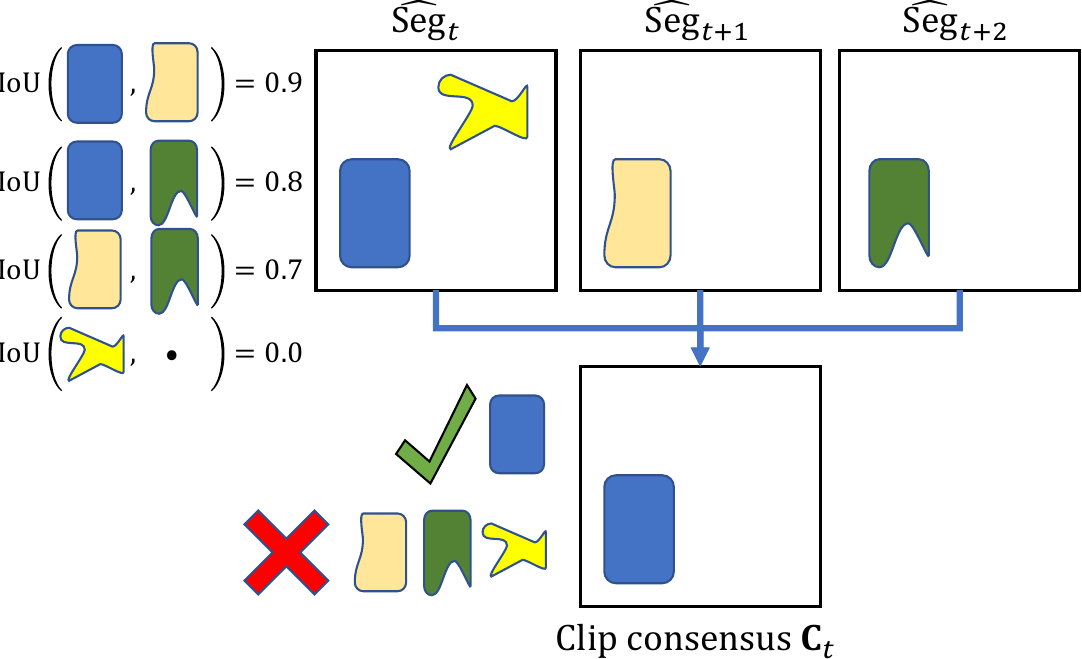}
    \caption{A simple illustration of in-clip consensus.
    The top three squares represent object proposals from three different frames aligned to time $t$.
    The blue shape is the most supported by other object proposals and is selected as output. The yellow shape is not supported by any and is ruled out as noise. The remaining are not used due to significant overlap with the selected (blue) shape.
    }
    \label{fig:consensus}
\end{figure}

\subsubsection{In-clip Consensus}\label{sec:in-clip-consensus}
\paragraph{Formulation.}
In-clip consensus operates on the image segmentations of a small future clip of $n$ frames ($\imseg_t$, $\imseg_{t+1}$, ..., $\imseg_{t+n-1}$) and outputs a denoised consensus $\consen_t$ for the current frame. 
In the online setting, $n=1$ and $\consen_t=\imseg_t$. In the subsequent discussion, we focus on the semi-online setting, as consensus computation in the online setting is straightforward. 
As an overview, we first obtain a set of \textit{object proposals} on the target frame $t$ via spatial alignment, merge the object proposals into a combined representation in a second step, and optimize for an indicator variable to choose a subset of proposals as the output in an integer program. Figure~\ref{fig:consensus} illustrates this in-clip consensus computation in a stylized way and we provide details regarding each of the three aforementioned steps (spatial alignment, representation, and integer programming) next.

\paragraph{Spatial Alignment.}
As the segmentations ($\imseg_t$, $\imseg_{t+1}$, ..., $\imseg_{t+n-1}$) correspond to different time steps, they might be spatially misaligned. This misalignment complicates the computation of correspondences between segments. 
To align segmentations $\imseg_{t+i}$ with frame $t$, techniques like optical flow warping are applicable. In this paper, we simply re-use the temporal propagation model to find the aligned segmentation $\widehat{\imseg}_{t+i}$ (note $\widehat{\imseg}_{t}=\imseg_t$) via
\begin{equation}
    \widehat{\imseg}_{t+i} = \prop\left( \{I_{t+i}, \imseg_{t+i}\}, I_t \right), 0<i<n.
\end{equation}
Note, the propagation model here only uses one frame as memory at a time and this temporary memory $\{I_{t+i}, \imseg_{t+i}\}$ is discarded immediately after alignment. It does not interact with the global memory $\mem$.

\paragraph{Representation.}
Recall that we represent a segmentation as a set of non-overlapping per-object binary segments. 
After aligning all the segmentations to frame $t$, each segment is an \textit{object proposal} for frame $I_t$. 
We refer to the union of all these proposals via $\proposal$ (time index omitted for clarity):
\begin{equation}
    \proposal=\bigcup_{i=0}^{n-1} \widehat{\imseg}_{t+i}=\{ p_i, 0<i\leq \lvert\proposal\rvert \}.
\end{equation}
The output of consensus voting is represented by an indicator variable $v^*\in\{0,1\}^{|\proposal|}$ that combines segments into the consensus output $\consen_t$:
\begin{equation}
    \consen_t = \{ p_i | v^*_i=1\} = \{c_i, 0<i\leq\lvert\consen\rvert\}.\label{eq:consensus_output}
\end{equation}
We resolve overlapping segments $c_i$ in $\consen_t$ by prioritizing smaller segments as they are more vulnerable to being majorly displaced by overlaps.
This priority is implemented by sequentially rendering the segments $c_i$ on an image in descending order of area. 
We optimize for $v$ based on two simple criteria:
\begin{enumerate}[nosep]
    \item Lone proposals $p_i$ are likely to be noise and should not be selected. Selected proposals should be supported by other (unselected) proposals.
    \item Selected proposals should not overlap significantly with each other.
\end{enumerate}
We combine these criteria in an integer programming problem which we describe next.

\paragraph{Integer Programming.}
We aim to optimize the indicator variable $v$ to achieve the above two objectives, by addressing the following integer programming problem:
\begin{align}
v^* = {\arg\!\max}_{v} \sum_i \left( \text{Supp}_i + \text{Penal}_i \right) \ \text{s.t.} \sum_{i,j} \text{Overlap}_{ij} = 0.
\end{align}
Next, we discuss each of the terms in the program in detail.

First, we define the pairwise Intersection-over-Union (IoU) between the $i$-th proposal and the $j$-th proposal as:
\begin{equation}
    \text{IoU}_{ij} = \text{IoU}_{ji} = \frac{\lvert p_i \cap p_j \rvert}{\lvert p_i \cup p_j \rvert}, 0\leq \text{IoU}_{ij} \leq 1.
\end{equation}
The $i$-th proposal \textit{supports} the $j$-th proposal if $\text{IoU}_{ij}>0.5$ -- the higher the IoU, the stronger the support. The more support a segment has, the more favorable it is to be selected. 
To maximize the total support of selected segments, we maximize the below objective for all $i$:
\begin{equation}
\text{Supp}_i = v_i\sum_{j} \begin{cases}
      \text{IoU}_{ij}, & \text{if}\ \text{IoU}_{ij}>0.5  \text{ and } i\neq j\\
      0, & \text{otherwise}
    \end{cases}.
\end{equation}
Additionally, proposals that support each other should not be selected together as they significantly overlap. 
This is achieved by constraining the following term to zero:
\begin{equation}
    \text{Overlap}_{ij} = \begin{cases}
      v_i v_j, & \text{if}\ \text{IoU}_{ij}>0.5 \text{ and } i\neq j \\
      0, & \text{otherwise}
    \end{cases}.
\end{equation}
Lastly, we introduce a penalty for selecting any segment for 1)~tie-breaking when a segment has no support, and 2)~excluding noisy segments, with weight $\alpha$:
\begin{equation}
    \text{Penal}_i = -\alpha v_i.
\end{equation}

We set the tie-breaking weight $\alpha=0.5$.
For all but the first frame, we merge $\consen_t$ with the propagated segmentation $\prop(\mem, I_t)$ into the final output $\seg_t$ as described next.

\subsubsection{Merging Propagation and Consensus}\label{sec:merging-past-future}
\paragraph{Formulation.}
Here, we seek to merge the propagated segmentation $\prop(\mem, I_t)=\propseg_t=\{r_i, 0<i\leq\lvert\propseg\rvert\}$ (from the past) with the consensus $\consen_t=\{c_j, 0<j\leq\lvert\consen\rvert\}$ (from the near future) into a single segmentation $\seg_t$.
We associate segments from these two segmentations and denote the association with an indicator $a_{ij}$ which is 1 if $r_i$ associates with $c_j$, and $0$ otherwise. 
Different from the in-clip consensus, these two segmentations contain fundamentally different information.
Thus, we do not eliminate any segments and instead fuse all pairs of associated segments while letting the unassociated segments pass through to the output. Formally, we obtain the final segmentation via
\begin{equation}
\seg_t = \{ r_i \cup c_j | a_{ij}=1 \} \cup \{ r_i | \forall_j a_{ij}=0 \} \cup \{ c_j | \forall_i a_{ij}=0 \}, 
\label{eq:merge_output}
\end{equation}
where overlapping segments are resolved by prioritizing the smaller segments as discussed in Section~\ref{sec:in-clip-consensus}.

\paragraph{Maximizing Association IoU.}
We find $a_{ij}$ by maximizing the pairwise IoU of all associated pairs, with a minimum association IoU of $0.5$. This is equivalent to a maximum bipartite matching problem, with $r_i$ and $c_j$ as vertices and edge weight $e_{ij}$ given by 
\begin{equation}
    e_{ij} = \begin{cases}
      \text{IoU}(r_i, c_j), & \text{if}\ \text{IoU}(r_i, c_j)>0.5 \\
      -1, & \text{otherwise}
    \end{cases}.
\end{equation}
Requiring any matched pairs from two non-overlapping segmentations to have $\text{IoU}>0.5$ leads to a unique matching, as shown in~\cite{kirillov2019panoptic}. 
Therefore, a greedy solution of setting $a_{ij}=1$ if $e_{ij}>0$ and $0$ otherwise suffices to obtain an optimal result. 

\paragraph{Segment Deletion.}
As an implementation detail, we delete inactive segments from the memory to reduce computational costs. We consider a segment $r_i$ inactive when it fails to associate with any segments $c_j$ from the consensus for consecutive $L$ times.
Such objects might have gone out of view or were a misdetection.
Concretely, we associate a counter $\text{cnt}_i$ with each propagated segment $r_i$, initialized as 0.
When $r_i$ is not associated with any segments $c_j$ from the consensus, i.e., $\forall_{j} a_{ij}=0$, we increment $\text{cnt}_i$ by 1 and reset $\text{cnt}_i$ to 0 otherwise.
When $\text{cnt}_i$ reaches the pre-defined threshold $L$, the segment $r_i$ is deleted from the memory. We set $L=5$ in all our experiments.





%% file: 04-experiments.tex
\section{Experiments}
We first present our main results using a large-scale video panoptic segmentation dataset (VIPSeg~\cite{miao2022large}) and an open-world video segmentation dataset (BRUST~\cite{athar2023burst}). 
Next, we show that our method also works well for referring video object segmentation and unsupervised video object segmentation. 
We present additional results on the smaller-scale YouTubeVIS dataset in the appendix, but unsurprisingly recent end-to-end specialized approaches perform better because a sufficient amount of data is available in this case.
Figure~\ref{fig:teaser} visualizes some results of the integration of our approach with universal image segmentation models like SAM~\cite{kirillov2023segment} or Grounding-Segment-Anything~\cite{liu2023grounding,kirillov2023segment}.
By default, we merge in-clip consensus with temporal propagation every 5 frames with a clip size of $n=3$ in the semi-online setting, and $n=1$ in the online setting.
We evaluate all our results using either official evaluation codebases or official servers.
We use image models trained with standard training data for each task (using open-sourced models whenever available) and a universal temporal propagation module for all tasks unless otherwise specified.

The temporal propagation model is based on XMem~\cite{cheng2022xmem}, and is trained in a class-agnostic fashion with image segmentation datasets~\cite{shi2015hierarchicalECSSD,wang2017DUTS,zeng2019towardsHRSOD,li2020fss,cheng2020cascadepsp} and video object segmentation datasets~\cite{xu2018youtubeVOS, perazzi2016benchmark,qi2022occluded}.
With the long-term memory of XMem~\cite{cheng2022xmem}, our model can handle long videos with ease. We use top-k filtering~\cite{cheng2021mivos} with $k=30$ following~\cite{cheng2022xmem}. 
The performance of our modified propagation model on common video object segmentation benchmarks (DAVIS~\cite{perazzi2016benchmark}, YouTubeVOS~\cite{xu2018youtubeVOS}, and MOSE~\cite{MOSE}) are listed in the appendix.

\subsection{Large-Scale Video Panoptic Segmentation}
We are interested in addressing the large vocabulary setting.
To our best knowledge, VIPSeg~\cite{miao2022large} is currently the largest scale in-the-wild panoptic segmentation dataset, with 58 things classes and 66 stuff classes in 3,536 videos of 232 different scenes.

\paragraph{Metrics.}
To evaluate the quality of the result, we adopt the commonly used VPQ (Video Panoptic Quality)~\cite{kim2020video} and STQ (Segmentation and Tracking Quality)~\cite{weber2021step} metrics.
VPQ extends image-based PQ (Panoptic Quality)~\cite{kirillov2019panoptic} to video data by matching objects in sliding windows of $k$ frames (denoted $\vpq^k$). When $k=1$, VPQ $=$ PQ and associations of segments between frames are ignored. 
Correct long-range associations, which are crucial for object tracking and video editing tasks, are only evaluated with a large value of $k$.
For a more complete evaluation of VPS, we evaluate $k\in\{1,2,4,6,8,10,\infty\}$.
Note, $\vpq^\infty$ considers the entire video as a tube and requires global association.
We additionally report $\overline{\vpq}$, which is the average of $\vpq^\infty$ and the arithmetic mean of $\vpq^{\{1,2,4,6,8,10\}}$. 
This weights $\vpq^\infty$ higher as it  represents video-level performance, while the other metrics only assess frame-level or clip-level results. 
STQ is proposed in STEP~\cite{weber2021step} and is the geometric mean of AQ (Association Quality) and SQ (Segmentation Quality). It evaluates pixel-level associations and semantic segmentation quality respectively. We refer readers to~\cite{kim2020video} and~\cite{weber2021step} for more details on VPQ and STQ.

\begin{table*}[t]
\small
\input{tabs/tab-vipseg}
\caption{Comparisons of end-to-end approaches (e.g., state-of-the-art Video-K-Net~\cite{li2022video}) with our decoupled approach on the large-scale video panoptic segmentation dataset VIPSeg~\cite{miao2022large}. 
Our method scales with better image models and performs especially well with large $k$ where long-term associations are considered.
All baselines are reproduced using official codebases.
}
\label{tab:vipseg}
\end{table*}

\begin{table*}[]
    \input{tabs/tab-burst}
    \caption{Comparison to baselines in the open-world video segmentation dataset BURST~\cite{athar2023burst}. `com' stands for `common classes' and `unc' stands for `uncommon classes'.
    Our method performs better in both -- in the common classes with Mask2Former~\cite{cheng2022masked} image backbone, and in the uncommon classes with EntitySeg~\cite{qi2021open}. 
    The agility to switch image backbones is one of the main advantages of our decoupled formulation.
    Baseline performances are transcribed from~\cite{athar2023burst}. 
    }
    \label{tab:burst-results}
\end{table*}

\begin{figure}
\input{figs/k-trend}
\vspace{-1em}
\caption{Performance trend comparison of Video-K-Net~\cite{li2022video} and our decoupled approach with the same base model. Ours decreases slower with larger $k$, indicating that the proposed decoupled method has a better long-term propagation.}
\label{fig:k-trend}
\end{figure}
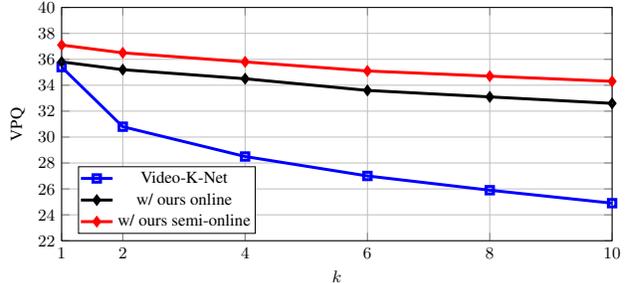

\paragraph{Main Results.}
Table~\ref{tab:vipseg} summarizes our findings.
To assess generality, we study three models as image segmentation input (PanoFCN~\cite{li2021panopticfcn}, Mask2Former~\cite{cheng2022masked}, and Video-K-Net~\cite{li2022video}) to our decoupled approach.
The weights of these image models are initialized by pre-training on the COCO panoptic dataset~\cite{lin2014microsoft} and subsequently fine-tuned on VIPSeg~\cite{miao2022large}.
Our method outperforms both baseline Clip-PanoFCN~\cite{miao2022large} and state-of-the-art Video-K-Net~\cite{li2022video} with the same backbone, especially if $k$ is large, \ie, when long-term associations are more important.
Figure~\ref{fig:k-trend} shows the performance trend with respect to $k$. 
The gains for large values of $k$ highlight the use of a decoupled formulation over end-to-end training: the latter struggles with associations eventually, as training sequences aren't arbitrarily long.
Without any changes to our generalized mask propagation module, using a better image backbone (\eg, SwinB~\cite{liu2021swin}) leads to noticeable improvements. Our method can likely be coupled with future advanced methods in image segmentation for even better performance.

\subsection{Open-World Video Segmentation}\label{sec:expr-open-world}
Open-world video segmentation addresses the difficult problem of discovering, segmenting, and tracking objects in the wild.
BURST~\cite{athar2023burst} is a recently proposed dataset that evaluates open-world video segmentation. It contains diverse scenarios and 2,414 videos in its validation/test sets. There are a total of 482 object categories, 78 of which are `common' classes while the rest are `uncommon'. 

\paragraph{Metrics.}
Following~\cite{athar2023burst}, we assess Open World Tracking Accuracy (OWTA), computed separately for `all', `common', and `uncommon' classes.
False positive tracks are not directly penalized in the metrics as the ground-truth annotations are not exhaustive for all objects in the scene, but indirectly penalized by requiring the output mask to be mutually exclusive.
We refer readers to~\cite{athar2023burst,luiten2021hota} for details.

\paragraph{Main Results.}
Table~\ref{tab:burst-results} summarizes our findings.
We study two image segmentation models: Mask2Former~\cite{cheng2022masked}, and EntitySeg~\cite{qi2021open}, both of which are pretrained on the COCO~\cite{lin2014microsoft} dataset.
The Mask2Former weight is trained for the instance segmentation task, while EntitySeg is trained for `entity segmentation', that is to segment all visual entities without predicting class labels. 
We find EntitySeg works better for novel objects, as it is specifically trained to do so. 
Being able to plug and play the latest development of open-world image segmentation models without any finetuning is one of the major advantages of our formulation.

\begin{figure}[t]
\centering
\input{figs/fig-burst-vis}
\caption{An in-the-wild result in the BURST~\cite{athar2023burst} dataset. Note, we can even track the small skateboarder (pink mask on the road).}
\label{fig:burst-vis}
\end{figure}
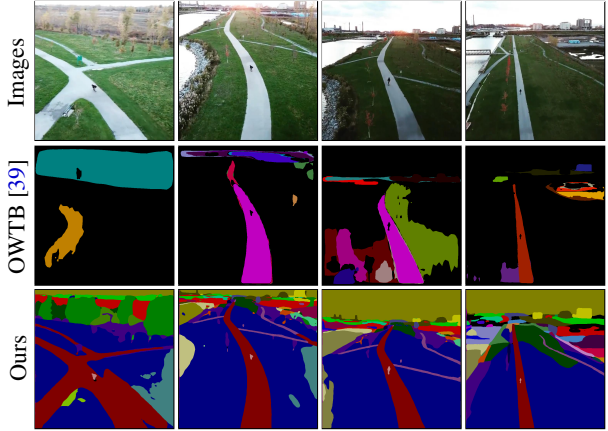

Our approach outperforms the baselines, which all follow the `tracking-by-detection' paradigm. In these baselines, segmentations are detected every frame, and a short-term temporal module is used to associate these segmentations between frames. 
This paradigm is sensitive to misdetections in the image segmentation model.
`Box tracker' uses per-frame object IoU; `STCN tracker' uses a pretrained STCN~\cite{cheng2021stcn} mask propagation network; and OWTB~\cite{liu2022opening} uses a combination of IoU, optical flow, and Re-ID features.
We also make use of mask propagation, but we go beyond the setting of simply associating existing segmentations -- our bi-directional propagation allows us to improve upon the image segmentations and enable long-term tracking.
Figure~\ref{fig:burst-vis} compares our results on one of the videos in BURST to OWTB~\cite{liu2022opening}.

\subsection{Referring Video Segmentation}\label{sec:expr-refer}
Referring video segmentation takes a text description of an object as input and segments the target object. We experiment on Ref-DAVIS17~\cite{khoreva2019video} and Ref-YouTubeVOS~\cite{seo2020urvos} which augments existing video object segmentation datasets~\cite{perazzi2016benchmark,xu2018youtubeVOS} with language expressions. 
Following~\cite{wu2022language}, we assess \mjf~which is the average of Jaccard index (\mj), and boundary F1-score (\mf). 

Table~\ref{tab:refer-vos} tabulates our results.
We use an image-level ReferFormer~\cite{wu2022language} as the image segmentation model. 
We find that the quality of referring segmentation has a high variance across the video (e.g., the target object might be too small at the beginning of the video). As in all competing approaches~\cite{seo2020urvos,wu2022language,ding2022vlt}, we opt for an offline setting to reduce this variance. 
Concretely, we perform the initial in-clip consensus by selecting 10 uniformly spaced frames in the video and using the frame with the highest confidence  given by the image model as a `key frame' for aligning the other frames. We then forward- and backward-propagate from the key frame without incorporating additional image segmentations. We give more details in the appendix.
Our method outperforms other approaches. 


\begin{table}[h]
\small
    \input{tabs/tab-refer-vos}
    \caption{\mjf~comparisons on two referring video segmentation datasets. Ref-YTVOS stands for Ref-YouTubeVOS~\cite{seo2020urvos}.}
    \label{tab:refer-vos}
\end{table}


\subsection{Unsupervised Video Object Segmentation}
Unsupervised video object segmentation aims to find and segment salient target object(s) in a video. We evaluate on DAVIS-16~\cite{perazzi2016benchmark} (single-object) and DAVIS-17~\cite{caelles2019} (multi-object). 
In the single-object setting, we use the image saliency model DIS~\cite{qin2022highly} as the image model and employ an offline setting as in Section~\ref{sec:expr-refer}. 
In the multi-object setting, since the image saliency model only segments one object, we instead use EntitySeg~\cite{qi2021open} and follow our semi-online protocol on open-world video segmentation in Section~\ref{sec:expr-open-world}. 
Table~\ref{tab:unsup-vos} summarizes our findings. Please refer to the appendix for details.

\begin{table}[h]
\small
    \input{tabs/tab-unsup-vos}
    \caption{\mjf~comparisons on three unsupervised video object segmentation datasets: DAVIS16 validation (D16-val), DAVIS17 validation (D17-val), and DAVIS17 test-dev (D17-td).
    Missing entries mean that the method did not report results on that dataset.
    }
    \label{tab:unsup-vos}
\end{table}

\subsection{Ablation Studies}

\subsubsection{Varying Training Data}~\label{sec:expr-data}
Here, we vary the amount of training data in the target domain (VIPSeg~\cite{miao2022large}) to measure the sensitivity of end-to-end approaches \vs our decoupled approach. 
We subsample different percentages of videos from the training set to train Video-K-Net-R50~\cite{li2022video} (all networks are still pretrained with COCO-panoptic~\cite{lin2014microsoft}). We then compare end-to-end performances with our (semi-online) decoupled performances (the temporal propagation model is unchanged as it does not use any data from the target domain). Figure~\ref{fig:teaser} plots our findings -- our model has a much higher relative $\overline{\vpq}$ improvement over the baseline Video-K-Net for rare classes if little training data is available. 

\begin{table}[]
\small
    \input{tabs/tab-consensus-clips-size}
    \caption{Performances of our method on VIPSeg~\cite{miao2022large} with different hyperparameters and design choices.
    By default, we use a clip size of $n=3$ and a merge frequency of every 5 frames with spatial alignment for a balance between performance and speed. }
    \label{tab:consensus-clips-size}
\end{table}

\subsubsection{In-Clip Consensus}
Here we explore hyperparameters and design choices in in-clip consensus. Table~\ref{tab:consensus-clips-size} tabulates our performances with different \emph{clip sizes}, different \emph{frequencies} of merging in-clip consensus with temporal propagation, and whether to use \emph{spatial alignment} during in-clip consensus.
Mask2Former-R50 is used as the backbone in all entries.
For clip size $n=2$, tie-breaking is ambiguous. A large clip is more computationally demanding and potentially leads to inaccurate spatial alignment as the appearance gap between frames in the clip increases.
A high merging frequency reduces the delay between the appearance of a new object and its detection in our framework but requires more computation.
By default, we use a clip size $n=3$, merge consensus with temporal propagation every 5 frames, and enable spatial alignment for a balance between performance and speed. 

\subsubsection{Using Temporal Propagation}
Here, we compare different approaches for using temporal propagation in a decoupled setting.
Tracking-by-detection approaches~\cite{kim2015multiple,tang2017multiple,bergmann2019tracking} typically detect segmentation at every frame and use temporal propagation to associate these per-frame segmentations. 
We test these short-term association approaches using 1) mask IoU between adjacent frames, 2) mask IoU of adjacent frames warped by optical flow from RAFT~\cite{teed2020raft}, and 3) query association~\cite{huang2022minvis} of query-based segmentation~\cite{cheng2022masked} between adjacent frames.
We additionally compare with variants of our temporal propagation method: 4) `ShortTrack', where we consider only short-term tracking by re-initializing the memory $\mem$ every frame, and 5) `TrustImageSeg', where we explicitly trust the consensus given by the image segmentations over temporal propagation by discarding segments that are not associated with a segment in the consensus (i.e., dropping the middle term in Eq.~\eqref{eq:merge_output}).
Table~\ref{tab:temporal-propagation} tabulates our findings. For all entries, we use Mask2Former-R50~\cite{cheng2022masked} in the online setting on VIPSeg~\cite{miao2022large} for fair comparisons.

\begin{table}[h]
\small
    \vspace{-1ex}
    \input{tabs/tab-temporal-propagation}
    \caption{Performances of different temporal schema on VIPSeg~\cite{miao2022large}. Our bi-directional propagation scheme is necessary for the final high performance.}
    \label{tab:temporal-propagation}
    \vspace{-2ex}
\end{table}

\subsection{Limitations}
As the temporal propagation model is task-agnostic, it cannot detect new objects by itself. As shown by the red boxes in Figure~\ref{fig:overview}, the new object in the scene is missing from $\seg_{k-1}$ and can only be detected in $\seg_k$ -- this results in delayed detections relating to the frequency of merging with in-clip consensus.
Secondly, we note that end-to-end approaches still work better when training data is sufficient, i.e., in smaller vocabulary settings like YouTubeVIS~\cite{yang2019video} as shown in the appendix. But we think  decoupled methods are  more promising in large-vocabulary/open-world settings.

%% file: tabs/tab-vipseg.tex
\centering
\begin{tabular}{l@{\hspace{1mm}}l@{\hspace{3mm}}l@{\hspace{3mm}}l@{\hspace{2mm}}c@{\hspace{2mm}}c@{\hspace{2mm}}c@{\hspace{2mm}}c@{\hspace{2mm}}c@{\hspace{2mm}}c@{\hspace{2mm}}c@{\hspace{2mm}}c@{\hspace{2mm}}cc}
\toprule
Backbone & & & & VPQ$^1$ & VPQ$^2$ & VPQ$^4$ & VPQ$^6$ & VPQ$^{8}$ & VPQ$^{10}$ & VPQ$^\infty$ & $\overline{\text{VPQ}}$ & STQ\\
\midrule
Clip-PanoFCN & & end-to-end~\cite{miao2022large} & semi-online & 27.3 & 26.0 & 24.2 & 22.9 & 22.1 & 21.5 & 18.1 & 21.1 & 28.3 \\
Clip-PanoFCN & & decoupled (ours) & online & 29.5 & 28.9 & 28.1 & 27.2 & 26.7 & 26.1 & 25.0 & 26.4 & 35.7 \\
Clip-PanoFCN & & decoupled (ours) & semi-online & \textbf{31.3} & \textbf{30.8} & \textbf{30.1} & \textbf{29.4} & \textbf{28.8} & \textbf{28.3} & \textbf{27.1} & \textbf{28.4} & \textbf{35.8} \\
\midrule
Video-K-Net & R50 & end-to-end~\cite{li2022video} & online & 35.4 & 30.8 & 28.5 & 27.0 & 25.9 & 24.9 & 21.7 & 25.2 & 33.7 \\
Video-K-Net & R50 & decoupled (ours) & online & 35.8 & 35.2 & 34.5 & 33.6 & 33.1 & 32.6 & 30.5 & 32.3 & 38.4 \\
Video-K-Net & R50 & decoupled (ours) & semi-online & 37.1 & 36.5 & 35.8 & 35.1 & 34.7 & 34.3 & 32.3 & 33.9 & 38.6 \\
Mask2Former & R50 & decoupled (ours) & online & 41.0 & 40.2 & 39.3 & 38.4 & 37.9 & 37.3 & 33.8 & 36.4 & 41.1 \\
Mask2Former & R50 & decoupled (ours) & semi-online & \textbf{42.1} & \textbf{41.5} & \textbf{40.8} & \textbf{40.1} & \textbf{39.7} & \textbf{39.3} & \textbf{36.1} & \textbf{38.3} & \textbf{41.5} \\
\midrule
Video-K-Net & Swin-B & end-to-end~\cite{li2022video} & online & 49.8 & 45.2 & 42.4 & 40.5 & 39.1 & 37.9 & 32.6 & 37.5 & 45.2 \\
Video-K-Net & Swin-B & decoupled (ours) & online & 48.2 & 47.4 & 46.5 & 45.6 & 45.1 & 44.5 & 42.0 & 44.1 & 48.6 \\
Video-K-Net & Swin-B & decoupled (ours) & semi-online & 50.0 & 49.3 & 48.5 & 47.7 & 47.3 & 46.8 & 44.5 & 46.4 & 48.9 \\
Mask2Former & Swin-B & decoupled (ours) & online & 55.3 & 54.6 & 53.8 & 52.8 & 52.3 & 51.9 & 49.0 & 51.2 & \textbf{52.4} \\
Mask2Former & Swin-B & decoupled (ours) & semi-online & \textbf{56.0} & \textbf{55.4} & \textbf{54.6} & \textbf{53.9} & \textbf{53.5} & \textbf{53.1} & \textbf{50.0} & \textbf{52.2} & 52.2 \\
\midrule
\bottomrule
\end{tabular}

%% file: tabs/tab-burst.tex
\centering
\begin{tabular}{llcccccc}
\toprule
& & \multicolumn{3}{c}{\textbf{Validation}} & \multicolumn{3}{c}{\textbf{Test}} \\
\cmidrule(lr){3-5} \cmidrule(lr){6-8}
Method & & OWTA$_{\text{all}}$ & OWTA$_{\text{com}}$ & OWTA$_{\text{unc}}$ & OWTA$_{\text{all}}$ & 
OWTA$_{\text{com}}$ & OWTA$_{\text{unc}}$\\
\midrule
Mask2Former & w/ Box tracker~\cite{athar2023burst} & 60.9 & 66.9 & 24.0 & 55.9 & 61.0 & 24.6 \\
Mask2Former & w/ STCN tracker~\cite{athar2023burst} & 64.6 & 71.0 & 25.0 & 57.5 & 62.9 & 23.9 \\
OWTB~\cite{liu2022opening} & & 55.8 & 59.8 & 38.8 & 56.0 & 59.9 & 38.3 \\
Mask2Former & w/ ours online & 69.5 & 74.6 & 42.3 & 70.1 & 75.0 & 44.1 \\
Mask2Former & w/ ours semi-online & \textbf{69.9} & \textbf{75.2} & 41.5 & \textbf{70.5} & \textbf{75.4} & 44.1 \\
EntitySeg & w/ ours online & 68.8 & 72.7 & 49.6 & 69.5 & 72.9 & 53.0 \\
EntitySeg & w/ ours semi-online & 69.5 & 73.3 & \textbf{50.5} & 69.8 & 73.1 & \textbf{53.3} \\
\midrule
\bottomrule
\end{tabular}

%% file: figs/k-trend.tex
\centering
\resizebox{\columnwidth}{!}{%

\begin{tikzpicture}

\begin{axis}[
    xlabel={$k$},
    ylabel={VPQ},
    xmin=1, xmax=10,
    ymin=22, ymax=40,
    xtick={1,2,4,6,8,10},
    ytick={22,24,26,28,30,32,34,36,38,40},
    legend pos=north west,
    ymajorgrids=true,
    xmajorgrids=true,
    width=12cm,
    height=6cm,
    legend pos=south west,
    label style = {font=\small},
    tick label style = {font=\small} , 
    legend style = {font=\small, inner sep=0.5pt},
    every axis plot/.append style={ultra thick},
]

\addplot[
    color=blue,
    mark=square,
    ]
    coordinates {
    (1, 35.4)
    (2, 30.8)
    (4, 28.5)
    (6, 27.0)
    (8, 25.9)
    (10, 24.9)
    };
    \addlegendentry{Video-K-Net}
\addplot[
    color=black,
    mark=diamond,
    ]
    coordinates {
    (1, 35.8)
    (2, 35.2)
    (4, 34.5)
    (6, 33.6)
    (8, 33.1)
    (10, 32.6)
    };
    \addlegendentry{w/ ours online}
\addplot[
    color=red,
    mark=diamond,
    ]
    coordinates {
    (1, 37.1)
    (2, 36.5)
    (4, 35.8)
    (6, 35.1)
    (8, 34.7)
    (10, 34.3)
    };
    \addlegendentry{w/ ours semi-online}
    
\end{axis}
\end{tikzpicture}
}

%% file: figs/fig-burst-vis.tex
\begin{tabular}{c@{}c@{\hspace{-1mm}}c@{\hspace{-1mm}}c@{\hspace{-1mm}}c}

\rotatebox[origin=c]{90}{\small Images \hspace{1mm}}&
\raisebox{-0.5\height}{
    \frame{\includegraphics[trim={9.9cm 0 9.9cm 0},clip,width=0.22\linewidth]{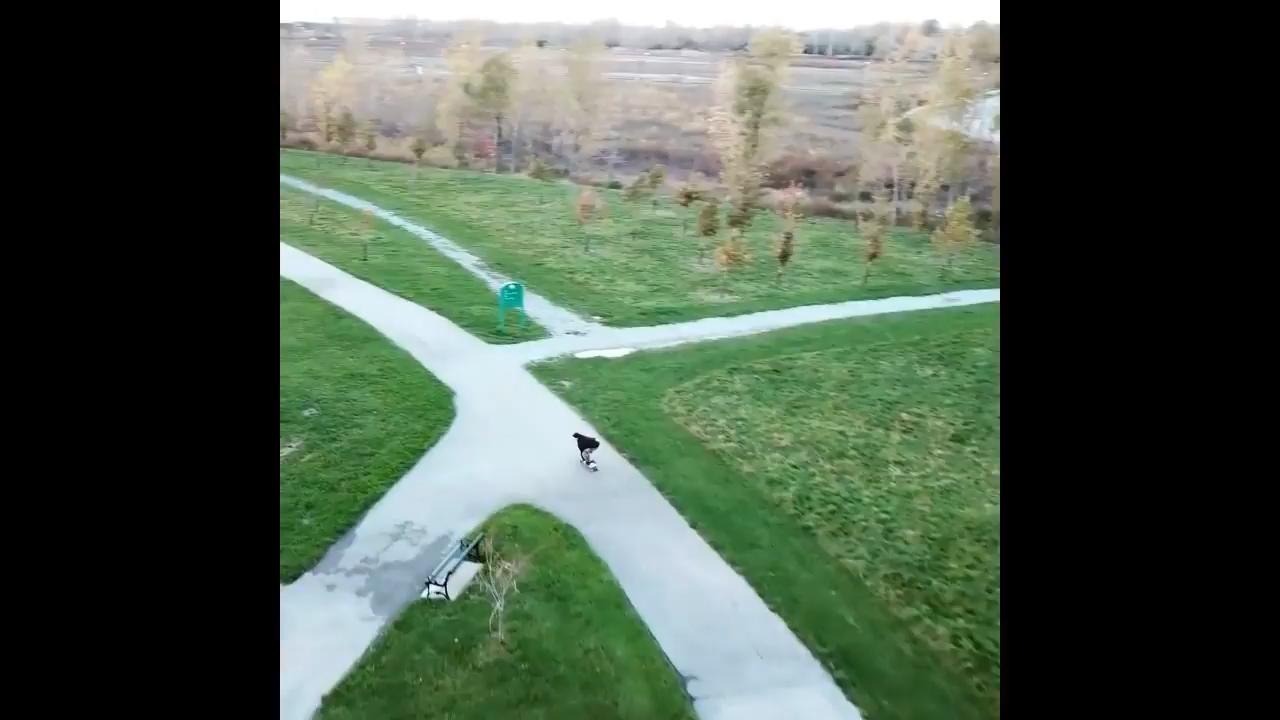}}
}&
\raisebox{-0.5\height}{
    \frame{\includegraphics[trim={9.9cm 0 9.9cm 0},clip,width=0.22\linewidth]{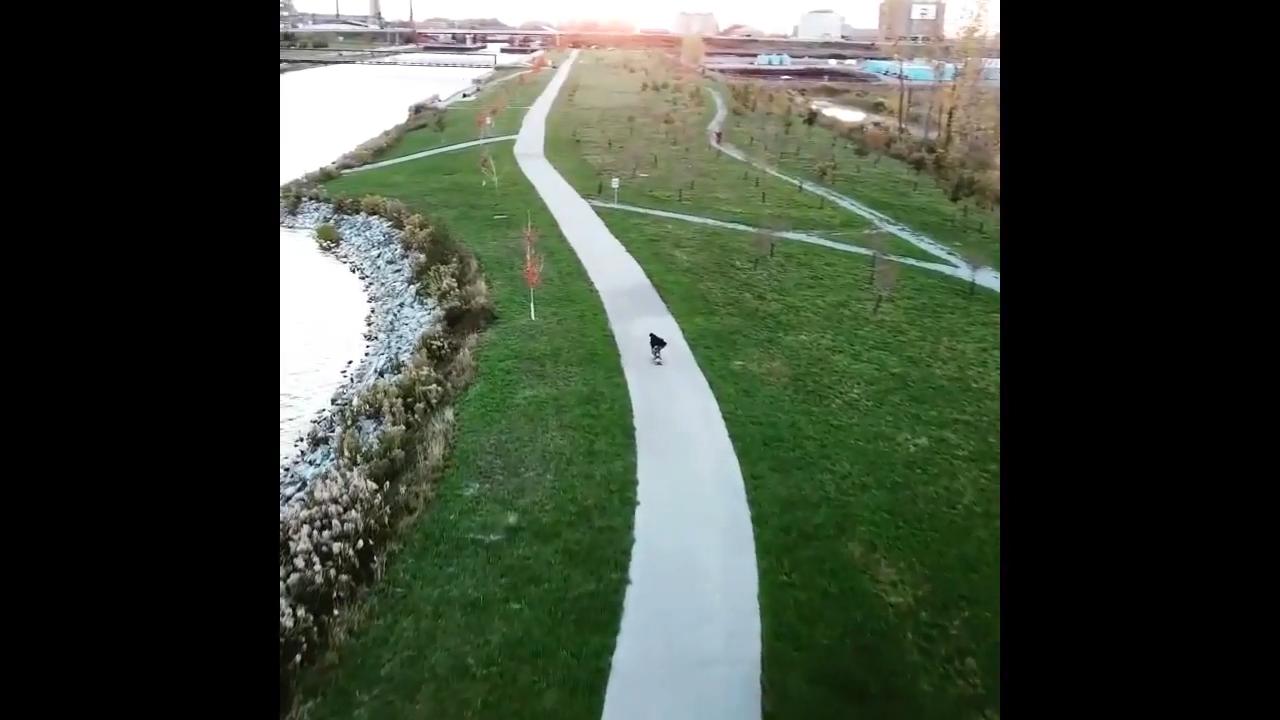}}
}&
\raisebox{-0.5\height}{
    \frame{\includegraphics[trim={9.9cm 0 9.9cm 0},clip,width=0.22\linewidth]{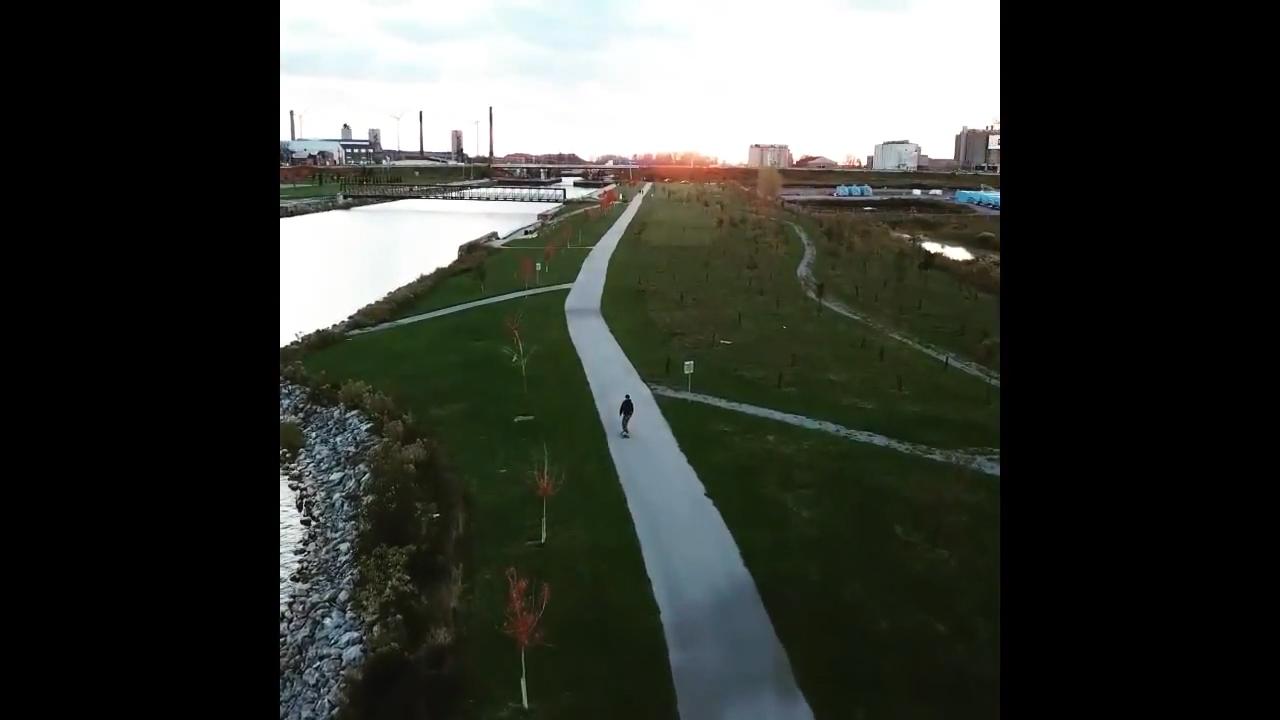}}
}&
\raisebox{-0.5\height}{
    \frame{\includegraphics[trim={9.9cm 0 9.9cm 0},clip,width=0.22\linewidth]{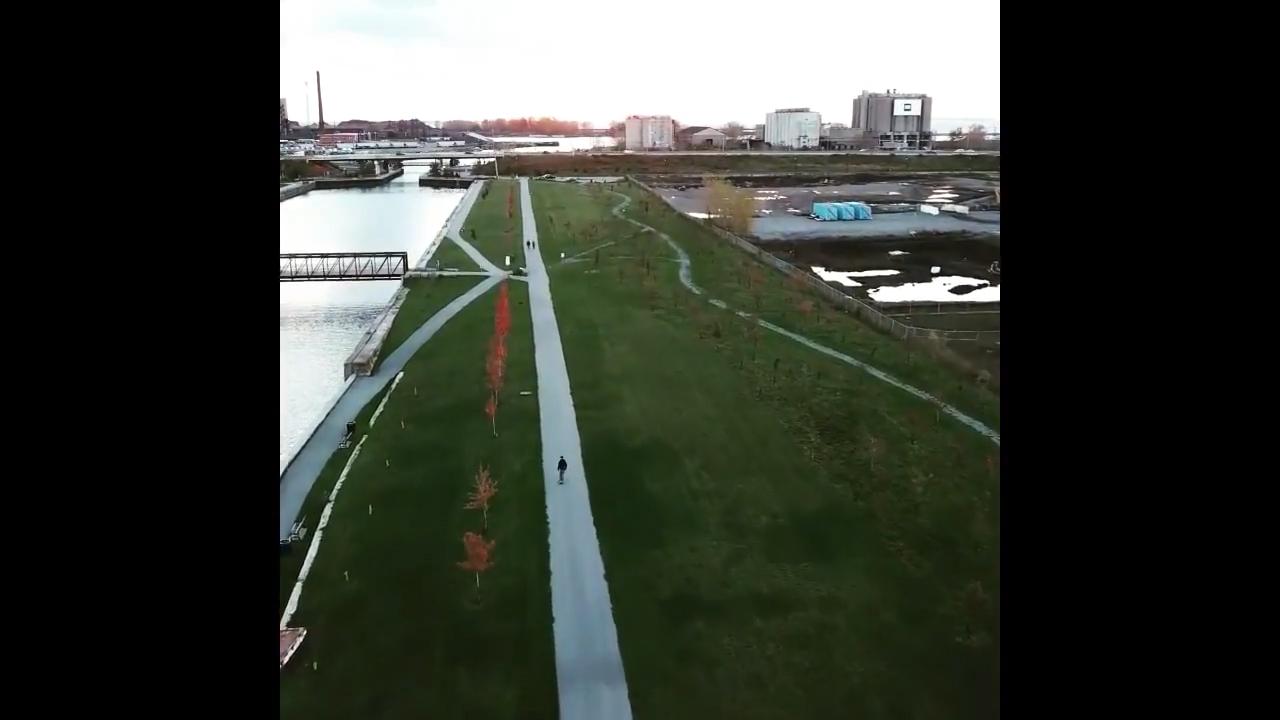}}
}\\
\vspace{-3.5mm}\\
\rotatebox[origin=c]{90}{\small OWTB~\cite{liu2022opening} \hspace{1mm}}&
\raisebox{-0.5\height}{
    \frame{\includegraphics[trim={9.9cm 0 9.9cm 0},clip,width=0.22\linewidth]{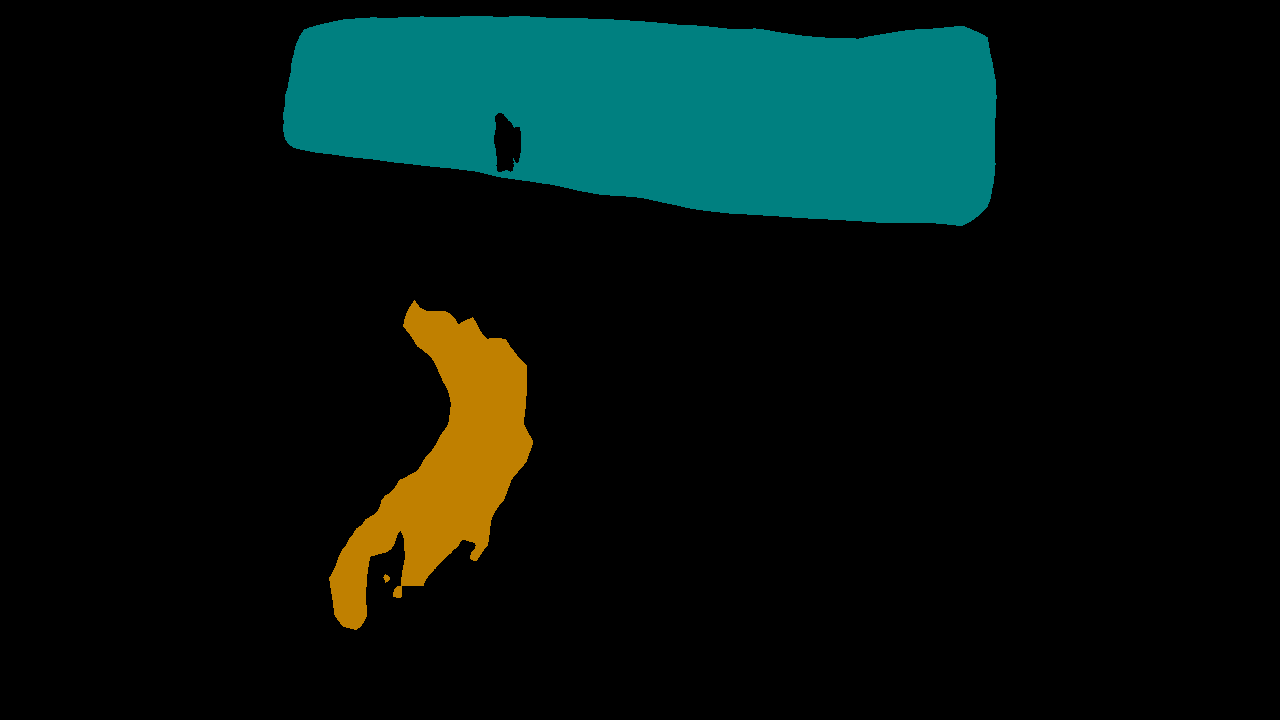}}
}&
\raisebox{-0.5\height}{
    \frame{\includegraphics[trim={9.9cm 0 9.9cm 0},clip,width=0.22\linewidth]{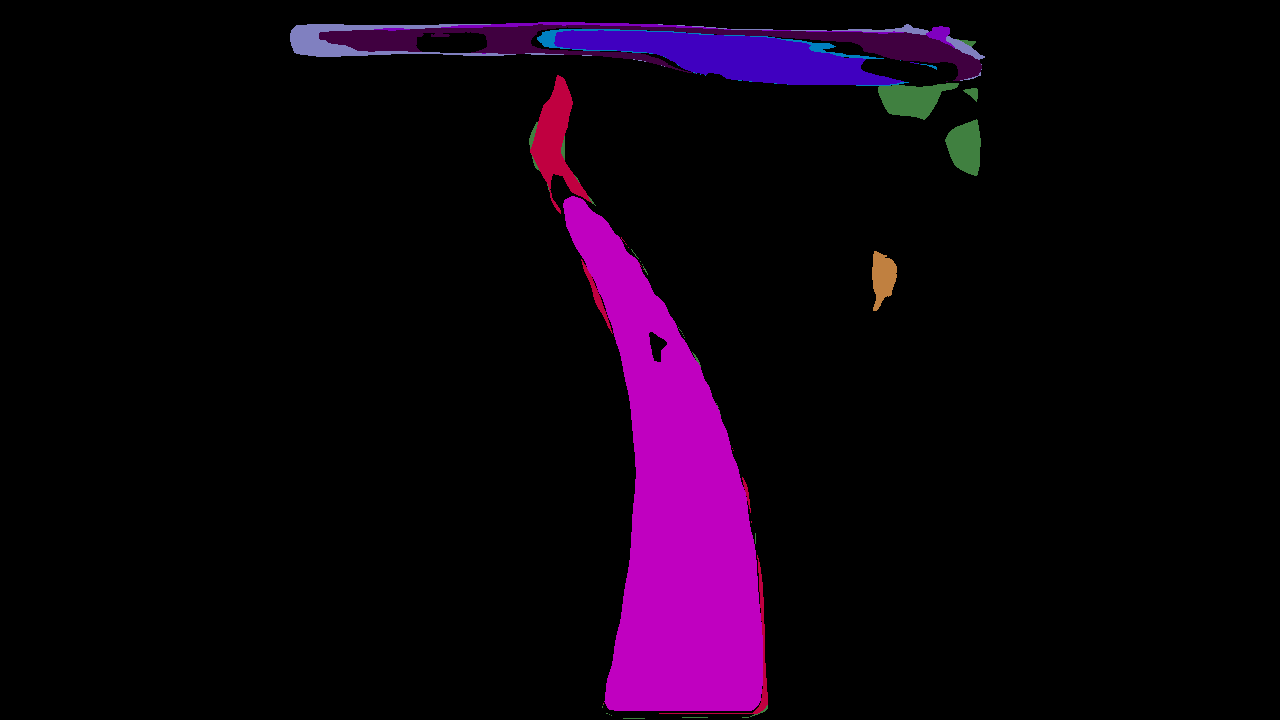}}
}&
\raisebox{-0.5\height}{
    \frame{\includegraphics[trim={9.9cm 0 9.9cm 0},clip,width=0.22\linewidth]{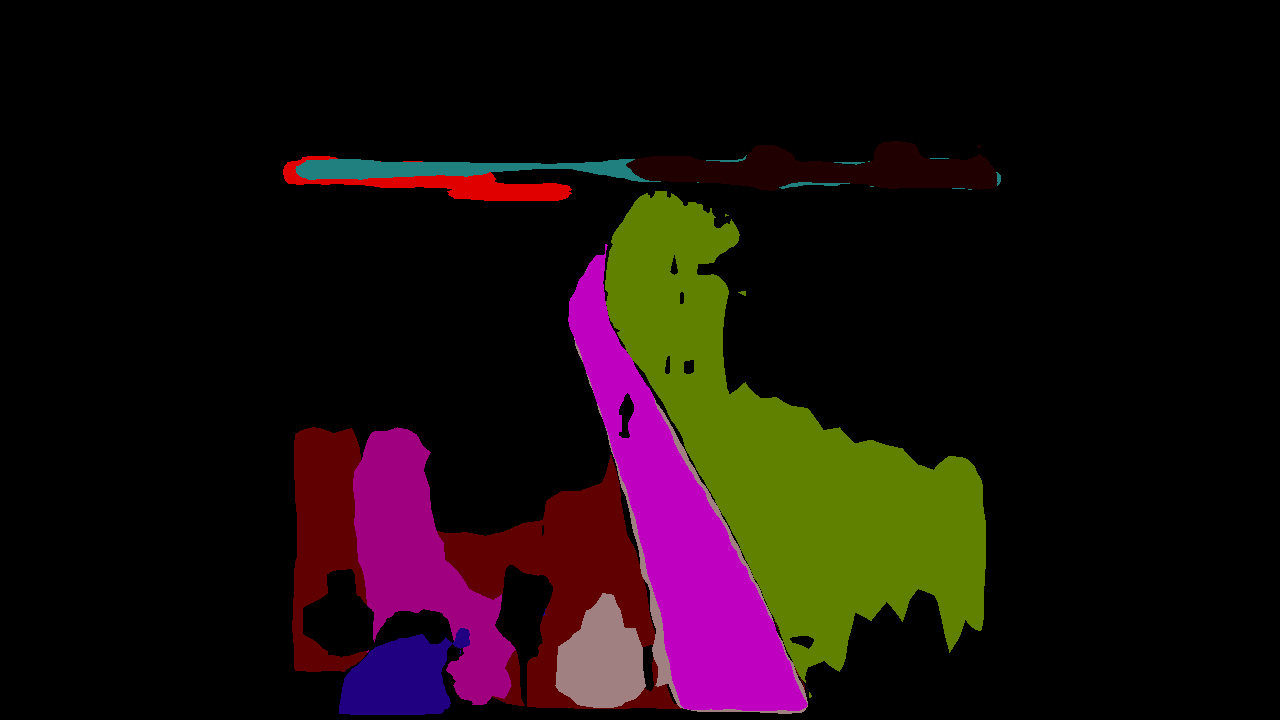}}
}&
\raisebox{-0.5\height}{
    \frame{\includegraphics[trim={9.9cm 0 9.9cm 0},clip,width=0.22\linewidth]{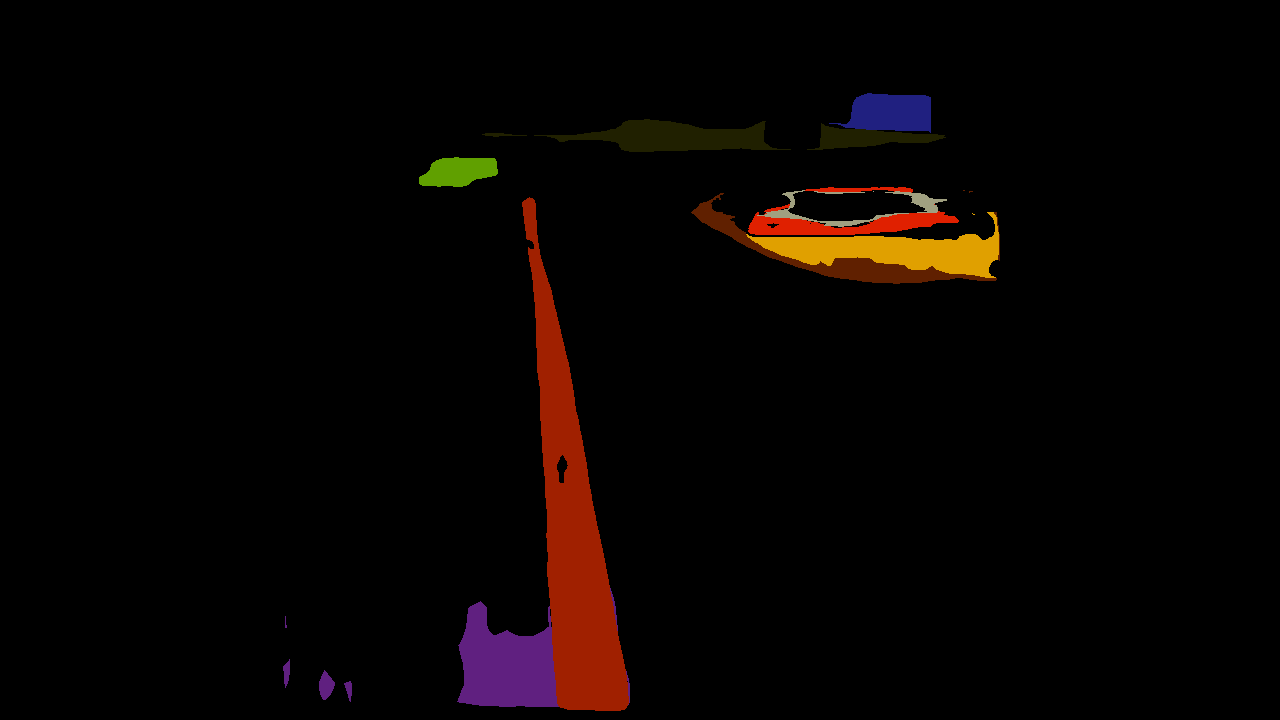}}
}\\
\vspace{-3.5mm}\\
\rotatebox[origin=c]{90}{\small Ours \hspace{1mm}}&
\raisebox{-0.5\height}{
    \frame{\includegraphics[trim={9.9cm 0 9.9cm 0},clip,width=0.22\linewidth]{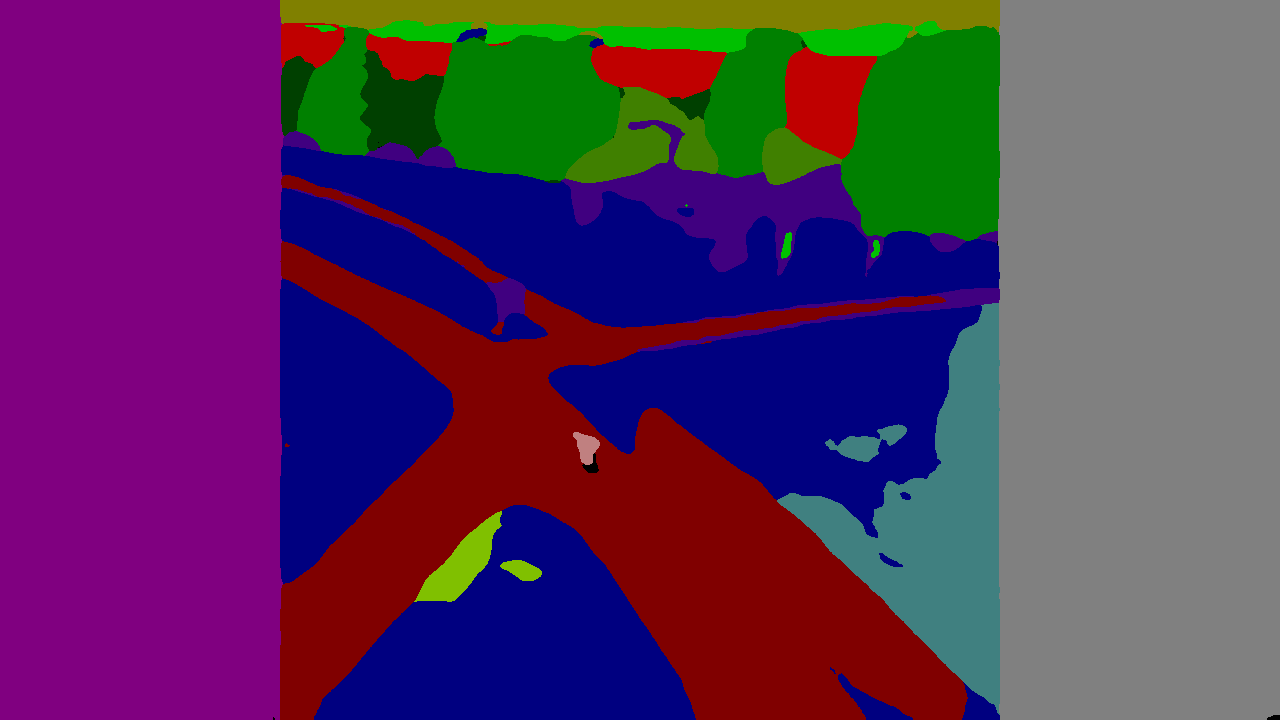}}
}&
\raisebox{-0.5\height}{
    \frame{\includegraphics[trim={9.9cm 0 9.9cm 0},clip,width=0.22\linewidth]{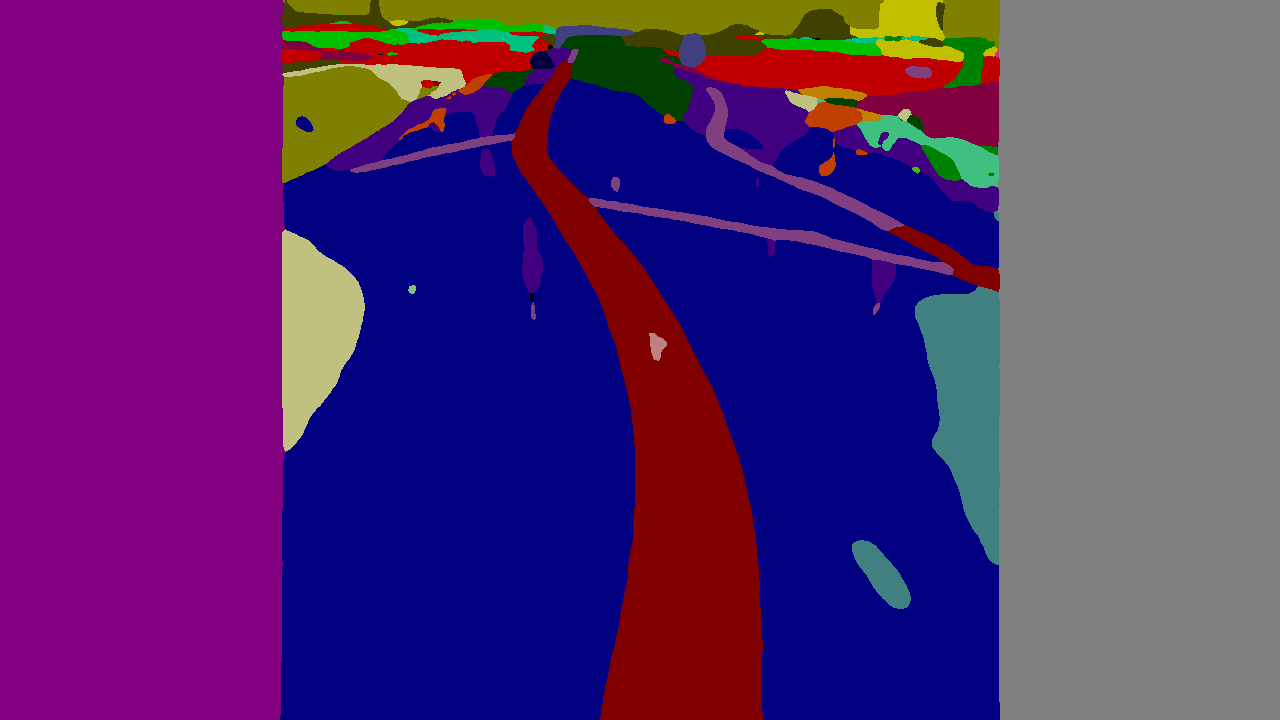}}
}&
\raisebox{-0.5\height}{
    \frame{\includegraphics[trim={9.9cm 0 9.9cm 0},clip,width=0.22\linewidth]{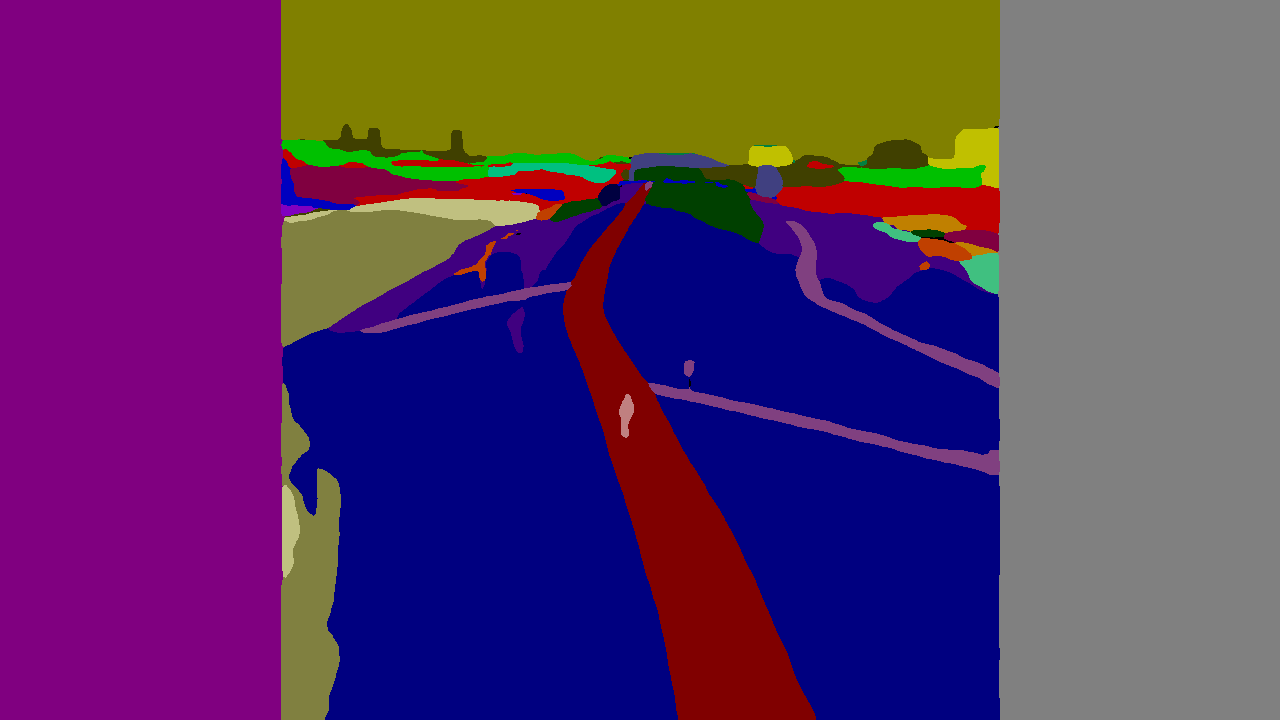}}
}&
\raisebox{-0.5\height}{
    \frame{\includegraphics[trim={9.9cm 0 9.9cm 0},clip,width=0.22\linewidth]{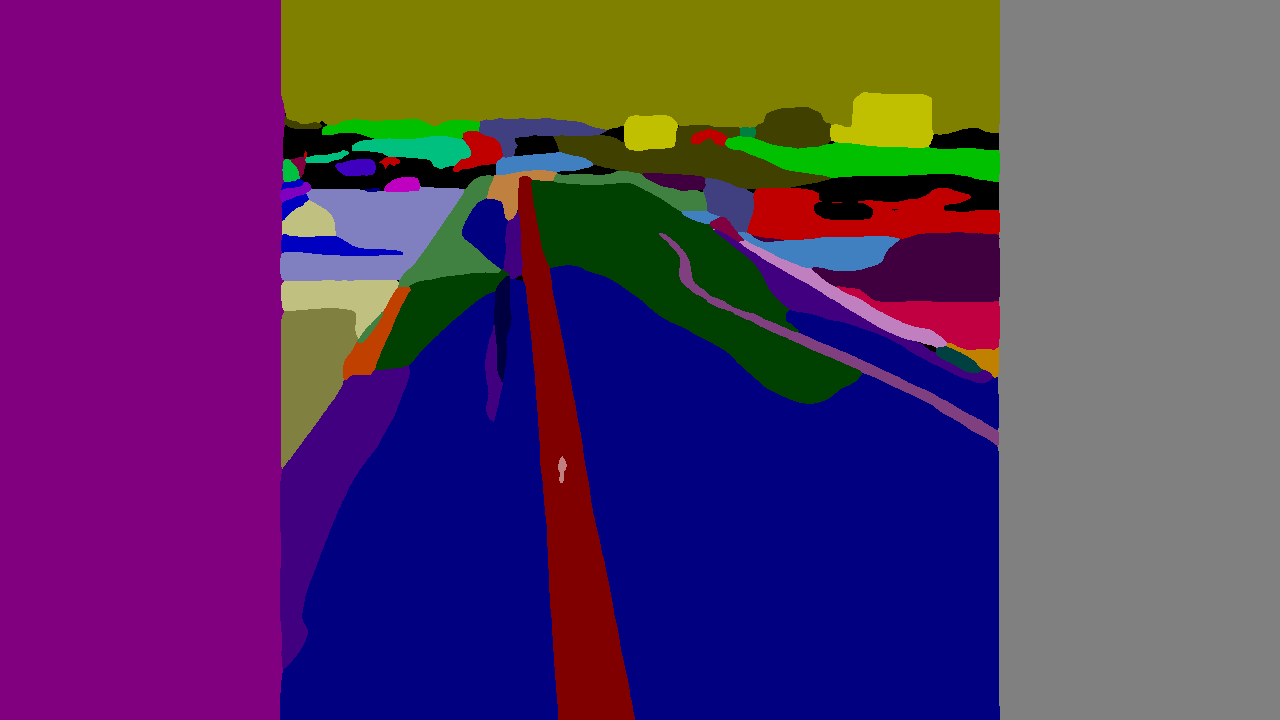}}
}\\
\vspace{-3.5mm}\\

\end{tabular}

%% file: tabs/tab-refer-vos.tex
\centering
\begin{tabular}{lcc}
\toprule
Method & Ref-DAVIS~\cite{khoreva2019video} & Ref-YTVOS~\cite{seo2020urvos} \\
\midrule
URVOS~\cite{seo2020urvos} & 51.6 & 47.2 \\
ReferFormer~\cite{wu2022language} & 60.5 & 62.4 \\
VLT~\cite{ding2022vlt} & 61.6 & 63.8 \\
Ours & \textbf{66.3} & \textbf{66.0} \\
\midrule
\bottomrule
\end{tabular}

%% file: tabs/tab-unsup-vos.tex
\centering
\begin{tabular}{lccc}
\toprule
Method & D16-val & D17-val & D17-td \\
\midrule
RTNet~\cite{ren2021reciprocal} & 85.2 & - & - \\
PMN~\cite{lee2023unsupervised} & 85.9 & - & - \\
UnOVOST~\cite{luiten2020unovost} & - & 67.9 & 58.0 \\
Propose-Reduce~\cite{lin2021video} & - & 70.4 & - \\
Ours & \textbf{88.9} & \textbf{73.4} & \textbf{62.1} \\
\midrule
\bottomrule
\end{tabular}

%% file: tabs/tab-consensus-clips-size.tex
\centering
\begin{tabular}{l@{\hspace{2mm}}c@{\hspace{2mm}}c@{\hspace{3mm}}ccc}
\toprule
\textbf{\textit{Varying clip size}} & VPQ$^1$ & VPQ$^{10}$ & $\overline{\vpq}$ & STQ & FPS\\
\midrule
$n=1$ & 41.0 & 37.3 & 36.4 & 41.1 & \textbf{10.3} \\
$n=2$ & 40.4 & 37.2 & 36.3 & 39.0 & 9.8 \\
$n=3$ & \textbf{42.1} & \textbf{39.3} & 38.3 & 41.5 & 7.8 \\
$n=4$ & \textbf{42.1} & 39.1 & \textbf{38.5} & 42.3 & 6.6 \\
$n=5$ & 41.7 & 38.9 & 38.3 & \textbf{42.8} & 5.6 \\
\midrule
\textbf{\textit{Varying merge freq.}}& VPQ$^1$ & VPQ$^{10}$ & $\overline{\vpq}$ & STQ & FPS \\
\midrule
Every 3 frames & \textbf{42.2} & 39.2 & \textbf{38.4} & \textbf{42.6} & 5.2 \\
Every 5 frames & 42.1 & \textbf{39.3} & 38.3 & 41.5 & 7.8 \\
Every 7 frames & 41.5 & 39.0 & 35.7 & 40.5 & \textbf{8.4} \\
\midrule
\textbf{\textit{Spatial Align?}}& VPQ$^1$ & VPQ$^{10}$ & $\overline{\vpq}$ & STQ & FPS \\
\midrule
Yes & \textbf{42.1} & \textbf{39.3} & \textbf{38.3} & \textbf{41.5} & 7.8 \\
No & 36.7 & 33.9 & 32.8 & 33.7 & \textbf{9.2} \\
\midrule
\bottomrule
\end{tabular}

%% file: tabs/tab-temporal-propagation.tex
\centering
\begin{tabular}{l@{\hspace{3mm}}c@{\hspace{3mm}}c@{\hspace{3mm}}c@{\hspace{3mm}}ccc}
\toprule
Temporal scheme & VPQ$^1$ & VPQ$^4$ & VPQ$^{10}$ & $\overline{\vpq}$ & STQ \\
\midrule
Mask IoU & 39.9 & 32.7 & 27.7 & 27.6 & 34.5 \\
Mask IoU+flow & 40.2 & 33.7 & 28.8 & 28.6 & 37.0 \\
Query assoc. & 40.4 & 33.1 & 28.1 & 28.0 & 35.8 \\
`ShortTrack' & 40.6 & 33.3 & 28.3 & 28.2 & 37.2 \\
`TrustImageSeg' & 40.3 & 37.5 & 33.7 & 33.2 & 37.9 \\
Ours, bi-direction & \textbf{41.0} & \textbf{39.3} & \textbf{37.3} & \textbf{36.4} & \textbf{41.1} \\
\midrule
\bottomrule
\end{tabular}

%% file: 05-conclusion.tex
\section{Conclusion}
We present \textbf{DEVA}, a decoupled video segmentation approach for `tracking anything'. It uses a bi-directional propagation technique that effectively scales image segmentation methods to video data.
Our approach critically leverages external task-agnostic data to reduce reliance on the target task, thus generalizing better to tasks with scarce data than end-to-end approaches. 
Combined with universal image segmentation models, our decoupled paradigm demonstrates state-of-the-art performance as a first step towards open-world large-vocabulary video segmentation.

{
\small
\noindent\textbf{Acknowledgments}. Work supported in part by NSF grants 2008387, 2045586, 2106825, MRI 1725729 (HAL~\cite{kindratenko2020hal}), and NIFA award 2020-67021-32799.
}

%% file: 09-appendix.tex
\appendix
\beginsupplement
\noindent This appendix is structured as follows:
\begin{itemize}
    \item We first provide implementation details of our temporal propagation network (Section~\ref{sec:app:temporal-details}).
    \item We then analyze the class-agnostic training data of the temporal propagation network (Section~\ref{sec:app:temporal-training-data}).
    \item After that, we list additional details regarding our experimental settings and results (Section~\ref{sec:app:experimental-details}).
    \item Next, we provide results on the small-vocabulary YouTube-VIS~\cite{yang2019video} dataset for reference (Section~\ref{sec:app:youtube-vis-results}).
    \item Lastly, we present qualitative results (Section~\ref{sec:app:qualitive-details}).
\end{itemize}

\section{Implementation Details of Temporal Propagation}\label{sec:app:temporal-details}
\subsection{Overview}
Recall that the temporal propagation model~$\prop(\mem, I)$ takes a set of segmented frames (memory) $\mem$ and a query image $I$ as input, and segments the query frame with the objects in the memory. 
For instance, $\prop\left(\{I_1,\seg_1\}, I_2\right)$ propagates the segmentation $\seg_1$ from the first frame $I_1$ to the second frame $I_2$. 
The memory $\mem$ is a compact representation computed from all past segmented frames.

In our implementation, we adopt the design of the internal memory $\mem$ from the recent Video Object Segmentation (VOS) approach XMem~\cite{cheng2022xmem}. 
VOS algorithms are initialized by a first-frame segmentation (in our case, the first in-clip consensus output), and segment new incoming query frames.
XMem is an online algorithm that maintains an internal feature memory representation $\mem$. For each incoming frame $I$, it computes a query representation which is used to read from the feature memory. It then uses the memory readout $\readout$ to segment the query frame. The segmentation result $(\prop(\mem, I))$ is used to update the internal representation $\mem$. 
With an internal memory management mechanism~\cite{cheng2022xmem}, this design has a bounded GPU memory cost with respect to the number of processed frames which is suitable for processing long video sequences.

We refer readers to~\cite{cheng2022xmem} for details regarding XMem. 
We describe core details below for completeness. 
We make a few technical modifications to XMem to increase robustness in our generalized setting, which we also document below. We  provide the full code at {\href{https://hkchengrex.github.io/Tracking-Anything-with-DEVA}{\nolinkurl{hkchengrex.github.io/Tracking-Anything-with-DEVA}}.}

\subsection{Network Architecture}
The temporal propagation network consists of four network modules: a \textit{key encoder}, a \textit{value encoder}, a \textit{mask decoder}, and a \textit{Convolutional Gated Recurrent Unit~(Conv-GRU)}~\cite{chung2014empirical}.

The \textit{key encoder}, implemented with a ResNet-50~\cite{he2016deepResNet}, takes an image as input and produces multi-scale features at the first (stride 4), second (stride 8), and third (stride 16) stages. The fourth stage is discarded.
The feature in the third stage is projected to a `key', which is used for querying during memory reading.
After segmentation, if we decide to add the segmented query frame into the memory $\mem$, we will re-use this `key' in the memory.

The \textit{value encoder}, implemented with a ResNet-18~\cite{he2016deepResNet}, takes an image and a corresponding object mask as inputs and produces a `value' representation as part of the memory. 
We discard the fourth stage and only use the stride-16 output feature in the third stage.
Objects are processed independently (done in mini-batches during inference).

The \textit{mask decoder} takes the memory readout $\readout$ and multi-scale skip connections from the key encoder as inputs and produces an object mask. 
It consists of three upsampling blocks. 
Each upsampling block uses the output from the previous layer as input, upsamples it bilinearly by a factor of two, and fuses the upsampled result with the skip connection at the corresponding scale with a residual block with two $3\times3$ convolutions.
A $3\times3$ convolution is used as the last output layer to generate a single-channel (stride 4) logit and bilinearly upsamples it by four times to the original resolution.
Similar to the value encoder, objects are processed independently, which can be done in mini-batches during inference.
Soft-aggregation~\cite{oh2019videoSTM} is used to combine logits for different objects as in~\cite{cheng2022xmem}.

The \textit{Convolutional Gated Recurrent Unit~(Conv-GRU)}~\cite{chung2014empirical} takes the last hidden state and the output of every upsampling block in the mask decoder as input and produces an updated hidden state.
$3\times3$ convolutions are used as projections in the Conv-GRU.

\paragraph{Our Modifications.}
Firstly, in XMem~\cite{cheng2022xmem}, the 1024\nobreakdash-channel third-stage feature from the key encoder is directly concatenated with the memory readout for mask decoding. For efficiency, we instead first project the 1024\nobreakdash-channel feature to 512 channels with a $1\times1$ convolutional layer before concatenating it with the memory readout. 
Secondly, in each upsampling block of the mask decoder, XMem uses a $3\times3$ convolution to pre-process the skip-connected feature. We replace it with a $1\times1$ convolution.
Moreover, XMem~\cite{cheng2022xmem} and prior works~\cite{oh2019videoSTM,cheng2021stcn} take the image, the target mask, and the sum of all non-target masks (excluding background) as input for the value encoder. We discard the `sum of all non-target masks' as we note that it becomes uninformative when there are many objects in the scene -- typical in open-world scenarios.
We notice a moderate speed-up (22.6$\to$25.8~FPS in DAVIS-2017~\cite{caelles2019}) from these modifications.

\subsection{Feature Memory}
\paragraph{Representation.}
The feature memory consists of three parts: a sensory memory, a working memory, and a long-term memory. 
The sensory memory is represented by the hidden state of the Conv\nobreakdash-GRU and contains positional information for temporal consistency that is updated every frame.
Both the working memory and the long-term memory are attention-based and contain key-value pairs. The working memory is updated every $r$ frames and has a maximum capacity of $T_{\text{max}}$ frames. During each update, the `key' feature from the key encoder and the `value' feature from the value encoder will be appended to the working memory after segmentation of the current frame.
When the working memory reaches its capacity, the oldest $T_{\text{max}}-T_{\text{min}}$  frames will be consolidated into the long-term memory. Please refer to~\cite{cheng2022xmem} for details.

\paragraph{Memory Reading.}
The last hidden state of the Conv\nobreakdash-GRU is used as the memory readout of the sensory memory. For the working and long-term memory, we compute a query from the query frame and perform space-time memory reading~\cite{oh2019videoSTM} to read from both types of memory.
For spatial dimensions $H, W$, the memory readout for the sensory memory is $C_h\times H \times W$ and the memory readout for the working/long-term memory is $C_v\times H \times W$. In XMem, $C_h=64$ and $C_v=512$ and these two features are concatenated together as the final memory readout $\readout$.

\paragraph{Our Modifications.}
For better temporal consistency, we expand the channel size $C_h$ of the sensory memory to $C_h=C_v=512$. 
For efficiency, we use `addition' instead of the original `concatenation' to fuse the memory readout from the sensory memory with the working/long-term memory.
Besides, we supervise the sensory memory with an auxiliary loss -- a $1\times1$ convolution is applied to the sensory memory to produce the weights and biases of a linear classifier on the stride 16 image feature (from the key encoder) for mask prediction. Cross-entropy loss with a weight of $0.1$ is applied on this predicted mask and the network is trained end-to-end.

\subsection{Inference Hyperparameters}
Following~\cite{cheng2022xmem}, the sensory memory is updated every frame. A new memory frame is added to the working memory every $r$-th frame. We synchronize $r$ with our in-clip consensus frequency, such that every in-clip consensus result is added to the working memory. Following the default hyperparameters in~\cite{cheng2022xmem}, we set $T_{\text{max}}=10$, $T_{\text{min}}=5$, the maximum number of long-term memory elements to be $10,000$, and use top-$k$ filtering~\cite{cheng2021mivos} with $k=30$. 

\subsection{Training}\label{sec:app:temporal-training}
XMem is first pretrained on static image segmentation datasets~\cite{shi2015hierarchicalECSSD,wang2017DUTS,zeng2019towardsHRSOD,li2020fss,cheng2020cascadepsp} by synthesizing small video clips of three frames with affine and thin-spline deformations. It is then trained on two video datasets: YouTubeVOS~\cite{xu2018youtubeVOS} and DAVIS~\cite{perazzi2016benchmark} by sampling clips of length eight.

\begin{figure*}
\input{figs/fig-app-data-aug}
\caption{AOT~\cite{yang2021associating} and XMem~\cite{cheng2022xmem} use different crops and rotations within a sequence respectively. We fix both within a sequence to encourage the learning of positional information.}
\label{fig:app-data-aug}
\end{figure*}
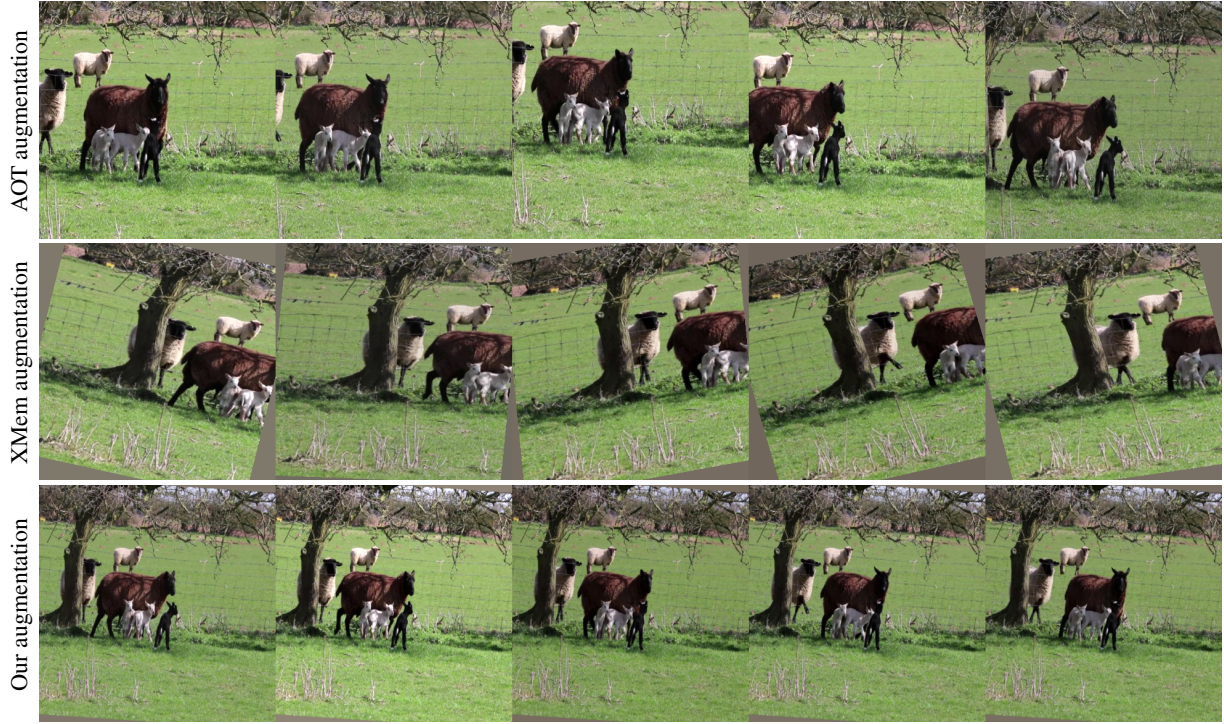

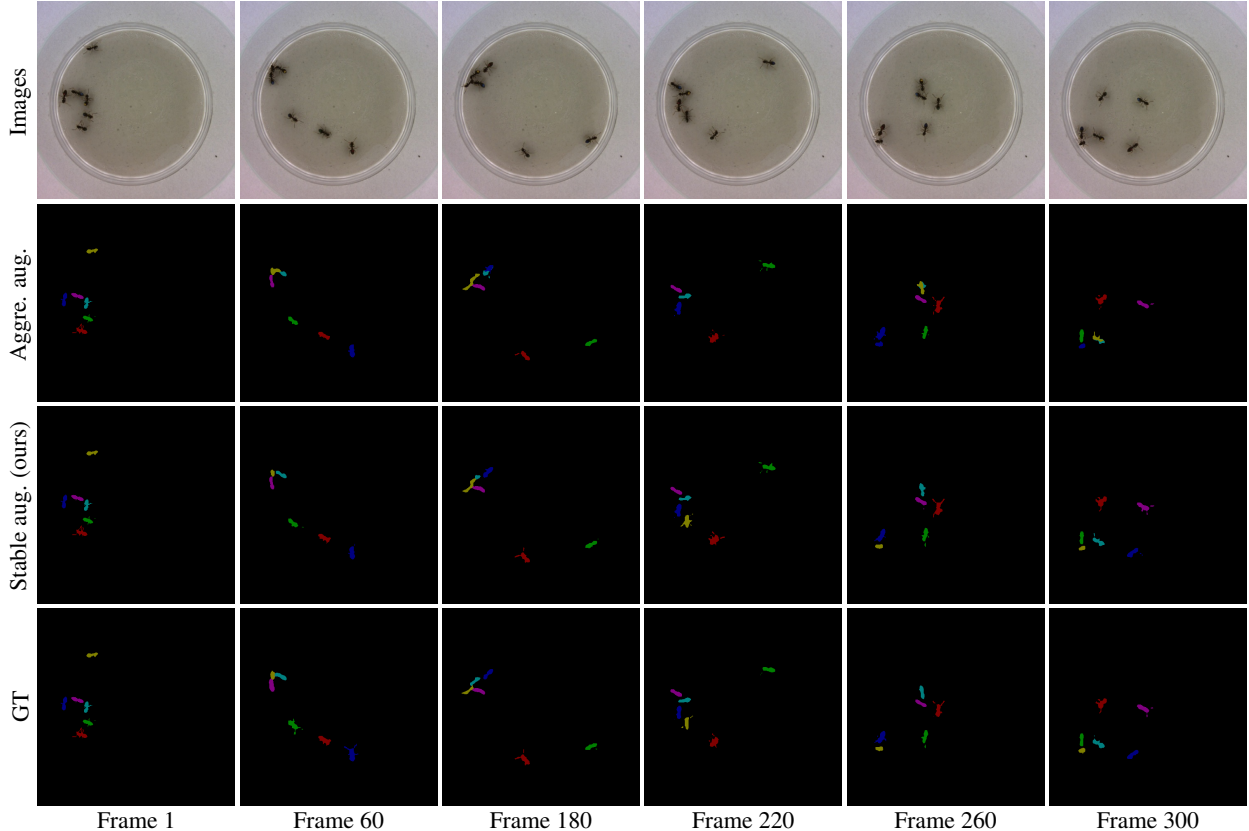
\begin{figure*}
\input{figs/fig-app-ants}
\caption{Comparison of methods tracking a group of ants with almost identical appearance.
The variant with aggressive augmentation fails for the yellow, blue, and cyan ants toward the end while ours with stable data augmentation tracks all ants successfully.
Ground-truth is annotated by us with an interactive image segmentation method, f-BRS~\cite{sofiiuk2020f}.
Zoom in for details.}
\label{fig:app-ants}
\end{figure*}

\paragraph{Our Modifications.}
We make three major modifications to the training process for better robustness:
\begin{enumerate}
    \item We introduce the more challenging OVIS~\cite{qi2022occluded} data into training as we find models have already saturated and produced almost perfect segmentation results on DAVIS and YouTubeVOS during training.
    \item We use a `stable' data augmentation pipeline which leads to better temporal consistency. 
    Current state-of-the-art data augmentation pipelines use aggressive augmentation, applying different rotations~\cite{cheng2022xmem} or crops~\cite{yang2021associating} to frames within the same sequence. 
    This encourages an invariant appearance model but harms the learning of temporal information. We instead use the same rotation and crop augmentation for a video sequence. Figure~\ref{fig:app-data-aug} visualizes the difference.
    \item We clip the norm of the gradient at $3.0$ during training. We find that this leads to faster convergence and more stable training.
\end{enumerate}

We use a batch size of 16, the same loss function (hard-mining cross-entropy loss with a warm-up and soft DICE loss) as XMem~\cite{cheng2022xmem}, and the AdamW~\cite{loshchilov2017decoupledAdamW} optimizer. During pre-training, we use a learning rate of 2$e$-5 for 80,000 iterations. During main training, we use a learning rate of 1$e$-5 for 150,000 iterations with a learning rate decay of $0.1$ at the 120,000-th iteration and the 140,000-th iteration.

\subsection{Video Object Segmentation Evaluation}
We compare our temporal propagation model with state-of-the-art methods on three common video object segmentation benchmarks: DAVIS-2017 validation/test-dev~\cite{perazzi2016benchmark}, YouTubeVOS-2019 validation~\cite{xu2018youtubeVOS}, and MOSE validation~\cite{MOSE}. Table~\ref{tab:app:vos-1} tabulates our results.
We resize all input  such that the shorter side is 480px and bilinearly upsample the output back to the original resolution following XMem~\cite{cheng2022xmem}. All frames in YouTubeVOS~\cite{xu2018youtubeVOS} are used by default.
Our simple modifications bring noticeable improvements on all benchmarks. Though, we find that the overall framework is more important than these design choices (Section~\ref{sec:app:vps-ablation}).

\begin{table*}[t]
    \input{tabs/tab-app-vos-1}
    \caption{Comparison of DEVA's temporal propagation module with state-of-the-art video object segmentation methods. FPS is measured on YouTubeVOS-2019 validation with a V100 GPU. All available frames in YouTubeVOS are used by default.}
    \label{tab:app:vos-1}
\end{table*}

\subsection{Ablation Studies}
\subsubsection{On VOS Tasks}
We assess the effects of our modifications on the training process on VOS tasks which purely evaluate temporal propagation performance.
In addition to the standard DAVIS~\cite{perazzi2016benchmark} dataset, we additionally convert the validation sets of OVIS~\cite{qi2022occluded} and UVO~\cite{wang2021unidentified} to the VOS format.
Following DAVIS~\cite{perazzi2016benchmark}, we discard any segments that do not appear in the first frame and provide the first-frame ground-truth segmentation as input.
These datasets are more diverse and allow for a more complete evaluation of temporal propagation performance.
Table~\ref{tab:app:temporal-prop-ablation-vos} tabulates our findings.
For a fair comparison, we also re-train the original XMem~\cite{cheng2022xmem} with additional OVIS~\cite{qi2022occluded} data.
A qualitative comparison of aggressive \vs stable data augmentation is illustrated in Figure~\ref{fig:app-ants}.

\begin{table}[h]
\small
    \input{tabs/tab-app-temporal-ablation-vos}
    \caption{\mjf~performance comparisons of XMem~\cite{cheng2022xmem} and our different modifications on VOS tasks.}
    \label{tab:app:temporal-prop-ablation-vos}
\end{table}

\begin{table*}[t]
    \input{tabs/tab-app-temporal-ablation-vipseg}
    \caption{Performance comparisons of our method with different temporal propagation model settings on the VIPSeg~\cite{miao2022large} validation set. 
    For a fair comparison, all are semi-online with a Mask2Former-R50~\cite{cheng2022masked} image model input.}
    \label{tab:app:temporal-prop-ablation-vipseg}
\end{table*}

\subsubsection{On Large-Scale Video Panoptic Segmentation}\label{sec:app:vps-ablation}
Next, we assess whether these improvements in VOS tasks transfer to target tasks like video panoptic segmentation.
We compare our final model with/without these modifications on a large-scale video panoptic segmentation dataset VIPSeg~\cite{miao2022large} with the Mask2Former-R50~\cite{cheng2022masked} backbone. 
Table~\ref{tab:app:temporal-prop-ablation-vipseg} (top) tabulates our findings.
Note, our method still works well even without our modifications to the temporal propagation network. 
We find the overall framework to be more important than particular design choices within the temporal propagation model.

\section{Training Data of Temporal Propagation}\label{sec:app:temporal-training-data}

\subsection{Sensitivity to Training Data}
We train the temporal propagation on class-agnostic image segmentation and mask propagation data as described in Section~\ref{sec:app:temporal-training}.
We note that these datasets are cheap to access and amass as they do not require class-specific annotations. 
Here, we evaluate the importance of large-scale training of the temporal propagation model.
We vary the amount of class-agnostic video-level training data under two settings: 1) with full image pretraining, and all three mask propagation datasets (DAVIS~\cite{perazzi2016benchmark}, YouTubeVOS~\cite{xu2018youtubeVOS} and OVIS~\cite{qi2022occluded}), and 2) without image pre-taining and using YouTubeVOS as the only training data.
Table~\ref{tab:app:temporal-prop-ablation-vos} (bottom) tabulates our findings on the VIPSeg~\cite{miao2022large} validation set. 
The performance of our model decays gracefully with fewer training data.

\subsection{Class Overlaps with VIPSeg}
While we train the temporal propagation network in a class-agnostic setting, the segmented objects in the training set might have object categories that overlap with the target task (e.g., with the classes in VIPSeg~\cite{miao2022large}).
Here, we investigate the effect of this overlap of temporal propagation training data with target task data on the final performance.

For this, we train the temporal propagation network with only YouTubeVOS~\cite{xu2018youtubeVOS} data which has 65 object categories (other datasets that we use for training have no class annotation).
We manually match these 65 categories with the classes in VIPSeg~\cite{miao2022large} to partition the classes of VIPSeg into three sets: `overlapping', `non-overlapping', or `ambiguous'.\footnote{The overlapping set includes flag, parasol\_or\_umbrella, car, bus, truck, bicycle, motorcycle, ship\_or\_boat, airplane, person, cat, dog, horse, cattle, skateboard, ball, box, bottle\_or\_cup, table\_or\_desk, mirror, and train (21 in total).
The ambiguous set includes other\_animal, bag\_or\_package, toy, and textiles (4 in total).
The remaining (99) classes in VIPSeg are in the non-overlapping set.}
We then evaluate the final task performance on the overlapping and the non-overlapping sets separately, while ignoring the `ambiguous' set.
We perform the same evaluation on an end-to-end method, Video-K-Net~\cite{li2022video}, as a measure of `baseline difficulty' for each set.
Table~\ref{tab:app:class-overlap} tabulates our findings. 
We observe no significant difference between the overlapping and non-overlapping set when accounting for the difficulty delta ($\Delta$) observed in the baseline.
This indicates that our class-agnostic training does not overfit to the object categories in the training set.

\begin{table}[h]
    \small
    \input{tabs/tab-app-class-overlap}
    \caption{Performance comparison on different classes of VIPSeg that overlap or do not overlap with the training data of temporal propagation.
    As a baseline, we use Video-K-Net-R50~\cite{li2022video}. 
    For ours, we use Mask2Former-R50 with a temporal propagation model that is only trained on YouTubeVOS~\cite{xu2018youtubeVOS} and evaluated in a semi-online setting.}
    \label{tab:app:class-overlap}
\end{table}

\section{Detailed Experimental Settings and Results}\label{sec:app:experimental-details}

\subsection{Large-Scale Video Panoptic Segmentation}\label{sec:app:expr-vipseg}
Following the standard practice~\cite{miao2022large}, we use the 720p version of the VIPSeg~\cite{miao2022large} dataset. 
We evaluate using its validation set (343 videos) and compute VPQ/STQ using the official codebase.
During temporal propagation, we downsample the videos such that the shortest side is 480px and bilinearly upsample the result back to the original resolution following~\cite{cheng2022xmem}.

Video Panoptic Segmentation (VPS) requires the prediction of class labels. 
We obtain these labels from the image segmentation model and use online majority voting to determine the output label.
Formally, we keep a list of class labels $\textbf{Cl}_i$ for each object $r_i$. 
When an existing (propagated) segment $r_i$ matches with a segment from the in-clip consensus $c_j$, i.e., $a_{ij}=1$, we take the class label from the consensus $c_j$ and append it to the list $\textbf{Cl}_i$. 
At the output of every frame, we determine the class label associated with segment $r_i$ by performing majority voting in $\textbf{Cl}_i$.
Note, in accordance with VPS evaluation~\cite{kim2020video}, an object can only have one class label throughout the video.
This means a change in class label necessitates a change in object id, which we also implemented.
Thus, a change in class label might lead to lower association accuracy.
An alternative algorithm would be to use the final major voting result to retroactively apply the class label in all frames, which would however not be strictly online/semi-online.

\paragraph{Running time Analysis}
Under our default semi-online setting, we use a clip size of 3 and perform merging every 5 frames (i.e., invoking the image model on 60\% of all frames).
We report time on VIPSeg~\cite{miao2022large}, averaged across all frames, on an A6000 GPU.
The mask propagation module takes 83ms per frame (VIPSeg has more objects per video than the standard VOS timing benchmark DAVIS-2017). 
For every merge, pre-processing (spatial alignment and finding pairwise IoU) takes 211ms, and solving the integer program takes 15ms.
For the image model (R50 backbone), both Video-K-Net~\cite{li2022video} and Mask2Former~\cite{cheng2022masked} take around 200ms per frame.
Overall, our method runs at 4.0fps.
Meanwhile, state-of-the-art Video-K-Net runs at 4.9fps. Ours is 18\% slower but has a 52\% higher $\overline{\text{VPQ}}$.

\subsection{Open-World Video Segmentation}
We evaluate on the validation (993 videos) and test (1421 videos) sets of BURST~\cite{athar2023burst}. 
As in Section~\ref{sec:app:expr-vipseg}, we downsample the videos during temporal propagation such that the shortest side is 480px and bilinearly upsample the result back to the original resolution following~\cite{cheng2022xmem}. For efficiency, we process only every three frames. Since the ground-truth is sparse (annotated every 24 or 30 frames), we can still perform a complete evaluation.

For the Mask2Former~\cite{cheng2022masked} image model, we follow BURST~\cite{athar2023burst} and use the best-performing Swin-L checkpoint trained on COCO~\cite{lin2014microsoft} provided by the authors. 
For the EntitySeg~\cite{qi2021open} image model, we also use the best available Swin-L model checkpoint trained on COCO~\cite{lin2014microsoft}.
For overlapping predictions, we use the post-processing for panoptic segmentation in Mask2Former~\cite{cheng2022masked} to resolve them.

We assess Open World Tracking Accuracy (OWTA) using official tools. OWTA is the geometric mean of Detection Recall (DetRe) and Association Accuracy (AssA). Please refer to~\cite{athar2023burst} for details. For completeness, we additionally report DetRe and AssA of baselines and our method in Table~\ref{tab:app:burst-deta-assa}.

\begin{table*}
\small
    \input{tabs/tab-app-burst-deta-assa}
    \caption{Extended results comparing baselines and our methods in the validation/test sets of BURST~\cite{athar2023burst}. Baseline performances are transcribed from~\cite{athar2023burst}.}
    \label{tab:app:burst-deta-assa}
\end{table*}

\subsection{Referring Video Segmentation}\label{sec:app:referring-details}
To evaluate on Ref-DAVIS~\cite{khoreva2019video} and Ref-YouTubeVOS~\cite{seo2020urvos}, we use ReferFormer Swin-L~\cite{wu2022language} as the image model. The network is first pretrained on Ref-COCO~\cite{yu2016modeling}, Ref-COCO+~\cite{yu2016modeling}, and G-Ref~\cite{mao2016generation} datasets and finetuned on Ref-YouTubeVOS~\cite{seo2020urvos} following~\cite{wu2022language}.
Unlike in video panoptic segmentation or open-world video segmentation, we do not need to use integer programming to associate segments from the image model in different frames.
This is because each segment corresponds to a known language expression. 
Thus, we process each object independently and use $\arg\!\max$ to fuse the final segmentations.
As mentioned in the main paper, we employ an offline setting as in prior works~\cite{seo2020urvos,wu2022language,ding2022vlt}.

In the offline setting, we first perform in-clip consensus by selecting 10 uniformly spaced frames in
the video and using the frame with the highest confidence given by the image model as a ‘key frame’ for aligning the other frames. 
Soft probability maps are used in the consensus to preserve confidence levels in the prediction.
We then forward- and backward-propagate from the key frame without incorporating additional image segmentations.

Formally, for a given object, we denote its soft probability map and confidence score given by the image model as $\mathbf{Pr}_t\in[0, 1]^{H\times W}$ and $c_t\in[0, 1]$ respectively.
We denote the frame index of the ten chosen frames as $\mathbf{T}_c=\{t_1, t_2, ..., t_{10}\}$.

We aim to compute a soft probability consensus $\mathbf{Cs}_{t_k}$ at a keyframe index $t_k$ by a weighted summation of the soft probability maps of the chosen frames $\{ \mathbf{Pr}_{t_1}, \mathbf{Pr}_{t_2}, ..., \mathbf{Pr}_{t_{10}} \}$:
\begin{equation}
    \mathbf{Cs}_{t_k} = \sum_{i \in \mathbf{T}_c} w_i \mathbf{Pr}_i, 
\end{equation}
where $w_i$ is a weighting coefficient, with $\sum_i w_i=1$.

We use the frame with the highest confidence predicted by the image model as the keyframe:
\begin{equation}
    t_k = {\arg\!\max}_{i \in \mathbf{T}_c} c_i.
\end{equation}

We compute the weighting coefficients using a softmax of the confidences such that we weigh confident predictions more:
\begin{equation}
    w_i = \frac{e^{c_i}}{\sum_{i \in \mathbf{T}_c} e^{c_i}}.
\end{equation}

After the consensus, $\mathbf{Cs}_{t_k}$ is used to initialize forward and backward propagation from frame $t_k$ without incorporating additional image segmentations.
The propagation is implemented as standard semi-supervised video object segmentation inference with the keyframe as initial guidance.
During propagation, the internal memory $\mem$ is updated every 5 frames using its own prediction as in~\cite{cheng2022xmem}.

\subsection{Unsupervised Video Object Segmentation}
For single-object unsupervised video object segmentation (DAVIS-2016~\cite{perazzi2016benchmark}), we use DIS~\cite{qin2022highly} as the image segmentation model. 
Since it does not provide segmentation confidence, we approximate it with the normalized area of the predicted mask to ignore null detections, i.e., $c_i=\frac{1}{HW} \lVert \mathbf{Pr}_{i} \rVert_1$.

For multi-object unsupervised video object segmentation (DAVIS-2017~\cite{caelles2019}), we follow our semi-online protocol in open-world video segmentation. 
The exception being DAVIS-2017~\cite{caelles2019} allows a maximum of 20 objects in the prediction. 
We overcome this limitation online by only accepting the first 20 objects and discarding the rest. 
When there are more than 20 objects in the frame, we prioritize the ones with larger areas as they are less likely to be noisy.

\section{Results on YouTube-VIS}\label{sec:app:youtube-vis-results}
Here we present additional results on the small-vocabulary YouTube-VIS~\cite{yang2019video} dataset, but unsurprisingly recent end-to-end specialized approaches perform better because a sufficient amount of data is available in this case.
For this task, we use our online video panoptic segmentation setting.
Besides the difference in the scale of vocabularies, our method assumes that no two objects occupy the same pixel, and produces a non-overlapping mask. 
Although this assumption is usually true, it harms the Average Precision (AP) evaluation of our method in VIS, with other methods typically outputting many ($\geq$100) potentially overlapping proposals for higher recall. We provide our result in Table~\ref{tab:app:youtubevis}.

\begin{table}[h]
    \centering
    \begin{tabular}{lcc}
    \toprule
    Method & mAP & AP@75\\
    \midrule
    MaskProp~\cite{bertasius2020classifying} & 40.0 & 42.9\\
    Video-K-Net~\cite{li2022video} & 40.5 & 44.5 \\
    MinVIS~\cite{huang2022minvis} & \textbf{47.4} & \textbf{52.1} \\
    Mask2Former~\cite{cheng2022masked} w/ Ours & 40.8 & 44.3 \\
    \midrule
    \bottomrule
    \end{tabular}
    \caption{Performance comparisons on YouTube-VIS 2019 validation. 
    All models use a ResNet-50 backbone.
    Note MinVIS is optimized for small-vocabulary YouTube-VIS and underperforms by 8.4 $\overline{\text{VPQ}}$ compared with our method in large-vocabulary VIPSeg (Tab.\ 6 of the main paper, Query assoc.\ \vs Ours).}
    \label{tab:app:youtubevis}
    \end{table}

\section{Qualitative Results}\label{sec:app:qualitive-details}
\subsection{Visualization}
For all results (see our project page), we associate each object id with a unique color. 
When a segment changes color, its object id has changed. 
This change might happen often (e.g., flicker) if the method is not stable.
We additionally show an `overlay' which is a composite of the colored segmentation with the input image.

\subsection{Large-Scale Video Panoptic Segmentation}
We compare our method with state-of-the-art Video-K-Net~\cite{li2022video}.
We use the semi-online setting Mask2Former~\cite{cheng2022masked} as the image model.
Videos are taken from VIPSeg~\cite{miao2022large} validation set.

\subsection{Open-World Video Segmentation}
We compare our method with the best open-world segmentation baseline (Mask2Former + STCN tracker).
We use the semi-online setting EntitySeg~\cite{qi2021open} as the image model.
Videos are collected from BURST~\cite{athar2023burst} and the Internet.

\subsection{Referring Video Segmentation in the Wild}
We compare our method with state-of-the-art referring video segmentation ReferFormer~\cite{wu2022language}.
We are interested in the open-world setting beyond standard Ref-DAVIS~\cite{khoreva2019video} and Ref-YouTubeVOS~\cite{seo2020urvos}. 
For this, we use a recent open-world referring image segmentation model X-Decoder~\cite{zou2022generalized} as our image model. 
The agility to switch image backbones and use the latest advancements in image segmentation is one of the main advantages of our decoupled formulation. 
We employ an offline setting following our referring video segmentation evaluation protocol (Section~\ref{sec:app:referring-details}). 
Note, ReferFormer~\cite{wu2022language} is also offline.
Our model can segment rare objects like `wheel of fortune' accurately.


%% file: figs/fig-app-data-aug.tex
\centering
\begin{tabular}{c@{}c}

\rotatebox[origin=c]{90}{\small AOT augmentation \hspace{1mm}}&
\raisebox{-0.5\height}{
    \includegraphics[width=0.9\linewidth]{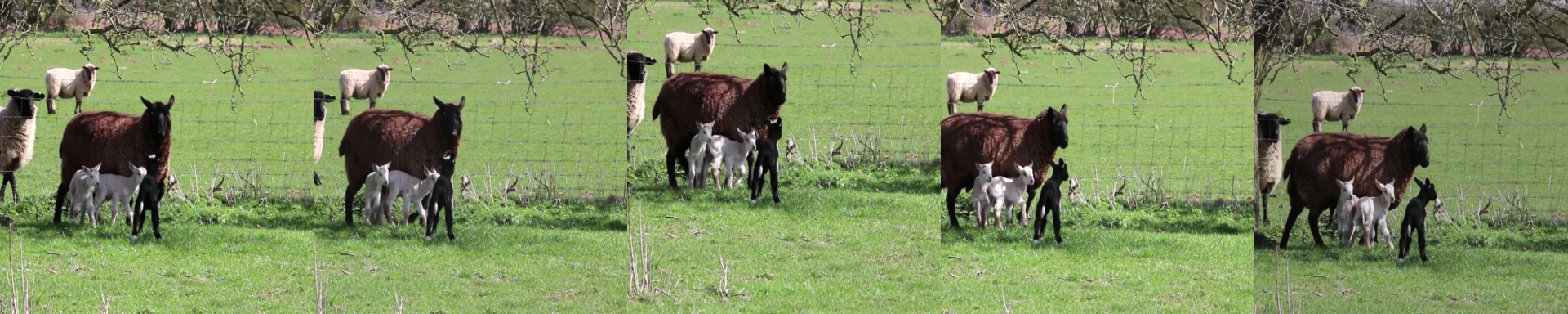}
}\\
\vspace{-3.5mm}\\
\rotatebox[origin=c]{90}{\small XMem augmentation \hspace{1mm}}&
\raisebox{-0.5\height}{
    \includegraphics[width=0.9\linewidth]{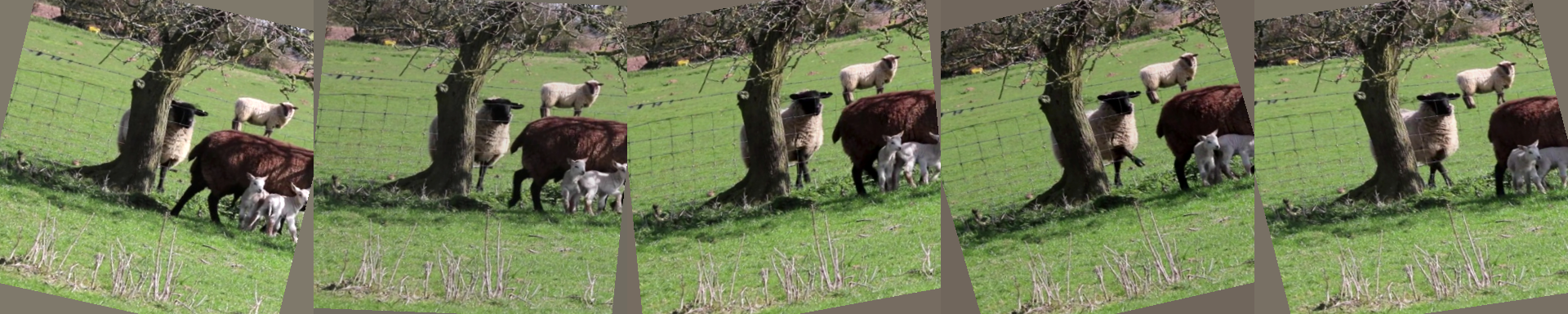}
}\\
\vspace{-3.5mm}\\
\rotatebox[origin=c]{90}{\small Our augmentation \hspace{1mm}}&
\raisebox{-0.5\height}{
    \includegraphics[width=0.9\linewidth]{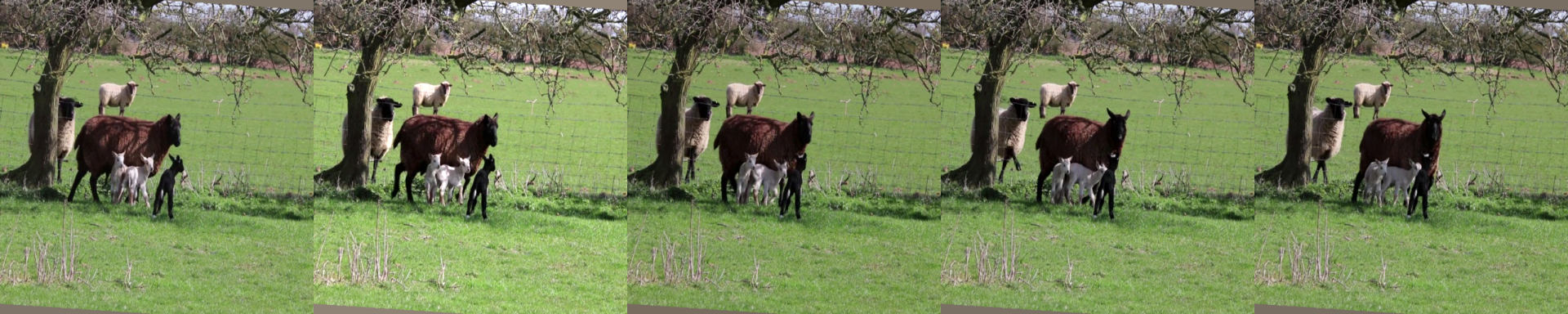}
}\\

\end{tabular}

%% file: figs/fig-app-ants.tex
\centering
\begin{tabular}{c@{\hspace{0cm}}c@{\hspace{-3pt}}c@{\hspace{-3pt}}c@{\hspace{-3pt}}c@{\hspace{-3pt}}c@{\hspace{-3pt}}c}

\rotatebox[origin=c]{90}{\small Images \hspace{1mm}}&
\raisebox{-0.5\height}{
    \includegraphics[width=0.15\linewidth]{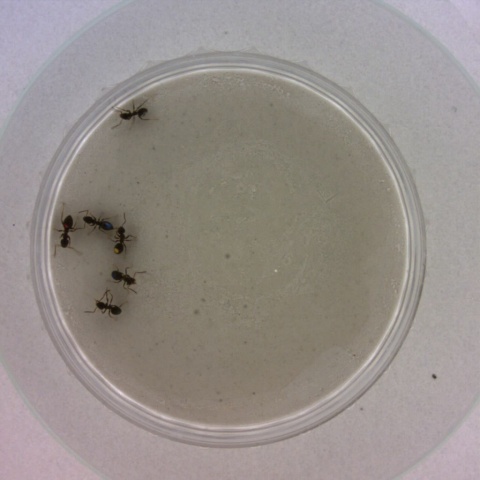}
}&
\raisebox{-0.5\height}{
    \includegraphics[width=0.15\linewidth]{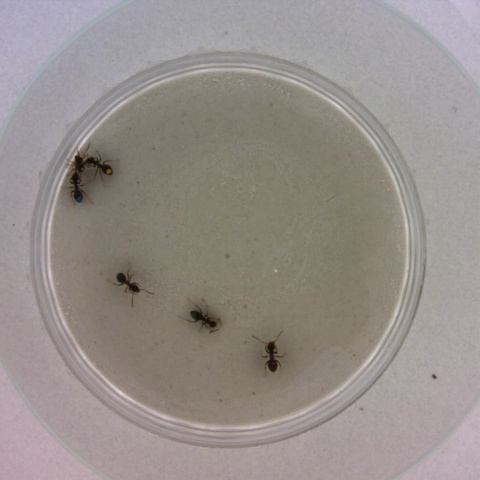}
}&
\raisebox{-0.5\height}{
    \includegraphics[width=0.15\linewidth]{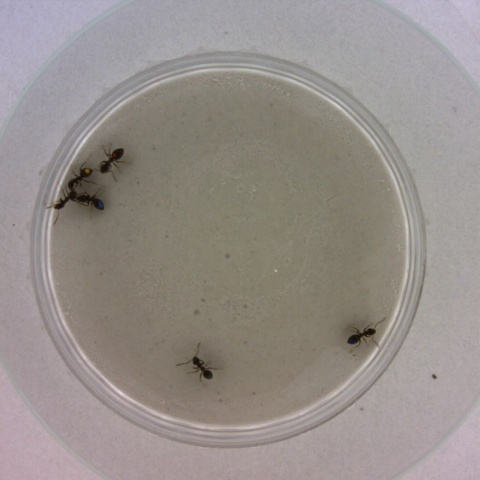}
}&
\raisebox{-0.5\height}{
    \includegraphics[width=0.15\linewidth]{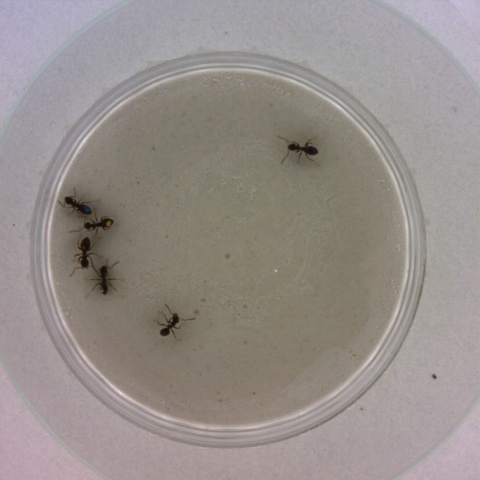}
}&
\raisebox{-0.5\height}{
    \includegraphics[width=0.15\linewidth]{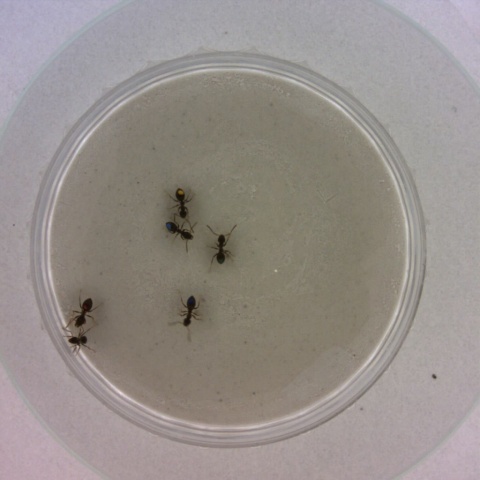}
}&
\raisebox{-0.5\height}{
    \includegraphics[width=0.15\linewidth]{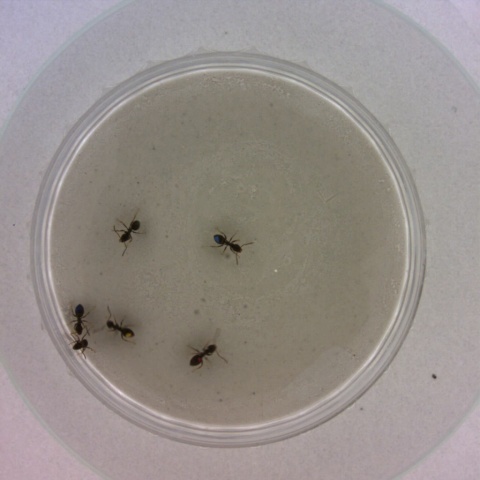}
}\\
\vspace{-3.5mm}\\


\rotatebox[origin=c]{90}{\small Aggre. aug. \hspace{1mm}}&
\raisebox{-0.5\height}{
    \includegraphics[width=0.15\linewidth]{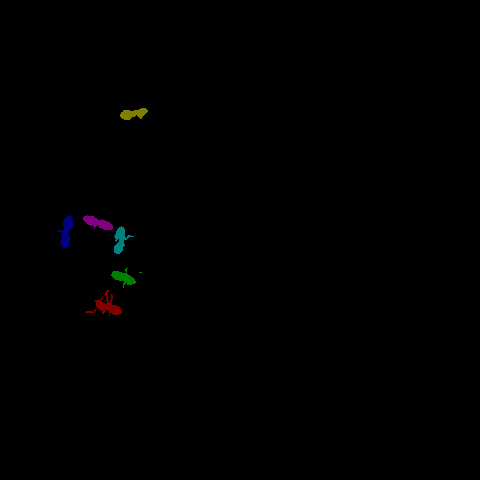}
}&
\raisebox{-0.5\height}{
    \includegraphics[width=0.15\linewidth]{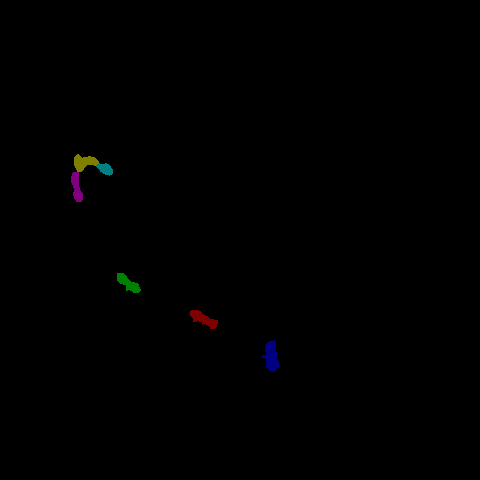}
}&
\raisebox{-0.5\height}{
    \includegraphics[width=0.15\linewidth]{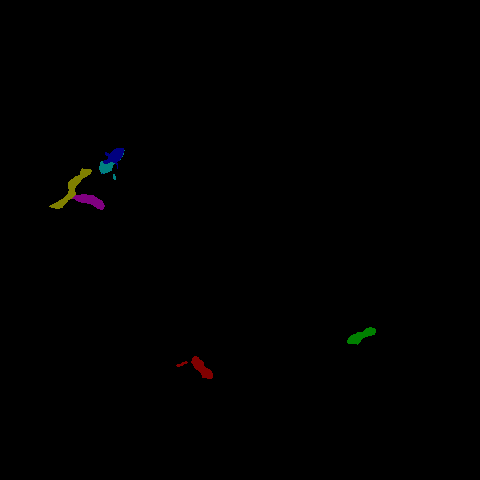}
}&
\raisebox{-0.5\height}{
    \includegraphics[width=0.15\linewidth]{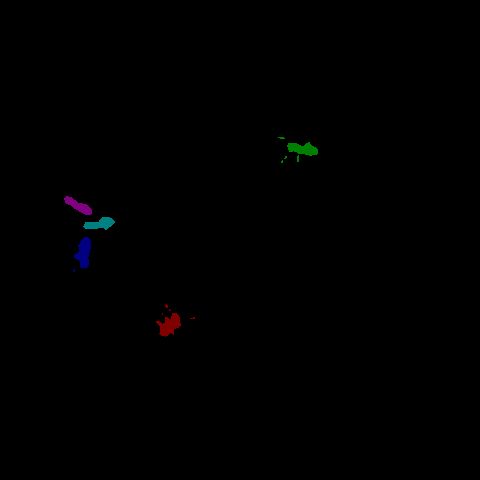}
}&
\raisebox{-0.5\height}{
    \includegraphics[width=0.15\linewidth]{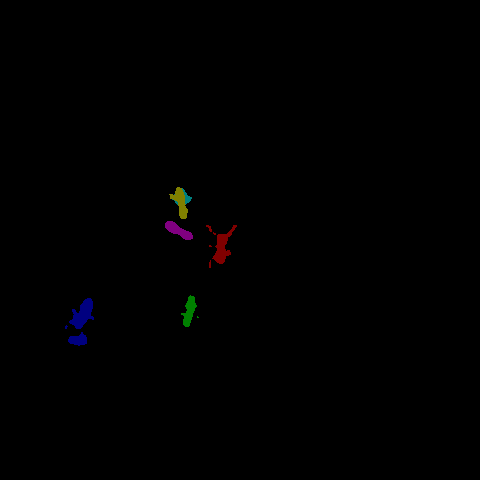}
}&
\raisebox{-0.5\height}{
    \includegraphics[width=0.15\linewidth]{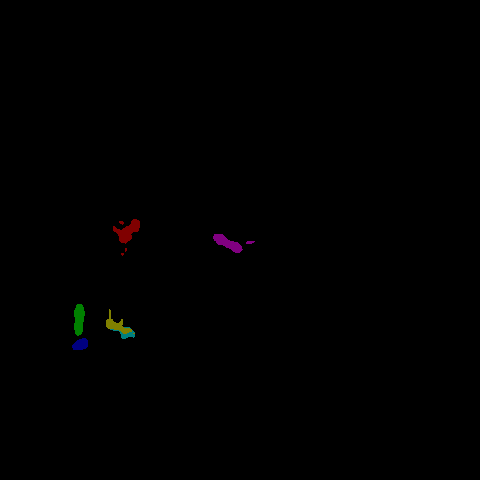}
}\\
\vspace{-3.5mm}\\

\rotatebox[origin=c]{90}{\small Stable aug. (ours) \hspace{1mm}}&
\raisebox{-0.5\height}{
    \includegraphics[width=0.15\linewidth]{imgs/ants/gt/00000001.png}
}&
\raisebox{-0.5\height}{
    \includegraphics[width=0.15\linewidth]{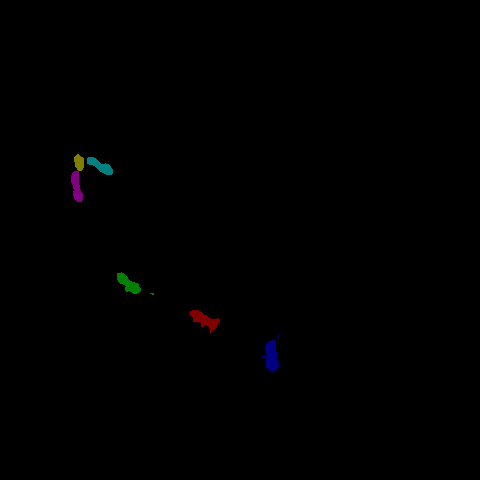}
}&
\raisebox{-0.5\height}{
    \includegraphics[width=0.15\linewidth]{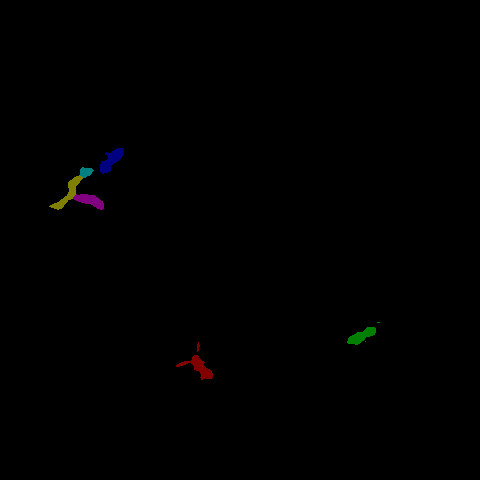}
}&
\raisebox{-0.5\height}{
    \includegraphics[width=0.15\linewidth]{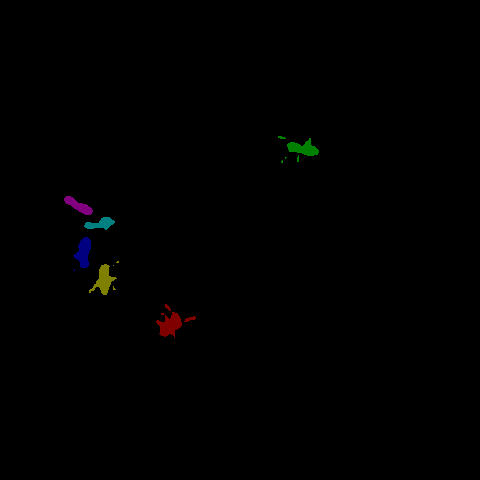}
}&
\raisebox{-0.5\height}{
    \includegraphics[width=0.15\linewidth]{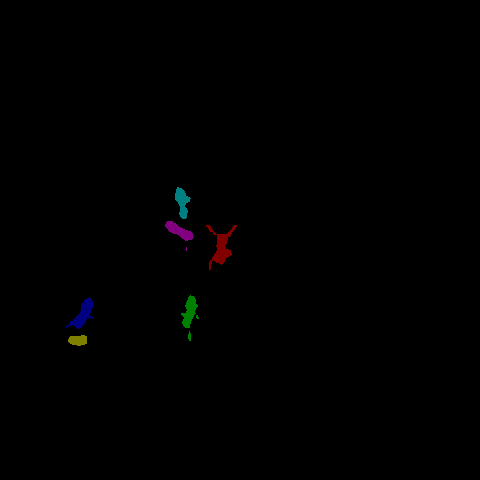}
}&
\raisebox{-0.5\height}{
    \includegraphics[width=0.15\linewidth]{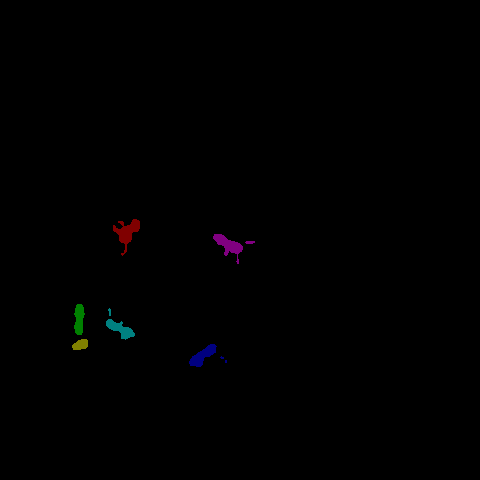}
}\\
\vspace{-3.5mm}\\

\rotatebox[origin=c]{90}{\small GT \hspace{1mm}}&
\raisebox{-0.5\height}{
    \includegraphics[width=0.15\linewidth]{imgs/ants/gt/00000001.png}
}&
\raisebox{-0.5\height}{
    \includegraphics[width=0.15\linewidth]{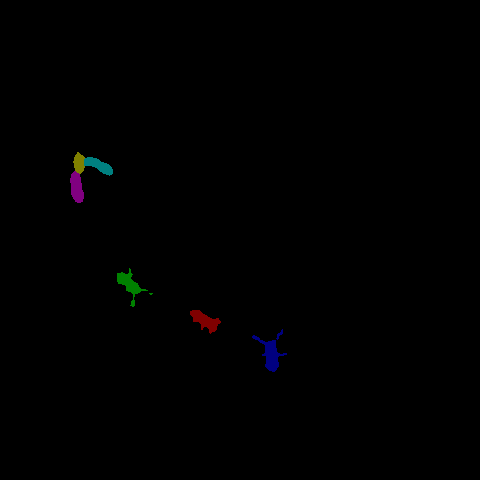}
}&
\raisebox{-0.5\height}{
    \includegraphics[width=0.15\linewidth]{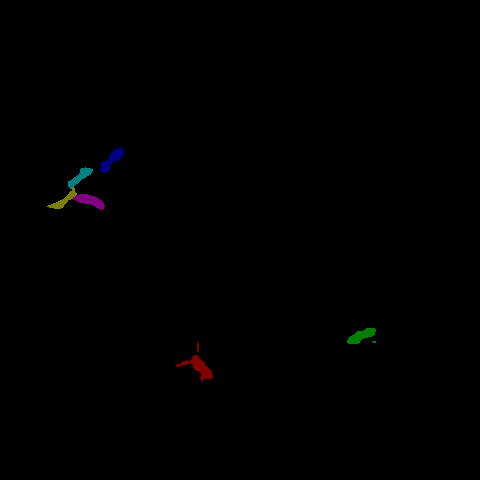}
}&
\raisebox{-0.5\height}{
    \includegraphics[width=0.15\linewidth]{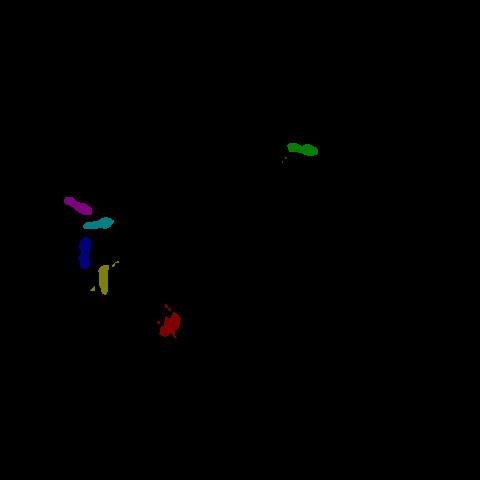}
}&
\raisebox{-0.5\height}{
    \includegraphics[width=0.15\linewidth]{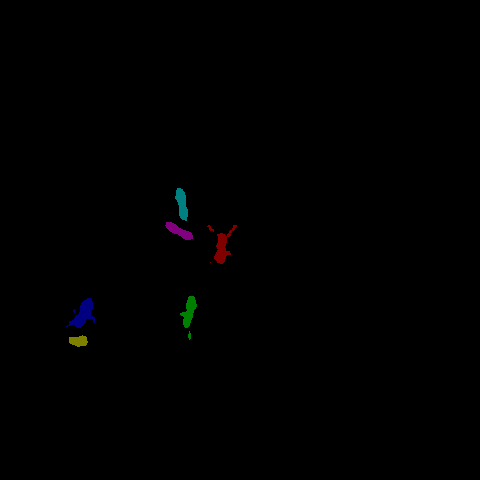}
}&
\raisebox{-0.5\height}{
    \includegraphics[width=0.15\linewidth]{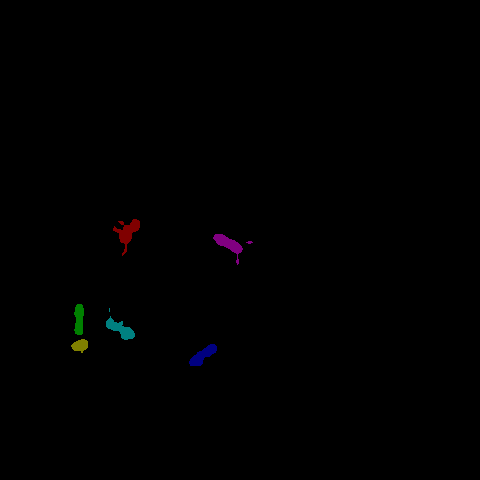}
}\\

& \small Frame 1 & \small Frame 60 & \small Frame 180 & \small Frame 220 & \small Frame 260 & \small Frame 300 \\
\end{tabular}

%% file: tabs/tab-app-vos-1.tex
\centering
\begin{tabular}
{l@{\hspace{10pt}}C{2.2em}@{}C{2.2em}@{}C{2.2em}@{\hspace{10pt}}C{2.2em}@{}C{2.2em}@{}C{2.2em}@{\hspace{10pt}}C{2.2em}@{}C{2.2em}@{}C{2.2em}@{\hspace{10pt}}C{2.2em}@{}C{2.2em}@{}C{2.2em}@{}C{2.2em}@{}C{2.2em}@{}R{2em}}
\toprule
& \multicolumn{3}{c}{MOSE}  & \multicolumn{3}{c}{\hspace{-12pt}DAVIS-17 val} & \multicolumn{3}{c}{\hspace{-12pt}DAVIS-17 test-dev} & \multicolumn{6}{c}{YouTubeVOS-2019 val} \\
\cmidrule(lr{\dimexpr 4\tabcolsep+12pt}){2-5} \cmidrule(lr{\dimexpr 4\tabcolsep+12pt}){5-8} \cmidrule(lr{\dimexpr 4\tabcolsep+12pt}){8-11} \cmidrule(lr){11-16}
Method & \mjf & \mj & \mf & \mjf & \mj & \mf & \mjf & \mj & \mf & \mg & \mjs & \mfs & \mju & \mfu & FPS \\
\midrule
STCN~\cite{cheng2021stcn} & 52.5 & 48.5 & 56.6 & 85.4 & 82.2 & 88.6 & 76.1 & 72.7 & 79.6 & 82.7 & 81.1 & 85.4 & 78.2 & 85.9 & 13.2 \\
AOT-R50~\cite{yang2021associating} & 58.4 & 54.3 & 62.6 & 84.9 & 82.3 & 87.5 & 79.6 & 75.9 & 83.3 & 85.3 & 83.9 & 88.8 & 79.9 & 88.5 & 6.4 \\
XMem~\cite{cheng2022xmem} & 56.3 & 52.1 & 60.6 & 86.2 & 82.9 & 89.5 & 81.0 & 77.4 & 84.5 & 85.5 & 84.3 & 88.6 & 80.3 & 88.6 & 22.6 \\
DEVA (ours), w/o OVIS & 60.0 & 55.8 & 64.3 & 86.8 & 83.6 & 90.0 & 82.3 & 78.7 & 85.9 & 85.5 & 85.0 & 89.4 & 79.7 & 88.0 & \textbf{25.3} \\
DEVA (ours), w/ OVIS & \textbf{66.5} & \textbf{62.3} & \textbf{70.8} & \textbf{87.6} & \textbf{84.2} & \textbf{91.0} & \textbf{83.2} & \textbf{79.6} & \textbf{86.8} & \textbf{86.2} & \textbf{85.4} & \textbf{89.9} & \textbf{80.5} & \textbf{89.1} & \textbf{25.3} \\
\midrule
\bottomrule
\end{tabular}

%% file: tabs/tab-app-temporal-ablation-vos.tex
\centering
\begin{tabular}{l@{\hspace{2mm}}c@{\hspace{2mm}}c@{\hspace{2mm}}c@{\hspace{2mm}}c@{\hspace{2mm}}c}
\toprule
Variant & DAVIS & OVIS  & UVO & FPS \\
\midrule
XMem~\cite{cheng2022xmem} & 86.1 & 69.0 & 82.7 & 22.6 \\
XMem~\cite{cheng2022xmem} train w/ OVIS & 86.1 & 72.0 & 83.0 & 22.6 \\
\midrule
With all our modifications & \textbf{87.6} & \textbf{75.7} & \textbf{83.5} & \textbf{25.8} \\
w/o stable data aug. & 87.5 & 73.6 & 83.2 & \textbf{25.8} \\
w/o gradient clipping & 85.2 & 71.3 & 82.7  & \textbf{25.8} \\
\midrule
\bottomrule
\end{tabular}

%% file: tabs/tab-app-temporal-ablation-vipseg.tex
\centering
\begin{tabular}{lc@{\hspace{2mm}}c@{\hspace{2mm}}c@{\hspace{2mm}}c@{\hspace{2mm}}c@{\hspace{2mm}}c@{\hspace{2mm}}c@{\hspace{2mm}}c@{\hspace{2mm}}c@{\hspace{2mm}}c}
\toprule
\textbf{\textit{Varying Temporal Propagation Model}} & VPQ$^1$ & VPQ$^2$ & VPQ$^4$ & VPQ$^6$ & VPQ$^{8}$ & VPQ$^{10}$ & VPQ$^\infty$ & $\overline{\text{VPQ}}$ & STQ\\
\midrule
With standard XMem~\cite{cheng2022xmem} & 41.9 & 41.3 & 40.6 & 39.9 & 39.5 & 39.0 & 35.4 & 37.9 & 41.3 \\
With our modified XMem~\cite{cheng2022xmem} & \textbf{42.1} & \textbf{41.5} & \textbf{40.8} & \textbf{40.1} & \textbf{39.7} & \textbf{39.3} & \textbf{36.1} & \textbf{38.3} & \textbf{41.5} \\
\midrule
\textbf{\textit{Varying Training Data of Temporal Propagation}} & VPQ$^1$ & VPQ$^2$ & VPQ$^4$ & VPQ$^6$ & VPQ$^{8}$ & VPQ$^{10}$ & VPQ$^\infty$ & $\overline{\text{VPQ}}$ & STQ\\
\midrule
Image pretraining + 100\% video training & \textbf{42.1} & \textbf{41.5} & \textbf{40.8} & \textbf{40.1} & \textbf{39.7} & \textbf{39.3} & \textbf{36.1} & \textbf{38.3} & \textbf{41.5} \\
Image pretraining + 50\%  video training & 42.0 & 41.4 & 40.7 & 40.1 & 39.7 & 39.4 & 36.0 & 38.3 & 41.3 \\
Image pretraining + 10\% video training & 40.7 & 40.1 & 39.3 & 38.5 & 38.1 & 37.7 & 34.6 & 36.8 & 40.1 \\
\midrule
Training on 100\% YouTube-VOS~\cite{xu2018youtubeVOS} only & 41.4 & 40.9 & 40.2 & 39.5 & 39.1 & 38.7 & 35.6 & 37.8 & 41.0 \\
Training on 50\% YouTube-VOS~\cite{xu2018youtubeVOS} only & 40.5 & 39.4 & 38.0 & 36.6 & 35.6 & 34.4 & 31.3 & 34.4 & 37.8 \\

\midrule
\bottomrule
\end{tabular}

%% file: tabs/tab-app-class-overlap.tex
\centering
\begin{tabular}{l@{\hspace{0mm}}c@{\hspace{2mm}}c@{\hspace{2mm}}c}
\toprule
Method & $\overline{\text{VPQ}}_{\text{overlap}}$ & $\overline{\text{VPQ}}_{\text{no-overlap}}$ & $\Delta_{\text{overlap}\to\text{non-overlap}}$ \\
\midrule
Video-K-Net & 25.0 & 25.7 & +0.7 \\
Ours & 38.1 & 38.6 & +0.5 \\
\midrule
\bottomrule
\end{tabular}

%% file: tabs/tab-app-burst-deta-assa.tex
\centering
\begin{tabular}{llccccccccc}
\toprule
& & \multicolumn{9}{c}{\textbf{Validation}} \\
\cmidrule(lr){3-11}
& & \multicolumn{3}{c}{\textbf{All}} & \multicolumn{3}{c}{\textbf{Common}} & \multicolumn{3}{c}{\textbf{Uncommon}} \\
\cmidrule(lr){3-5} \cmidrule(lr){6-8} \cmidrule(lr){9-11}
Method & & DetRe & AssA & OWTA & DetRe & AssA & OWTA & DetRe & AssA & OWTA \\
\midrule
Mask2Former & w/ Box tracker~\cite{athar2023burst} & 66.9 & 55.8 & 60.9 & 78.7 & 57.1 & 60.9 & 20.1 & 30.5 & 24.0 \\
Mask2Former & w/ STCN tracker~\cite{athar2023burst} & 67.0 & 62.6 & 64.6 & 78.8 & 64.1 & 71.0 & 20.0 & 33.3 & 25.0 \\
OWTB~\cite{liu2022opening} & & 70.9 & 45.2 & 56.2 & 76.8 & 47.0 & 59.8 & 46.5 & 34.3 & 38.5 \\
Mask2Former & w/ ours online & 72.1 & 67.5 & 69.5 & 80.2 & 69.9 & 74.6 & 39.8 & 46.4 & 42.3 \\
Mask2Former & w/ ours semi-online & 71.8 & \textbf{68.5} & \textbf{69.9} & \textbf{80.3} & \textbf{70.7} & \textbf{75.2} & 37.9 & 46.8 & 41.5 \\
EntitySeg & w/ ours online & 72.3 & 66.0 & 68.8 & 77.7 & 68.4 & 72.7 & \textbf{50.3} & 50.2 & 49.6 \\
EntitySeg & w/ ours semi-online & \textbf{72.4} & 67.1 & 69.5 & 78.1 & 69.3 & 73.3 & 50.0 & \textbf{52.2} & \textbf{50.5} \\
\midrule
& & \multicolumn{9}{c}{\textbf{Test}} \\
\cmidrule(lr){3-11}
& & \multicolumn{3}{c}{\textbf{All}} & \multicolumn{3}{c}{\textbf{Common}} & \multicolumn{3}{c}{\textbf{Uncommon}} \\
\cmidrule(lr){3-5} \cmidrule(lr){6-8} \cmidrule(lr){9-11}
Method & & DetRe & AssA & OWTA & DetRe & AssA & OWTA & DetRe & AssA & OWTA \\
\midrule
Mask2Former & w/ Box tracker~\cite{athar2023burst} & 61.5 & 51.1 & 55.9 & 71.4 & 52.5 & 61.0 & 21.1 & 30.0 & 24.6 \\
Mask2Former & w/ STCN tracker~\cite{athar2023burst} & 61.6 & 54.1 & 57.5 & 71.5 & 55.7 & 62.9 & 21.0 & 28.6 & 23.9 \\
OWTB~\cite{liu2022opening} & & 70.9 & 45.2 & 56.2 & 76.8 & 47.0 & 59.8 & 46.5 & 34.3 & 38.5 \\
Mask2Former & w/ ours online & 72.2 & 68.6 & 70.1 & \textbf{79.8} & 70.8 & 75.0 & 40.7 & 49.2 & 44.1 \\
Mask2Former & w/ ours semi-online & 71.9 & \textbf{69.6} & \textbf{70.5} & 79.7 & \textbf{71.7} & \textbf{75.4} & 39.5 & 50.7 & 44.1 \\
EntitySeg & w/ ours online & \textbf{72.5} & 67.3 & 69.5 & 77.3 & 69.2 & 72.9 & \textbf{52.3} & 55.0 & 53.0 \\
EntitySeg & w/ ours semi-online & 72.4 & 67.7 & 69.8 & 77.4 & 69.5 & 73.1 & 51.9 & \textbf{55.9} & \textbf{53.3} \\
\midrule
\bottomrule
\end{tabular}

%% file: main.bbl
\begin{thebibliography}{10}\itemsep=-1pt

\bibitem{athar2023tarvis}
Ali Athar, Alexander Hermans, Jonathon Luiten, Deva Ramanan, and Bastian Leibe.
\newblock Tarvis: A unified approach for target-based video segmentation.
\newblock {\em arXiv preprint arXiv:2301.02657}, 2023.

\bibitem{athar2023burst}
Ali Athar, Jonathon Luiten, Paul Voigtlaender, Tarasha Khurana, Achal Dave,
  Bastian Leibe, and Deva Ramanan.
\newblock Burst: A benchmark for unifying object recognition, segmentation and
  tracking in video.
\newblock In {\em WACV}, 2023.

\bibitem{bergmann2019tracking}
Philipp Bergmann, Tim Meinhardt, and Laura Leal-Taixe.
\newblock Tracking without bells and whistles.
\newblock In {\em ICCV}, 2019.

\bibitem{bertasius2020classifying}
Gedas Bertasius and Lorenzo Torresani.
\newblock Classifying, segmenting, and tracking object instances in video with
  mask propagation.
\newblock In {\em CVPR}, 2020.

\bibitem{caelles2019}
Sergi Caelles, Jordi Pont-Tuset, Federico Perazzi, Alberto Montes,
  Kevis-Kokitsi Maninis, and Luc Van~Gool.
\newblock The 2019 davis challenge on vos: Unsupervised multi-object
  segmentation.
\newblock In {\em arXiv preprint arXiv:1905.00737}, 2019.

\bibitem{cheng2021}
Bowen Cheng, Anwesa~Choudhuri andd Ishan~Misra, Alexander Kirillov, Rohit
  Girdhar, and Alexander~G. Schwing.
\newblock Mask2former for video instance segmentation.
\newblock In {\em https://arxiv.org/abs/2112.10764}, 2021.

\bibitem{cheng2022masked}
Bowen Cheng, Ishan Misra, Alexander~G Schwing, Alexander Kirillov, and Rohit
  Girdhar.
\newblock Masked-attention mask transformer for universal image segmentation.
\newblock In {\em CVPR}, 2022.

\bibitem{cheng2020cascadepsp}
Ho~Kei Cheng, Jihoon Chung, Yu-Wing Tai, and Chi-Keung Tang.
\newblock Cascadepsp: Toward class-agnostic and very high-resolution
  segmentation via global and local refinement.
\newblock In {\em CVPR}, 2020.

\bibitem{cheng2022xmem}
Ho~Kei Cheng and Alexander~G Schwing.
\newblock Xmem: Long-term video object segmentation with an atkinson-shiffrin
  memory model.
\newblock In {\em ECCV}, 2022.

\bibitem{cheng2021mivos}
Ho~Kei Cheng, Yu-Wing Tai, and Chi-Keung Tang.
\newblock Modular interactive video object segmentation: Interaction-to-mask,
  propagation and difference-aware fusion.
\newblock In {\em CVPR}, 2021.

\bibitem{cheng2021stcn}
Ho~Kei Cheng, Yu-Wing Tai, and Chi-Keung Tang.
\newblock Rethinking space-time networks with improved memory coverage for
  efficient video object segmentation.
\newblock In {\em NeurIPS}, 2021.

\bibitem{cheng2023segment}
Yangming Cheng, Liulei Li, Yuanyou Xu, Xiaodi Li, Zongxin Yang, Wenguan Wang,
  and Yi Yang.
\newblock Segment and track anything.
\newblock In {\em arXiv preprint arXiv:2305.06558}, 2023.

\bibitem{ChoudhuriICCV2021}
A. Choudhuri, G. Chowdhary, and A.~G. Schwing.
\newblock {Assignment-Space-Based Multi-Object Tracking and Segmentation}.
\newblock In {\em ICCV}, 2021.

\bibitem{ChoudhuriCVPR2023}
A. Choudhuri, G. Chowdhary, and A.~G. Schwing.
\newblock {Context-Aware Relative Object Queries to Unify Video Instance and
  Panoptic Segmentation}.
\newblock In {\em CVPR}, 2023.

\bibitem{chung2014empirical}
Junyoung Chung, Caglar Gulcehre, KyungHyun Cho, and Yoshua Bengio.
\newblock Empirical evaluation of gated recurrent neural networks on sequence
  modeling.
\newblock {\em NIPS Workshop}, 2014.

\bibitem{MOSE}
Henghui Ding, Chang Liu, Shuting He, Xudong Jiang, Philip~HS Torr, and Song
  Bai.
\newblock {MOSE}: A new dataset for video object segmentation in complex
  scenes.
\newblock In {\em ICCV}, 2023.

\bibitem{ding2022vlt}
Henghui Ding, Chang Liu, Suchen Wang, and Xudong Jiang.
\newblock Vlt: Vision-language transformer and query generation for referring
  segmentation.
\newblock In {\em TPAMI}, 2022.

\bibitem{du2021uvo}
Yuming Du, Wen Guo, Yang Xiao, and Vincent Lepetit.
\newblock Uvo challenge on video-based open-world segmentation 2021: 1st place
  solution.
\newblock {\em ICCV Workshop}, 2021.

\bibitem{garg2021mask}
Shubhika Garg and Vidit Goel.
\newblock Mask selection and propagation for unsupervised video object
  segmentation.
\newblock In {\em WACV}, 2021.

\bibitem{goel2021msn}
Vidit Goel, Jiachen Li, Shubhika Garg, Harsh Maheshwari, and Humphrey Shi.
\newblock Msn: efficient online mask selection network for video instance
  segmentation.
\newblock In {\em CVPR Workshop}, 2021.

\bibitem{he2016deepResNet}
Kaiming He, Xiangyu Zhang, Shaoqing Ren, and Jian Sun.
\newblock Deep residual learning for image recognition.
\newblock In {\em CVPR}, 2016.

\bibitem{huang2022minvis}
De-An Huang, Zhiding Yu, and Anima Anandkumar.
\newblock Minvis: A minimal video instance segmentation framework without
  video-based training.
\newblock In {\em NeurIPS}, 2022.

\bibitem{hwang2021video}
Sukjun Hwang, Miran Heo, Seoung~Wug Oh, and Seon~Joo Kim.
\newblock Video instance segmentation using inter-frame communication
  transformers.
\newblock {\em NeurIPS}, 2021.

\bibitem{sam_hq}
Lei Ke, Mingqiao Ye, Martin Danelljan, Yifan Liu, Yu-Wing Tai, Chi-Keung Tang,
  and Fisher Yu.
\newblock Segment anything in high quality.
\newblock In {\em arXiv}, 2023.

\bibitem{khoreva2019video}
Anna Khoreva, Anna Rohrbach, and Bernt Schiele.
\newblock Video object segmentation with language referring expressions.
\newblock In {\em ACCV}, 2019.

\bibitem{kim2015multiple}
Chanho Kim, Fuxin Li, Arridhana Ciptadi, and James~M Rehg.
\newblock Multiple hypothesis tracking revisited.
\newblock In {\em ICCV}, 2015.

\bibitem{kim2020video}
Dahun Kim, Sanghyun Woo, Joon-Young Lee, and In~So Kweon.
\newblock Video panoptic segmentation.
\newblock In {\em CVPR}, 2020.

\bibitem{kindratenko2020hal}
Volodymyr Kindratenko, Dawei Mu, Yan Zhan, John Maloney, Sayed~Hadi Hashemi,
  Benjamin Rabe, Ke Xu, Roy Campbell, Jian Peng, and William Gropp.
\newblock Hal: Computer system for scalable deep learning.
\newblock In {\em PEARC}, 2020.

\bibitem{kirillov2019panoptic}
Alexander Kirillov, Kaiming He, Ross Girshick, Carsten Rother, and Piotr
  Doll{\'a}r.
\newblock Panoptic segmentation.
\newblock In {\em CVPR}, 2019.

\bibitem{kirillov2023segment}
Alexander Kirillov, Eric Mintun, Nikhila Ravi, Hanzi Mao, Chloe Rolland, Laura
  Gustafson, Tete Xiao, Spencer Whitehead, Alexander~C Berg, Wan-Yen Lo, et~al.
\newblock Segment anything.
\newblock In {\em arXiv preprint arXiv:2304.02643}, 2023.

\bibitem{lee2023unsupervised}
Minhyeok Lee, Suhwan Cho, Seunghoon Lee, Chaewon Park, and Sangyoun Lee.
\newblock Unsupervised video object segmentation via prototype memory network.
\newblock In {\em WACV}, 2023.

\bibitem{li2023semantic}
Feng Li, Hao Zhang, Peize Sun, Xueyan Zou, Shilong Liu, Jianwei Yang, Chunyuan
  Li, Lei Zhang, and Jianfeng Gao.
\newblock Semantic-sam: Segment and recognize anything at any granularity.
\newblock In {\em arXiv}, 2023.

\bibitem{li2020fss}
Xiang Li, Tianhan Wei, Yau~Pun Chen, Yu-Wing Tai, and Chi-Keung Tang.
\newblock Fss-1000: A 1000-class dataset for few-shot segmentation.
\newblock In {\em CVPR}, 2020.

\bibitem{li2022video}
Xiangtai Li, Wenwei Zhang, Jiangmiao Pang, Kai Chen, Guangliang Cheng, Yunhai
  Tong, and Chen~Change Loy.
\newblock Video k-net: A simple, strong, and unified baseline for video
  segmentation.
\newblock In {\em CVPR}, 2022.

\bibitem{li2021panopticfcn}
Yanwei Li, Hengshuang Zhao, Xiaojuan Qi, Liwei Wang, Zeming Li, Jian Sun, and
  Jiaya Jia.
\newblock Fully convolutional networks for panoptic segmentation.
\newblock In {\em CVPR}, 2021.

\bibitem{lin2021video}
Huaijia Lin, Ruizheng Wu, Shu Liu, Jiangbo Lu, and Jiaya Jia.
\newblock Video instance segmentation with a propose-reduce paradigm.
\newblock In {\em ICCV}, 2021.

\bibitem{lin2014microsoft}
Tsung-Yi Lin, Michael Maire, Serge Belongie, James Hays, Pietro Perona, Deva
  Ramanan, Piotr Doll{\'a}r, and C~Lawrence Zitnick.
\newblock Microsoft coco: Common objects in context.
\newblock In {\em ECCV}, 2014.

\bibitem{liu2023grounding}
Shilong Liu, Zhaoyang Zeng, Tianhe Ren, Feng Li, Hao Zhang, Jie Yang, Chunyuan
  Li, Jianwei Yang, Hang Su, Jun Zhu, et~al.
\newblock Grounding dino: Marrying dino with grounded pre-training for open-set
  object detection.
\newblock In {\em arXiv preprint arXiv:2303.05499}, 2023.

\bibitem{liu2022opening}
Yang Liu, Idil~Esen Zulfikar, Jonathon Luiten, Achal Dave, Deva Ramanan,
  Bastian Leibe, Aljo{\v{s}}a O{\v{s}}ep, and Laura Leal-Taix{\'e}.
\newblock Opening up open world tracking.
\newblock In {\em CVPR}, 2022.

\bibitem{liu2021swin}
Ze Liu, Yutong Lin, Yue Cao, Han Hu, Yixuan Wei, Zheng Zhang, Stephen Lin, and
  Baining Guo.
\newblock Swin transformer: Hierarchical vision transformer using shifted
  windows.
\newblock In {\em ICCV}, 2021.

\bibitem{loshchilov2017decoupledAdamW}
Ilya Loshchilov and Frank Hutter.
\newblock Decoupled weight decay regularization.
\newblock In {\em ICLR}, 2019.

\bibitem{luiten2021hota}
Jonathon Luiten, Aljosa Osep, Patrick Dendorfer, Philip Torr, Andreas Geiger,
  Laura Leal-Taix{\'e}, and Bastian Leibe.
\newblock Hota: A higher order metric for evaluating multi-object tracking.
\newblock {\em International journal of computer vision}, 129:548--578, 2021.

\bibitem{luiten2020unovost}
Jonathon Luiten, Idil~Esen Zulfikar, and Bastian Leibe.
\newblock Unovost: Unsupervised offline video object segmentation and tracking.
\newblock In {\em WACV}, 2020.

\bibitem{mao2016generation}
Junhua Mao, Jonathan Huang, Alexander Toshev, Oana Camburu, Alan~L Yuille, and
  Kevin Murphy.
\newblock Generation and comprehension of unambiguous object descriptions.
\newblock In {\em CVPR}, 2016.

\bibitem{miao2022large}
Jiaxu Miao, Xiaohan Wang, Yu Wu, Wei Li, Xu Zhang, Yunchao Wei, and Yi Yang.
\newblock Large-scale video panoptic segmentation in the wild: A benchmark.
\newblock In {\em CVPR}, 2022.

\bibitem{oh2019videoSTM}
Seoung~Wug Oh, Joon-Young Lee, Ning Xu, and Seon~Joo Kim.
\newblock Video object segmentation using space-time memory networks.
\newblock In {\em ICCV}, 2019.

\bibitem{perazzi2016benchmark}
Federico Perazzi, Jordi Pont-Tuset, Brian McWilliams, Luc Van~Gool, Markus
  Gross, and Alexander Sorkine-Hornung.
\newblock A benchmark dataset and evaluation methodology for video object
  segmentation.
\newblock In {\em CVPR}, 2016.

\bibitem{qi2022occluded}
Jiyang Qi, Yan Gao, Yao Hu, Xinggang Wang, Xiaoyu Liu, Xiang Bai, Serge
  Belongie, Alan Yuille, Philip~HS Torr, and Song Bai.
\newblock Occluded video instance segmentation.
\newblock {\em IJCV}, 2022.

\bibitem{qi2021open}
Lu Qi, Jason Kuen, Yi Wang, Jiuxiang Gu, Hengshuang Zhao, Zhe Lin, Philip Torr,
  and Jiaya Jia.
\newblock Open-world entity segmentation.
\newblock In {\em arXiv preprint arXiv:2107.14228}, 2021.

\bibitem{qiao2021vip}
Siyuan Qiao, Yukun Zhu, Hartwig Adam, Alan Yuille, and Liang-Chieh Chen.
\newblock Vip-deeplab: Learning visual perception with depth-aware video
  panoptic segmentation.
\newblock In {\em CVPR}, 2021.

\bibitem{qin2022highly}
Xuebin Qin, Hang Dai, Xiaobin Hu, Deng-Ping Fan, Ling Shao, and Luc Van~Gool.
\newblock Highly accurate dichotomous image segmentation.
\newblock In {\em ECCV}, 2022.

\bibitem{radford2018improving}
Alec Radford, Karthik Narasimhan, Tim Salimans, Ilya Sutskever, et~al.
\newblock Improving language understanding by generative pre-training.
\newblock 2018.

\bibitem{rajivc2023segment}
Frano Raji{\v{c}}, Lei Ke, Yu-Wing Tai, Chi-Keung Tang, Martin Danelljan, and
  Fisher Yu.
\newblock Segment anything meets point tracking.
\newblock In {\em arXiv preprint arXiv:2307.01197}, 2023.

\bibitem{ren2021reciprocal}
Sucheng Ren, Wenxi Liu, Yongtuo Liu, Haoxin Chen, Guoqiang Han, and Shengfeng
  He.
\newblock Reciprocal transformations for unsupervised video object
  segmentation.
\newblock In {\em CVPR}, 2021.

\bibitem{seo2020urvos}
Seonguk Seo, Joon-Young Lee, and Bohyung Han.
\newblock Urvos: Unified referring video object segmentation network with a
  large-scale benchmark.
\newblock In {\em ECCV}, 2020.

\bibitem{shi2015hierarchicalECSSD}
Jianping Shi, Qiong Yan, Li Xu, and Jiaya Jia.
\newblock Hierarchical image saliency detection on extended cssd.
\newblock In {\em TPAMI}, 2015.

\bibitem{sofiiuk2020f}
Konstantin Sofiiuk, Ilia Petrov, Olga Barinova, and Anton Konushin.
\newblock f-brs: Rethinking backpropagating refinement for interactive
  segmentation.
\newblock In {\em CVPR}, 2020.

\bibitem{tang2017multiple}
Siyu Tang, Mykhaylo Andriluka, Bjoern Andres, and Bernt Schiele.
\newblock Multiple people tracking by lifted multicut and person
  re-identification.
\newblock In {\em CVPR}, 2017.

\bibitem{teed2020raft}
Zachary Teed and Jia Deng.
\newblock Raft: Recurrent all-pairs field transforms for optical flow.
\newblock In {\em ECCV}, 2020.

\bibitem{wang2017DUTS}
Lijun Wang, Huchuan Lu, Yifan Wang, Mengyang Feng, Dong Wang, Baocai Yin, and
  Xiang Ruan.
\newblock Learning to detect salient objects with image-level supervision.
\newblock In {\em CVPR}, 2017.

\bibitem{wang2021unidentified}
Weiyao Wang, Matt Feiszli, Heng Wang, and Du Tran.
\newblock Unidentified video objects: A benchmark for dense, open-world
  segmentation.
\newblock In {\em CVPR}, 2021.

\bibitem{wang2021end}
Yuqing Wang, Zhaoliang Xu, Xinlong Wang, Chunhua Shen, Baoshan Cheng, Hao Shen,
  and Huaxia Xia.
\newblock End-to-end video instance segmentation with transformers.
\newblock In {\em CVPR}, 2021.

\bibitem{weber2021step}
Mark Weber, Jun Xie, Maxwell Collins, Yukun Zhu, Paul Voigtlaender, Hartwig
  Adam, Bradley Green, Andreas Geiger, Bastian Leibe, Daniel Cremers, et~al.
\newblock Step: Segmenting and tracking every pixel.
\newblock In {\em NeurIPS}, 2021.

\bibitem{wu2022language}
Jiannan Wu, Yi Jiang, Peize Sun, Zehuan Yuan, and Ping Luo.
\newblock Language as queries for referring video object segmentation.
\newblock In {\em CVPR}, 2022.

\bibitem{xu2018youtubeVOS}
Ning Xu, Linjie Yang, Yuchen Fan, Dingcheng Yue, Yuchen Liang, Jianchao Yang,
  and Thomas Huang.
\newblock Youtube-vos: A large-scale video object segmentation benchmark.
\newblock In {\em ECCV}, 2018.

\bibitem{yan2022towards}
Bin Yan, Yi Jiang, Peize Sun, Dong Wang, Zehuan Yuan, Ping Luo, and Huchuan Lu.
\newblock Towards grand unification of object tracking.
\newblock In {\em ECCV}, 2022.

\bibitem{yan2023universal}
Bin Yan, Yi Jiang, Jiannan Wu, Dong Wang, Ping Luo, Zehuan Yuan, and Huchuan
  Lu.
\newblock Universal instance perception as object discovery and retrieval.
\newblock In {\em CVPR}, 2023.

\bibitem{yang2023track}
Jinyu Yang, Mingqi Gao, Zhe Li, Shang Gao, Fangjing Wang, and Feng Zheng.
\newblock Track anything: Segment anything meets videos.
\newblock In {\em arXiv preprint arXiv:2304.11968}, 2023.

\bibitem{yang2019video}
Linjie Yang, Yuchen Fan, and Ning Xu.
\newblock Video instance segmentation.
\newblock In {\em ICCV}, 2019.

\bibitem{yang2021associating}
Zongxin Yang, Yunchao Wei, and Yi Yang.
\newblock Associating objects with transformers for video object segmentation.
\newblock In {\em NeurIPS}, 2021.

\bibitem{yu2016modeling}
Licheng Yu, Patrick Poirson, Shan Yang, Alexander~C Berg, and Tamara~L Berg.
\newblock Modeling context in referring expressions.
\newblock In {\em ECCV}, 2016.

\bibitem{zeng2019towardsHRSOD}
Yi Zeng, Pingping Zhang, Jianming Zhang, Zhe Lin, and Huchuan Lu.
\newblock Towards high-resolution salient object detection.
\newblock In {\em ICCV}, 2019.

\bibitem{mobile_sam}
Chaoning Zhang, Dongshen Han, Yu Qiao, Jung~Uk Kim, Sung-Ho Bae, Seungkyu Lee,
  and Choong~Seon Hong.
\newblock Faster segment anything: Towards lightweight sam for mobile
  applications.
\newblock In {\em arXiv}, 2023.

\bibitem{zhao2023fast}
Xu Zhao, Wenchao Ding, Yongqi An, Yinglong Du, Tao Yu, Min Li, Ming Tang, and
  Jinqiao Wang.
\newblock Fast segment anything.
\newblock In {\em arXiv}, 2023.

\bibitem{zou2022generalized}
Xueyan Zou, Zi-Yi Dou, Jianwei Yang, Zhe Gan, Linjie Li, Chunyuan Li, Xiyang
  Dai, Harkirat Behl, Jianfeng Wang, Lu Yuan, et~al.
\newblock Generalized decoding for pixel, image, and language.
\newblock In {\em CVPR}, 2023.

\bibitem{zou2023segment}
Xueyan Zou, Jianwei Yang, Hao Zhang, Feng Li, Linjie Li, Jianfeng Gao, and
  Yong~Jae Lee.
\newblock Segment everything everywhere all at once.
\newblock In {\em arXiv}, 2023.

\end{thebibliography}
